\documentclass{article} 
\usepackage{iclr2027_conference,times}
\iclrfinalcopy

\usepackage{amsmath,amsfonts,bm}

\def\eqref#1{equation~\ref{#1}}

\def\1{\bm{1}}

\DeclareMathAlphabet{\mathsfit}{\encodingdefault}{\sfdefault}{m}{sl}
\SetMathAlphabet{\mathsfit}{bold}{\encodingdefault}{\sfdefault}{bx}{n}

\usepackage{hyperref}
\usepackage{url}

\usepackage{graphicx}
\usepackage{xspace}
\usepackage{booktabs}
\usepackage{multirow}
\usepackage{enumitem}
\usepackage{array}
\usepackage{makecell}
\usepackage[table]{xcolor}
\usepackage{adjustbox}
\usepackage{minted}
\usepackage[most]{tcolorbox}
\usepackage{xcolor}
\usepackage{wrapfig}
\usepackage[T1]{fontenc}

\newcommand{\method}{\textbf{\textsc{\textcolor{mypurple2}{FraudBench}}}\xspace}
\newcommand{\huggingface}{\raisebox{-1.5pt}{\includegraphics[height=1.05em]{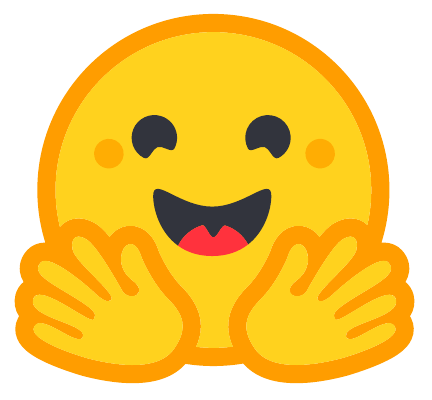}}\xspace}
\newcommand{\github}{\raisebox{-1.5pt}{\includegraphics[height=1.05em]{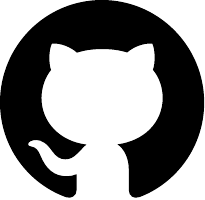}}\xspace}
\definecolor{mypurple}{HTML}{F1F0FF}
\definecolor{mypurple2}{HTML}{5252AD}
\definecolor{boxbg}{HTML}{F1F0FF}
\definecolor{boxframe}{HTML}{746FA8}
\definecolor{boxtitle}{HTML}{665C9E}

\newtcblisting{example}[2][]{
  listing only,
  colbacktitle = boxtitle,
  colframe = boxframe,
  colback = boxbg,
  coltitle = white,
  fonttitle=\bfseries\small,
  title={Example: #2},
  boxrule=1.2pt,
  arc=2mm,
  top=0mm,
  bottom=0mm,
  left=2mm,
  right=2mm,
  toptitle=0.5mm,
  bottomtitle=0.5mm,
  listing options={
    basicstyle=\ttfamily\footnotesize,
    breaklines=true,
    columns=fullflexible,
    keepspaces=true
  },
  #1
}

\newtcblisting{prompt}[2][]{
  listing only,
  colbacktitle = boxtitle,
  colframe = boxframe,
  colback = boxbg,
  coltitle = white,
  fonttitle=\bfseries\small,
  title={Prompt: #2},
  boxrule=1.2pt,
  arc=2mm,
  top=0mm,
  bottom=0mm,
  left=2mm,
  right=2mm,
  toptitle=0.5mm,
  bottomtitle=0.5mm,
  listing options={
    basicstyle=\ttfamily\footnotesize,
    breaklines=true,
    breakatwhitespace=true,
    breakindent=0pt,
    breakautoindent=false,
    columns=fullflexible,
    keepspaces=false,
    showstringspaces=false
  },
  #1
}

\newtcblisting{nocontent}[2][]{
  listing only,
  colbacktitle = boxtitle,
  colframe = boxframe,
  colback = boxbg,
  coltitle = white,
  fonttitle=\bfseries\small,
  title={#2},
  boxrule=1.2pt,
  arc=2mm,
  top=0mm,
  bottom=0mm,
  left=2mm,
  right=2mm,
  toptitle=0.5mm,
  bottomtitle=0.5mm,
  listing options={
    basicstyle=\ttfamily\footnotesize,
    breaklines=true,
    columns=fullflexible,
    keepspaces=true
  },
  #1
}

\title{\method{}: A Multimodal Benchmark for Detecting AI-Generated Fraudulent Refund Evidence}

\author{%
  Xinyu Yan$^{1,2}$, Boyang Chen$^{1}$, Jiaming Zhang$^{1}$, Tiantong Wu$^{1,2}$, Hong Xi Tae$^{1}$, \\ 
  \textbf{Yurong Hao$^{1}$, Yachun Mi$^{1}$, Tiantong Wang$^{1,2}$, Yichen He$^{1}$, Yilei Zhao$^{1}$,} \\ 
  \textbf{Lei Xiao$^{3}$, Longtao Huang$^{3}$, Pengjun Xie$^{3}$, Wei Liu$^{3}$, Wei Yang Bryan Lim$^{1}$} \\ 
  $^1$College of Computing and Data Science, Nanyang Technological University \\ 
  $^2$Alibaba-NTU Global e-Sustainability CorpLab (ANGEL) \\ 
  $^3$Alibaba Group \\
  \texttt{xinyu039@e.ntu.edu.sg} \quad \texttt{bryan.limwy@ntu.edu.sg}
  \vspace{0.4em} \\
  \github \textbf{Code}: \url{https://github.com/Tristan0318/FraudBench}
  \vspace{0.2em} \\
  \huggingface \textbf{Hugging Face Repo}:~\texttt{\href{https://huggingface.co/datasets/TristanYan/FraudBench}{TristanYan/FraudBench}}
  }

\begin{document}

\maketitle

\begin{abstract}
Artificial Intelligence (AI)-generated images have become increasingly realistic and readily adaptable to real-world claims, creating new challenges for verifying visual evidence. A concrete emerging risk is AI-generated refund fraud, in which manipulated or synthetic images are used to support claims about damaged products, poor delivery conditions, or service-related defects. Existing AI-generated image detection benchmarks mainly evaluate standalone authenticity classification, cross-generator transfer, or forensic localization, leaving claim-conditioned fraudulent evidence detection underexplored. To bridge this gap, we introduce \method{}, a multimodal benchmark for detecting AI-generated fraudulent refund evidence. \method{} is constructed from real-world user-review evidence across e-commerce, food delivery, and travel-service scenarios. We curate real evidence images together with their associated review and product metadata, identify genuine damaged and undamaged evidence through MLLM-assisted filtering and human annotation, and synthesize fake-damaged evidence from genuine undamaged reference images using 12 state-of-the-art image editing and generation models. Using \method{}, we evaluate MLLMs, specialized AI-generated image detectors, and human evaluators on the same evaluation images. Experiments show that current MLLMs often recognize real-damaged evidence but fail on many fake-damaged subsets, with fake-damage detection rates (TPR) far below the 50\% baseline on most generator subsets. Specialized detectors generally perform better but remain inconsistent across generators and can produce false positives on real-damaged samples, revealing a clear gap between generic AI image detection and reliable claim-conditioned refund-evidence verification.

\textcolor{red}{\textbf{Note:} This paper is intended solely for academic research purposes. The goal is not to facilitate refund fraud or provide actionable guidance for misuse.}
\end{abstract}

\section{Introduction}
The rapid evolution of generative artificial intelligence (AI) has reached a critical milestone: the visual quality of synthesized and edited images has crossed the threshold of human perception, making them nearly indistinguishable from authentic photographs~\citep{rombach2022high,saharia2022photorealistic}. While approaching or surpassing human-level realism represents a triumph in deep learning, it simultaneously introduces unprecedented vulnerabilities into trust-based digital ecosystems. A concrete and rapidly emerging threat is AI-generated refund fraud. On transaction-oriented platforms, from e-commerce to food delivery and travel services (e.g., Taobao, Lazada, and Amazon), malicious actors now leverage state-of-the-art image generation and editing models to fabricate highly realistic visual evidence. By seamlessly injecting claim-specific defects (e.g., shattered screens, spoiled food, or stained fabrics) into genuine product images, these manipulated visuals easily pass manual review and exploit dispute-resolution systems for illicit financial gain~\citep{chinadaily2025refundscams,chinadaily2025crabrefund}. This transforms the generic ``deepfake'' problem into an attack on platform integrity, where AI is weaponized to generate context-aware, fraudulent evidence.

Despite the growing urgency of this threat, existing detection benchmarks are not well-suited to evaluate it. Many established datasets were constructed before the recent leaps in generative capabilities, focusing largely on low-level artifacts from earlier models rather than the highly realistic edits prevalent today~\citep{rossler2019faceforensics++,zhu2023genimage}. Beyond image quality, there is also a structural mismatch between standard benchmark settings and how real-world evidence is presented. Most benchmarks treat images as isolated, single-view samples for generic authenticity classification~\citep{yan2024df40,wang2025dfbench}. In practice, however, refund evidence is rarely a standalone image. Users typically submit multiple photographs from different angles to demonstrate a specific defect, often accompanied by a textual claim or review. As a result, detecting fraudulent evidence goes beyond spotting pixel-level anomalies in a single picture; it requires verifying multi-view visual consistency and ensuring the images align with the user's stated claim. To properly evaluate this capability, the community needs to move beyond generic synthetic datasets and focus on realistic, transaction-oriented scenarios.

To bridge this gap, we introduce \method{}, a multimodal benchmark specifically designed for detecting AI-generated fraudulent refund evidence. Rather than collecting isolated images, we construct \method{} from 822 real-world user reviews across 29 diverse categories in e-commerce, food delivery, and travel services, yielding a total of 12,868 evaluation images. This comprehensive foundation naturally provides realistic, multi-view evidence paired with textual claims. To ensure the benchmark evaluates genuine verification capabilities rather than relying on ``shortcut learning'' (e.g., detecting generic AI styles or resolution drops), we employ a rigorous, controlled synthesis pipeline. We curate real, undamaged multi-view images as references and utilize 12 state-of-the-art image generation and editing models to synthesize \textit{fake-damaged} evidence. By applying category-specific damage patterns to authentic user photos, we preserve the original object, scene context, and lighting. This forces recognition models to look beyond superficial artifacts and perform rigorous cross-modal and multi-view consistency checks.

Using \method{}, we conduct a comprehensive evaluation of 30 Multimodal Large Language Models (MLLMs), four specialized detectors with eight checkpoints, and human evaluators. Our findings reveal a critical dilemma in current detection paradigms when faced with the asymmetric risks of real-world dispute resolution. In practice, false positives (wrongly accusing an honest customer of fraud) can trigger severe trust crises, while false negatives (failing to detect AI evidence) lead to direct financial losses. We observe that current models struggle to safely navigate this trade-off. MLLMs exhibit a tendency toward \textit{\textbf{over-credulity}}; while they rarely misjudge genuine images, their detection rates for AI-manipulated damage are low, often falling below random guessing. Conversely, specialized detectors tend toward \textit{\textbf{over-sensitivity}}; although they can identify synthetic manipulations, several checkpoints suffer from unacceptable false positive rates, frequently flagging genuine, unedited images of damaged goods as fake. Notably, while human evaluators outperform all evaluated MLLMs, they still exhibit non-negligible error rates, confirming that verifying these highly realistic manipulations remains a formidable challenge even for manual inspection. This collective failure underscores the unique difficulty of \method{} and highlights a profound gap in current multimodal verification capabilities. Our contributions are summarized as follows:
\begin{itemize}
    \item We introduce \method{}, to the best of our knowledge, the first multimodal benchmark dedicated to detecting AI-generated fraudulent refund evidence. Featuring 822 review samples and 12,868 evaluation images across 29 categories, it shifts the evaluation paradigm from standalone authenticity classification to claim-conditioned, multi-view evidence verification.
    \item We design a rigorous construction pipeline that synthesizes fake-damaged evidence from real-world references using 12 state-of-the-art generative models, ensuring the benchmark reflects realistic adversarial scenarios.
    \item We benchmark a diverse set of 30 MLLMs, four specialized detectors with eight checkpoints, and human evaluators. Our results expose a critical ``over-credulity vs.\ over-sensitivity'' dilemma, highlighting the urgent need for models capable of managing asymmetric risks in high-stakes applications.
\end{itemize}
\section{Related Work}
\paragraph{Benchmarks for AI-Generated Image Detection.}
Benchmarks for AI-generated image detection are essential for measuring whether detectors generalize across generators, visual domains, and deployment conditions. Recent efforts have moved beyond early settings that evaluate detectors on a narrow set of face manipulations or a small number of generators. Specifically, DF40 expands face deepfake evaluation to diverse manipulation techniques and evaluation settings~\citep{yan2024df40}. Community Forensics scales the generator space, highlighting generator diversity as a major bottleneck for cross-model generalization~\citep{park2025community}, while RRBench studies robustness under transmission and re-digitization~\citep{li2025bridging}. DFBench evaluates the deepfake detection capability of large multimodal models~\citep{wang2025dfbench}, and ForensicHub provides a unified benchmark and codebase across multiple image forensic domains~\citep{du2025forensichub}. Despite these advances, existing benchmarks mainly focus on generic authenticity classification, cross-generator transfer, or broad forensic localization. They do not directly evaluate refund-related fraudulent evidence, where an image must be assessed in relation to the specific claim it is intended to support.

\paragraph{AI-Generated Image Detectors.}
Detecting AI-generated images has become increasingly important as modern generative models can produce realistic images tailored for specific instructions and application scenarios. Existing approaches broadly fall into two categories. The first line of work develops specialized detectors for generalizable real/fake classification, from early artifact-based detection~\citep{wang2020cnn} to universal detectors for cross-generator generalization~\citep{ojha2023towards}. Recent methods improve generalization by combining foundation-model representations with local or frequency-level artifacts~\citep{yan2025sanity}, or by integrating semantic and pixel-level features~\citep{cheng2025co}. Structured and adaptive feature-learning methods further improve robustness across unseen generators and post-processing conditions~\citep{yan2025orthogonal,li2026towards}, yet these approaches remain optimized as task-specific binary classifiers without broader claim-level reasoning. A more recent direction explores multimodal large-model-based detection. Rather than only predicting real or fake labels, recent studies use MLLMs to provide artifact explanations, visual grounding, segmentation, or human-aligned reasoning~\citep{wen2025spot,zhou2025aigi,kang2025legion,ji2026fakexplain}. This direction makes AI-generated detection more interpretable and closer to user-facing authenticity assessment. However, both specialized detectors and MLLM-based detectors are primarily evaluated on generic synthetic images or broad forensic benchmarks, leaving their effectiveness for claim-conditioned fraudulent evidence detection unclear.
\section{Our Benchmark: \method{}}
\subsection{Overview of \method{}}
\begin{figure}[!t]
    \centering
    \includegraphics[width=1\linewidth]{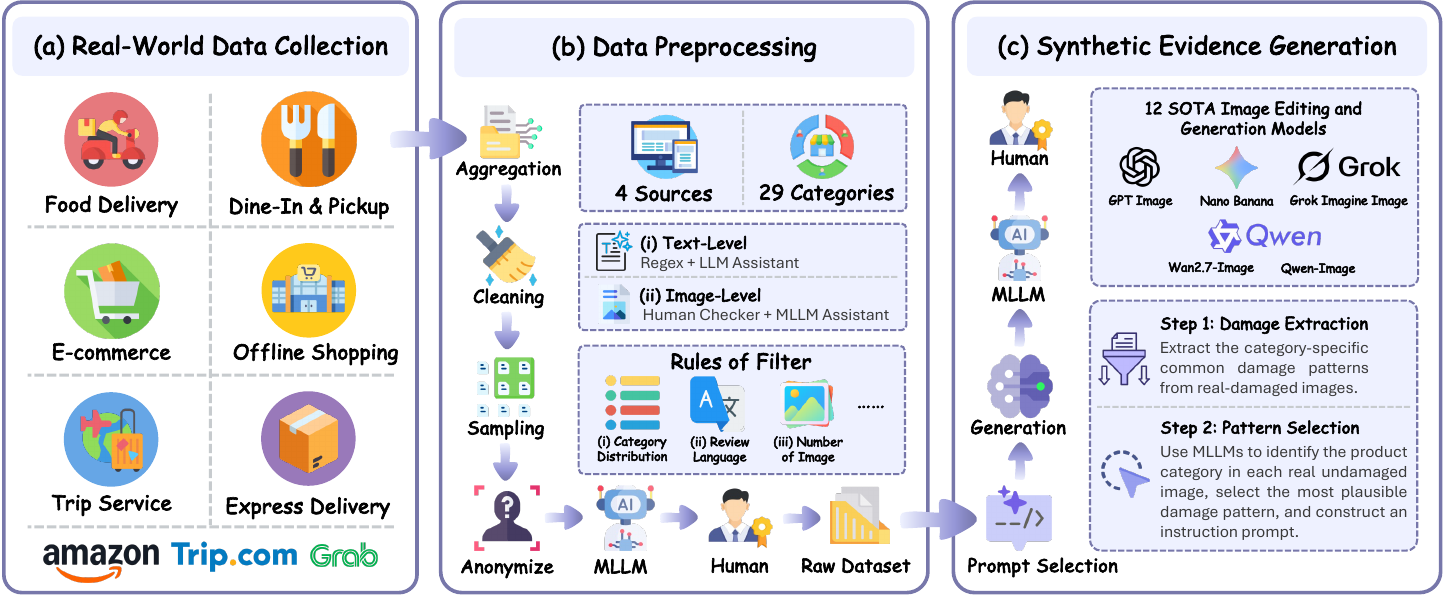}
    \caption{\textbf{Construction pipeline of our proposed \method{}.} The benchmark is constructed in three stages: \textbf{(1) Data Collection}: We collect real-world refund-related review data from four data sources and consolidate them into 29 categories. \textbf{(2) Data Preprocessing}: We perform multimodal aggregation, two-level cleaning, rule-based filtering, representative sampling, and anonymization, with screening and human verification for quality control. \textbf{(3) Synthetic Evidence Generation}: We construct image-specific prompts via a two-step procedure. SOTA models are then used to generate fraudulent refund evidence, which is further verified by human annotators.
    }
    \label{fig:Construction_Pipeline}
\end{figure}

\method{} is a multimodal benchmark for evaluating AI-generated fraudulent refund evidence detection under realistic transaction settings. It is constructed from a task-relevant subset of Amazon Reviews 2023~\citep{hou2024bridging}, self-collected review evidence from Trip.com and GrabFood, and a small set of in-house captured real-world samples. In total, \method{} contains \textbf{822} real review samples and \textbf{13,856} images. The 2,000 real images consist of 988 real-undamaged images and 1,012 real-damaged images. Using the 988 real-undamaged images as generation references, we further synthesize 11,856 fake-damaged images. Real-undamaged images serve solely as synthesis sources and are excluded from evaluation. The benchmark covers \textbf{four} data sources, \textbf{six} real-world transaction scenarios, and \textbf{29} product/service categories. The collected data span e-commerce, food delivery, trip service, offline shopping, dine-in \& pickup, and express delivery, supplemented by self-collected evidence. The detailed construction process is shown in Fig.~\ref{fig:Construction_Pipeline}. 

Unlike generic AI-generated image detection benchmarks, \method{} focuses on refund-evidence verification, where detectors must judge whether visual evidence is authentic and whether it is consistent with the associated review context. Each real review sample is organized as a multimodal evidence instance containing one or more images, review text, or a transaction-specific description, category metadata, and authenticity labels. For synthetic evidence, fake-damaged images are generated from retained real-undamaged images using \textbf{12} state-of-the-art image editing and generation models, together with image-specific prompts and corresponding fake review comments.

\begin{figure}[!ht]
    \centering
    \includegraphics[width=1\linewidth]{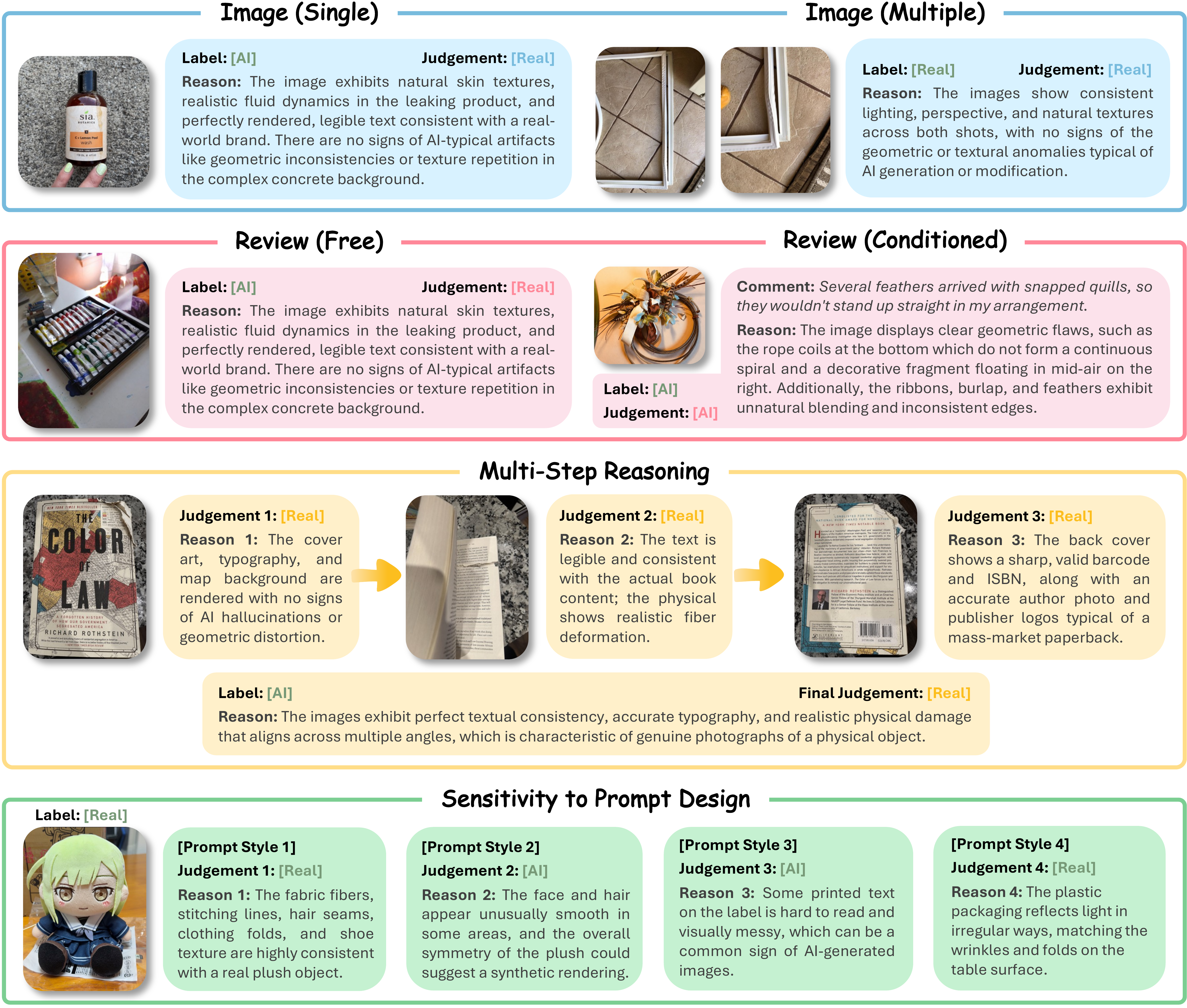}
    \caption{\textbf{Representative examples in \method{}.} We illustrate examples covering the main evaluation settings: input modality, contextual information, multi-step reasoning, and prompt sensitivity. Text prompts, rationales, and intermediate judgments are abbreviated for visualization.}
    \label{fig:Mode_Representative}
\end{figure}

As illustrated in Fig.~\ref{fig:Mode_Representative} and detailed in Sec.~\ref{sec:eval}, the benchmark is designed around five evaluation dimensions that reflect practical sources of uncertainty in refund-evidence verification. The input-modality dimension examines whether detectors remain reliable for either single images or multiple related images from the same review, while the contextual-information dimension tests whether review text helps distinguish genuine damage from synthetic or inconsistent evidence beyond visual cues alone. Meanwhile, the multi-step reasoning evaluates whether image-level decomposition improves claim-level decisions, and the prompt-sensitivity dimension measures the stability of detector outputs under semantically equivalent verification prompts. Finally, real-image preservation assesses whether models can detect synthetic refund evidence without over-flagging genuine user submissions, a prerequisite for trustworthy refund adjudication.

\subsection{Benchmark Construction}
\paragraph{Data Collection.} To evaluate AI-generated refund-evidence detection under realistic transaction settings, we construct the benchmark from four complementary data sources. The primary e-commerce source is Amazon Reviews 2023~\citep{hou2024bridging}, a large-scale review dataset with user reviews, item metadata, and image-related resources, from which we curate \textbf{590} task-relevant review samples across \textbf{27} product categories. Because Amazon Reviews 2023 was not designed for refund-evidence verification, repurposing it for this setting requires non-trivial review--image--metadata alignment, task-specific filtering, and new claim-conditioned damage annotations. Moreover, service-oriented refund scenarios remain outside its e-commerce scope.

We therefore add \textbf{225} non-Amazon review evidence from Trip.com and GrabFood, covering service-oriented scenarios such as travel services, food delivery, and dine-in \& pickup. Finally, we include a small set of in-house captured real evidence images collected under everyday conditions to improve coverage of practical visual acquisition settings. These sources provide real user-submitted or naturally captured images paired with review text or transaction-specific descriptions, allowing the benchmark to span diverse product categories, service contexts, image conditions, and damage or complaint patterns.

\paragraph{Data Preprocessing.}
After aggregating data from the four sources, we apply a multimodal preprocessing pipeline covering text cleaning, image cleaning and anonymization, image--review alignment filtering, distribution-aware sampling, and quality control. We use an MLLM as a first-stage screening model for both textual and visual content. For text cleaning, we remove empty, duplicated, corrupted, or structurally invalid reviews, and assess whether each review is relevant to the associated product, service, or complaint. For image cleaning and anonymization, the screening model flags invalid or risky samples, including privacy-sensitive images containing clear human faces or personal identifiers, technically invalid images such as corrupted files, duplicated images, screenshots, advertisements, or unreadable images, and task-irrelevant images that are unrelated to the target item or service. We further discard samples with weak image--review alignment, since review-conditioned detection requires consistency between textual context and visual evidence.

We then conduct distribution-aware sampling to preserve category diversity and evaluation usability. At the category level, we retain a post-filtering distribution that remains as close as possible to the distribution before sampling, avoiding over-representation of a small number of frequent categories while preserving the natural long-tail structure of the collected data. Within each category, we further control the image-count distribution across single-image and multi-image reviews, so that the benchmark supports both single-image evaluation and different levels of multi-image context. We also retain multilingual review text when available, rather than translating all reviews into English, to reflect real-world platform usage and evaluate robustness under multilingual user contexts. Finally, samples flagged during MLLM-based screening are verified by human annotators, who make the final inclusion decision before the samples are retained in the benchmark.

\paragraph{Synthetic Evidence Generation.} We generate synthetic fraudulent refund evidence from retained real-undamaged images through an image-specific prompt construction pipeline. We use 12 state-of-the-art image editing and generation models that are accessible through mainstream AI platforms or official API services, reflecting a realistic threat model for low-barrier synthetic evidence creation. The pipeline has two stages. First, we mine real damaged samples within each category to construct category-specific candidate damage lists, covering visually verifiable defects such as scratches, dents, stains, leakage, breakage, and related damage patterns. Second, for each target image, an MLLM analyzes the visible product, packaging, material, and physical attributes, and combines this visual analysis with product metadata, including the title, category, description, and features. The corresponding category-specific damage list is then retrieved, and the most plausible damage pattern is selected according to the object, material, scene, and metadata. When no category-specific candidate is suitable, the selection falls back to a universal damage vocabulary. The selected damage pattern is finally converted into an image-specific editing prompt.

Each prompt is shared across the 12 image editing and generation models, ensuring that the generated outputs are compared under the same editing intent. The prompt specifies that the modification should be visible, photorealistic, physically plausible, and localized to the target object, while preserving the original product identity, viewpoint, lighting, background, surrounding objects, and overall composition. We also generate a fake review comment for each synthetic sample for review-conditioned evaluation. Quality control verifies that generated comments are stylistically consistent with genuine user reviews and do not contain artifact-level linguistic cues that would permit reliable text-only classification, ensuring that claim verification requires joint reasoning over the visual evidence and the stated textual claim. This setup allows the benchmark to evaluate both visual authenticity detection and image--text consistency in refund-claim contexts. Generated samples are subsequently checked using the same quality-control procedure and regenerated when implausible damage, category mismatch, prompt misalignment, or trivial generation artifacts were found.

\paragraph{Quality Control.}
We apply quality control after both raw-data preprocessing and synthetic image generation. 
Details of the screening procedure are provided in Appx.~\ref{appx:quality_control}.

\subsection{Evaluation Protocol}
\label{sec:eval}
Detecting AI-generated review images in refund scenarios requires more than a single fixed evaluation setting. A reliable detector should operate across both single-image and multi-image inputs, use review text when it is available, reason over image sequences in a structured manner, remain stable under prompt phrasing changes, and, crucially, avoid misclassifying genuine user-submitted evidence as forged. We therefore evaluate each detector along five complementary dimensions: \textbf{(1) Input Modality}, \textbf{(2) Contextual Information}, \textbf{(3) Multi-Step Reasoning}, \textbf{(4) Prompt Sensitivity}, and \textbf{(5) Real Image Preservation}.

This design reflects the deployment conditions of refund-evidence verification. User reviews may contain one or multiple images, may or may not include informative text, and may be processed by downstream systems using different prompt templates or interaction settings. At the same time, the error costs are asymmetric: aggressively flagging all images can improve fake-image recall, but it also harms honest users by falsely rejecting genuine evidence. Evaluating only one setting, or reporting results only on synthetic images, would therefore obscure important failure modes. Fig.~\ref{fig:Mode_Representative} provides qualitative examples for the main settings, while the corresponding quantitative results are reported in Sec.~\ref{chap:exp_overall}. We describe each evaluation dimension below.

\paragraph{Dimension 1: Input Modality.}
Real reviews from e-commerce and service platforms often include multiple images of the same product, order, or service experience. We evaluate two input settings. In the \textbf{Single-Image} setting, each image is judged independently. In the \textbf{Multi-Image} setting, all images from the same review are provided jointly in one inference call. The multi-image setting allows the detector to use cross-image cues such as consistent lighting, viewpoint, background, object geometry, and damage appearance. To avoid a degenerate comparison, this setting is applied only to reviews containing at least two images.

\paragraph{Dimension 2: Contextual Information.}
Review text can provide evidence that changes the interpretation of an image. For example, a complaint about a broken or leaking item supplies a ready explanation for visual irregularities, which may otherwise have been flagged as generation artifacts. We compare two settings: \textbf{Review-Free}, where the detector receives only the image input, and \textbf{Review-Conditioned}, where the associated user review text is also provided. This dimension measures whether textual context improves detection quality or instead introduces misleading shortcuts.

\paragraph{Dimension 3: Multi-Step Reasoning.}
When a review contains multiple images, a detector may benefit from examining each image individually before reaching a review-level decision. We therefore introduce a \textbf{Multi-Step} setting, in which the images are presented one per turn in a multi-turn conversation: the model gives a short observation for each image and then issues a final verdict for the entire review. We compare this setting against the Multi-Image setting, where the model jointly infers a single verdict from all images in one pass. This comparison tests whether structured per-image decomposition improves reliability, particularly when different images provide complementary or conflicting evidence.

\paragraph{Dimension 4: Prompt Sensitivity.}
In practical systems, prompt templates are often revised for clarity, length, or integration with downstream pipelines. A robust detector should not change its decision solely because the same task is phrased differently. We measure prompt sensitivity by evaluating six MLLMs with five prompt variants that progressively remove task-specific guidance, on a stratified 10\% sample of the 27 Amazon categories, while holding the image input and ground-truth label fixed. Large variation across prompt styles indicates that the detector is responding to prompt wording rather than stable visual evidence.

\paragraph{Dimension 5: Real Image Preservation.}
Real image preservation evaluates whether a detector correctly retains genuine user uploads. This dimension is central to refund-evidence verification because false accusations of forgery can directly harm user experience and platform trust. Unlike the first four dimensions, which define setting variations, this dimension is evaluated across all settings using a shared real-image subset. It separates detectors that achieve high fake-image recall by over-flagging from detectors that detect synthetic evidence while preserving authentic content.
\section{Experiments}
\subsection{Experimental Setup}

\paragraph{Metrics.}
\label{sec:metrics}
We formulate the task as binary verification, where fake-damaged images are treated as the positive class, and genuine, real-damaged images are treated as the negative class. For each generator subset, we report Recall, i.e., True Positive Rate (TPR), to measure the model's ability to detect synthetic damage, using the shared real-damaged subset as the negative reference set. On the real-damaged subset, we report the True Negative Rate (TNR) to assess whether authentic damaged evidence is correctly retained rather than falsely rejected as synthetic. We further report the mean confidence on correct predictions for both generated and real-damaged images. For overall comparison, we report Balanced Accuracy and the F1 score of the fake-damaged (positive) class. All overall metrics are macro-averaged over product categories.

\paragraph{Models.} We evaluate a broad set of state-of-the-art MLLMs from major providers, including proprietary models accessed through official APIs and open-weight models, and spanning general-purpose, vision-language, and omni-modal variants. We further compare them with four specialized AI-generated image detectors: CO-SPY~\citep{cheng2025co}, ForgeLens~\citep{chen2025forgelens}, Effort~\citep{yan2025orthogonal}, and IAPL~\citep{li2026towards}, each evaluated with checkpoints trained on different datasets. The full list of evaluated MLLMs is provided in Appx.~\ref{appx:eval_full_list}.

\paragraph{Implementation Settings.}
For all MLLM settings, the model returns a binary prediction, a confidence score in $[0,1]$, and a free-text rationale. In multi-image settings, the model additionally returns a review-level verdict and per-image predictions. For proprietary models, we obtain outputs directly through their respective official API services using default inference settings unless otherwise specified. Specialized AI-generated image detectors are evaluated on a cloud server with one NVIDIA A100 80GB GPU, using officially released implementations, released checkpoints, and default inference settings. Additional details, including prompts and hyperparameter settings, are provided in Appx.~\ref{appx:specialized_detectors}. To interpret model performance, we conduct a small-scale human 
evaluation and report accuracy as a reference baseline, with details provided in Appx.~\ref{appx:human_eval}.

\subsection{Overall Results}
\label{chap:exp_overall}

Tables~\ref{tab:main} and~\ref{tab:main_results_review} present the primary evaluation results. We summarize the main findings below.

\begin{table}[t]
\centering
\caption{\textbf{Performance comparison of representative MLLMs and specialized AI-generated image detectors on \method{}.} Results are reported across six AI image editing and generation subsets under the single-image setting without review context, which evaluates the input-modality dimension. Full results for all 30 evaluated MLLMs are provided in Tab.~\ref{tab:T2}.}
\label{tab:main}
\resizebox{\textwidth}{!}{
\begin{tabular}{c|cc cc cc cc cc cc|cc|cc}
\toprule
\multirow{2}[2]{*}{\textbf{Models}}
& \multicolumn{2}{c}{\textbf{GPT Image 2}}
& \multicolumn{2}{c}{\textbf{Grok Imagine}}
& \multicolumn{2}{c}{\textbf{Nano Banana 2}}
& \multicolumn{2}{c}{\textbf{Wan2.7-Image}}
& \multicolumn{2}{c}{\textbf{Qwen-Image-2.0}}
& \multicolumn{2}{c|}{\textbf{Qwen-Image-Edit}}
& \multicolumn{2}{c|}{\textbf{Overall}}
& \multicolumn{2}{c}{\textbf{Real-Damaged}} \\
\cmidrule(lr){2-3}
\cmidrule(lr){4-5}
\cmidrule(lr){6-7}
\cmidrule(lr){8-9}
\cmidrule(lr){10-11}
\cmidrule(lr){12-13}
\cmidrule(lr){14-15}
\cmidrule(lr){16-17}
& \textbf{TPR} & \textbf{Conf.}
& \textbf{TPR} & \textbf{Conf.}
& \textbf{TPR} & \textbf{Conf.}
& \textbf{TPR} & \textbf{Conf.}
& \textbf{TPR} & \textbf{Conf.}
& \textbf{TPR} & \textbf{Conf.}
& \textbf{Bal.Acc.} & \textbf{F1}
& \textbf{TNR} & \textbf{Conf.} \\
\midrule

\rowcolor{mypurple}
\multicolumn{17}{c}{\textbf{Multimodal Large Language Models}} \\
\midrule
DeepSeek-V4.1-Flash & 0.084 & 0.693 & 0.171 & 0.701 & 0.213 & 0.690 & 0.255 & 0.700 & 0.357 & 0.710 & 0.423 & 0.721 & 0.601 & 0.391 & 0.952 & 0.764\\
Gemini 3.1 Pro & 0.091 & 0.714 & 0.122 & 0.712 & 0.167 & 0.712 & 0.260 & 0.719 & 0.327 & 0.715 & 0.519 & 0.717 & 0.619 & 0.391 & 0.990 & 0.704\\
Gemini 3.5 Flash-Lite & 0.050 & 0.965 & 0.066 & 0.958 & 0.085 & 0.964 & 0.119 & 0.968 & 0.178 & 0.965 & 0.244 & 0.968 & 0.560 & 0.212 & 0.996 & 0.950\\
Gemini 3.8 Flash & 0.025 & 0.703 & 0.013 & 0.708 & 0.066 & 0.703 & 0.102 & 0.708 & 0.157 & 0.707 & 0.199 & 0.710 & 0.547 & 0.165 & \textbf{\textcolor{mypurple2}{1.000}} & 0.658\\
GPT-6 Astra & 0.108 & 0.783 & 0.107 & 0.795 & 0.212 & 0.815 & 0.400 & 0.837 & \underline{0.453} & 0.830 & 0.530 & 0.840 & \underline{0.650} & 0.450 & \underline{0.998} & 0.930\\
GPT-6 Sol & 0.089 & 0.771 & 0.126 & 0.769 & 0.202 & 0.794 & 0.312 & 0.838 & 0.322 & 0.815 & 0.465 & 0.838 & 0.623 & 0.390 & 0.993 & 0.913\\
GPT-6 Luna & \textbf{\textcolor{mypurple2}{0.761}} & 0.896 & \textbf{\textcolor{mypurple2}{0.910}} & 0.919 & \textbf{\textcolor{mypurple2}{0.934}} & 0.924 & \textbf{\textcolor{mypurple2}{0.928}} & 0.927 & \textbf{\textcolor{mypurple2}{0.845}} & 0.902 & \textbf{\textcolor{mypurple2}{0.951}} & 0.934 & 0.547 & \textbf{\textcolor{mypurple2}{0.873}} & 0.206 & 0.763\\
Grok 4.6 & 0.050 & 0.829 & 0.047 & 0.806 & 0.131 & 0.823 & 0.173 & 0.835 & 0.198 & 0.818 & 0.250 & 0.814 & 0.568 & 0.237 & 0.995 & 0.843\\
Kimi K3 & 0.015 & 0.848 & 0.043 & 0.718 & 0.029 & 0.711 & 0.127 & 0.808 & 0.152 & 0.742 & 0.276 & 0.773 & 0.541 & 0.189 & 0.975 & 0.866\\
Qwen3-VL-Flash & 0.043 & 0.936 & 0.090 & 0.938 & 0.172 & 0.938 & 0.247 & 0.939 & 0.278 & 0.936 & 0.364 & 0.938 & 0.592 & 0.321 & 0.986 & 0.973\\
Qwen3-VL-Plus & 0.002 & 0.920 & 0.006 & 0.932 & 0.024 & 0.942 & 0.050 & 0.945 & 0.049 & 0.952 & 0.108 & 0.947 & 0.518 & 0.075 & 0.997 & 0.977\\
Qwen3.5-Omni-Flash & 0.013 & 0.911 & 0.050 & 0.910 & 0.083 & 0.910 & 0.147 & 0.910 & 0.152 & 0.911 & 0.185 & 0.911 & 0.551 & 0.183 & 0.996 & 0.952\\
Qwen3.5-Omni-Plus & 0.015 & 0.928 & 0.026 & 0.928 & 0.055 & 0.928 & 0.109 & 0.931 & 0.121 & 0.933 & 0.180 & 0.936 & 0.542 & 0.151 & \textbf{\textcolor{mypurple2}{1.000}} & 0.971\\
Qwen3.8-27B & 0.104 & 0.821 & 0.090 & 0.820 & 0.163 & 0.828 & 0.218 & 0.834 & 0.207 & 0.815 & 0.320 & 0.839 & 0.583 & 0.305 & 0.982 & 0.870\\
Qwen3.8-Flash & 0.095 & 0.896 & \underline{0.253} & 0.905 & \underline{0.295} & 0.904 & 0.412 & 0.891 & 0.390 & 0.903 & 0.561 & 0.904 & 0.641 & \underline{0.490} & 0.947 & 0.940\\
Qwen3.8-Max & \underline{0.112} & 0.789 & 0.180 & 0.806 & 0.234 & 0.811 & \underline{0.414} & 0.828 & 0.452 & 0.829 & \underline{0.619} & 0.850 & \textbf{\textcolor{mypurple2}{0.664}} & \underline{0.490} & 0.993 & 0.864\\
\midrule
\rowcolor{mypurple}
\multicolumn{17}{c}{\textbf{Specialized Detectors}} \\
\midrule
CO-SPY [ProGAN] & \textbf{\textcolor{mypurple2}{0.977}} & 0.991 & 0.099 & 0.903 & 0.312 & 0.903 & 0.218 & 0.869 & 0.894 & 0.972 & \textbf{\textcolor{mypurple2}{0.996}} & 0.997 & 0.782 & 0.733 & \textbf{\textcolor{mypurple2}{0.982}} & 0.998\\
CO-SPY [SD-v1.4] & 0.360 & 0.767 & 0.310 & 0.743 & 0.282 & 0.750 & 0.360 & 0.759 & 0.339 & 0.750 & 0.349 & 0.769 & 0.509 & 0.455 & 0.685 & 0.844\\
ForgeLens [ProGAN] & 0.700 & 0.955 & 0.217 & 0.869 & 0.516 & 0.905 & 0.823 & 0.949 & 0.859 & 0.966 & 0.925 & 0.973 & \underline{0.815} & 0.798 & 0.958 & 0.968\\
ForgeLens [GenImage] & 0.967 & 0.888 & 0.158 & 0.749 & \textbf{\textcolor{mypurple2}{1.000}} & 0.940 & \underline{0.990} & 0.895 & \textbf{\textcolor{mypurple2}{1.000}} & 0.925 & 0.989 & 0.910 & \textbf{\textcolor{mypurple2}{0.904}} & \underline{0.915} & 0.957 & 0.875\\
Effort [SD-v1.4] & 0.710 & 0.816 & 0.546 & 0.763 & 0.658 & 0.804 & 0.755 & 0.827 & 0.858 & 0.874 & 0.890 & 0.885 & 0.795 & 0.828 & 0.853 & 0.823\\
Effort [Chameleon] & 0.828 & 0.715 & \underline{0.824} & 0.716 & 0.868 & 0.730 & 0.882 & 0.756 & 0.849 & 0.734 & 0.872 & 0.736 & 0.452 & 0.838 & 0.051 & 0.575\\
IAPL [ProGAN] & 0.255 & 0.874 & 0.295 & 0.886 & 0.140 & 0.866 & 0.122 & 0.869 & 0.398 & 0.920 & 0.568 & 0.927 & 0.628 & 0.438 & \underline{0.960} & 0.984\\
IAPL [SD-v1.4] & \underline{0.973} & 0.883 & \textbf{\textcolor{mypurple2}{0.895}} & 0.855 & \underline{0.916} & 0.872 & \textbf{\textcolor{mypurple2}{0.996}} & 0.932 & \underline{0.955} & 0.882 & \underline{0.994} & 0.901 & 0.727 & \textbf{\textcolor{mypurple2}{0.932}} & 0.499 & 0.804\\

\midrule
\rowcolor{mypurple}
\multicolumn{17}{c}{\textbf{Baselines}} \\
\midrule
Random Guessing           & 0.500 & — & 0.500 & — & 0.500 & — & 0.500 & — & 0.500 & — & 0.500 &  —& 0.500 & — & 0.500 & — \\
Human Reference               & 0.420 & — & 0.509 & — & 0.539 & — & 0.598 & — & 0.697 & — & 0.709 & — & 0.686 & 0.704 & 0.793 & — \\
\bottomrule
\end{tabular}
}
\end{table}

\paragraph{MLLM Failures Are Driven by Synthetic-Damage Under-Detection.} In the single-image no-review setting, current MLLMs preserve genuine evidence more reliably than they detect synthetic damage. Across MLLMs, the average real-image TNR is 0.938, whereas the average TPR over fake-damaged subsets is only 0.230. This indicates that the dominant failure mode is not excessive rejection of genuine evidence, but systematic under-detection of synthetic refund evidence. GPT-6 Luna is the only exception, flagging most genuine evidence as fake with a TNR of 0.206. The category-level breakdown supports this interpretation: category-wise balanced accuracy is much more strongly associated with fake-damage recall than with real-image retention.

\paragraph{Failures Are Category-Dependent.}
The failure pattern is not uniform across categories. MLLMs are weakest on everyday-product and service-oriented categories such as Baby Products, Tools \& Home Improvement, and Food Delivery, which form the lower end of the category-wise breakdown. These categories often involve heterogeneous objects, unconstrained capture conditions, cluttered backgrounds, and non-standard evidence styles, making them less aligned with clean forensic authenticity cues. By contrast, categories with more regular object structure or visually salient physical defects tend to provide stronger cues for detection. This category-level variation shows that \method{} not only measures generator artifacts but also probes whether models can verify plausible damage under realistic product and service contexts. We provide detailed analysis in Appx.~\ref{app:category_difficulty}.

\begin{table*}[t]
\centering
\caption{\textbf{Performance comparison of representative MLLMs on \method{} under review-conditioned evaluation modes.} Results are reported across six 
AI image editing and generation subsets for single-image, multi-image, 
and multi-step settings with review context. A representative subset of 
six MLLMs is shown. Specialized detectors are omitted as they do not support review text, multi-image inputs, or multi-step interaction.}
\label{tab:main_results_review}
\resizebox{\textwidth}{!}{
\begin{tabular}{c|cc cc cc cc cc cc|cc|cc}
\toprule
\multirow{2}[2]{*}{\textbf{Models}}
& \multicolumn{2}{c}{\textbf{GPT Image 2}}
& \multicolumn{2}{c}{\textbf{Grok Imagine}}
& \multicolumn{2}{c}{\textbf{Nano Banana 2}}
& \multicolumn{2}{c}{\textbf{Wan2.7-Image}}
& \multicolumn{2}{c}{\textbf{Qwen-Image-2.0}}
& \multicolumn{2}{c|}{\textbf{Qwen-Image-Edit}}
& \multicolumn{2}{c|}{\textbf{Overall}}
& \multicolumn{2}{c}{\textbf{Real-Damaged}} \\
\cmidrule(lr){2-3}\cmidrule(lr){4-5}\cmidrule(lr){6-7}\cmidrule(lr){8-9}\cmidrule(lr){10-11}\cmidrule(lr){12-13}\cmidrule(lr){14-15}\cmidrule(lr){16-17}
& \textbf{TPR} & \textbf{Conf.}
& \textbf{TPR} & \textbf{Conf.}
& \textbf{TPR} & \textbf{Conf.}
& \textbf{TPR} & \textbf{Conf.}
& \textbf{TPR} & \textbf{Conf.}
& \textbf{TPR} & \textbf{Conf.}
& \textbf{Bal.Acc.} & \textbf{F1}
& \textbf{TNR} & \textbf{Conf.} \\
\midrule

\rowcolor{mypurple}
\multicolumn{17}{c}{\textbf{Single Image with Review}} \\
\midrule
DeepSeek-V4.1-Flash     & \textbf{\textcolor{mypurple2}{0.261}} & 0.727 & \textbf{\textcolor{mypurple2}{0.334}} & 0.689 & \textbf{\textcolor{mypurple2}{0.397}} & 0.689 & \textbf{\textcolor{mypurple2}{0.454}} & 0.724 & \textbf{\textcolor{mypurple2}{0.519}} & 0.720 & \underline{0.612} & 0.721 & \underline{0.658} & \textbf{\textcolor{mypurple2}{0.588}} & 0.887 & 0.782 \\
Gemini 3.8 Flash        & \underline{0.119} & 0.697 & 0.106 & 0.703 & 0.201 & 0.703 & 0.274 & 0.710 & 0.349 & 0.710 & 0.467 & 0.712 & 0.626 & 0.385 & \textbf{\textcolor{mypurple2}{1.000}} & 0.673 \\
GPT-5.6 Terra           & 0.069 & 0.780 & 0.086 & 0.801 & 0.119 & 0.807 & 0.295 & 0.870 & 0.275 & 0.843 & 0.458 & 0.864 & 0.608 & 0.346 & \underline{0.999} & 0.964 \\
Grok 4.6                & 0.037 & 0.854 & 0.041 & 0.821 & 0.110 & 0.828 & 0.189 & 0.842 & 0.200 & 0.824 & 0.273 & 0.828 & 0.570 & 0.241 & 0.998 & 0.865 \\
Kimi K3                 & 0.036 & 0.714 & 0.082 & 0.713 & 0.076 & 0.719 & 0.202 & 0.766 & 0.215 & 0.739 & 0.357 & 0.762 & 0.581 & 0.271 & \textbf{\textcolor{mypurple2}{1.000}} & 0.891 \\
Qwen3.8-Max             & 0.115 & 0.781 & \underline{0.198} & 0.798 & \underline{0.257} & 0.811 & \underline{0.414} & 0.831 & \underline{0.452} & 0.827 & \textbf{\textcolor{mypurple2}{0.638}} & 0.848 & \textbf{\textcolor{mypurple2}{0.672}} & \underline{0.502} & \underline{0.999} & 0.888 \\
\midrule
\rowcolor{mypurple}
\multicolumn{17}{c}{\textbf{Multi Image with Review}} \\
\midrule
DeepSeek-V4.1-Flash     & \textbf{\textcolor{mypurple2}{0.327}} & 0.707 & \textbf{\textcolor{mypurple2}{0.364}} & 0.735 & \textbf{\textcolor{mypurple2}{0.438}} & 0.727 & \underline{0.572} & 0.739 & \textbf{\textcolor{mypurple2}{0.633}} & 0.759 & \textbf{\textcolor{mypurple2}{0.789}} & 0.759 & \textbf{\textcolor{mypurple2}{0.748}} & \textbf{\textcolor{mypurple2}{0.671}} & \underline{0.980} & 0.803 \\
Gemini 3.8 Flash        & \underline{0.321} & 0.719 & \underline{0.278} & 0.717 & \underline{0.411} & 0.720 & 0.550 & 0.715 & \underline{0.561} & 0.718 & \underline{0.782} & 0.721 & \underline{0.742} & \underline{0.633} & \textbf{\textcolor{mypurple2}{1.000}} & 0.687 \\
GPT-5.6 Terra           & 0.152 & 0.899 & 0.170 & 0.897 & 0.276 & 0.902 & \textbf{\textcolor{mypurple2}{0.591}} & 0.944 & 0.434 & 0.929 & 0.699 & 0.945 & 0.688 & 0.545 & \textbf{\textcolor{mypurple2}{1.000}} & 0.967 \\
Grok 4.6                & 0.068 & 0.851 & 0.037 & 0.814 & 0.124 & 0.841 & 0.294 & 0.855 & 0.232 & 0.862 & 0.366 & 0.849 & 0.591 & 0.303 & \textbf{\textcolor{mypurple2}{1.000}} & 0.898 \\
Kimi K3                 & 0.094 & 0.746 & 0.154 & 0.731 & 0.222 & 0.778 & 0.372 & 0.769 & 0.361 & 0.780 & 0.582 & 0.789 & 0.648 & 0.444 & \textbf{\textcolor{mypurple2}{1.000}} & 0.900 \\
Qwen3.8-Max             & 0.108 & 0.824 & 0.180 & 0.817 & 0.238 & 0.828 & 0.472 & 0.853 & 0.521 & 0.861 & 0.709 & 0.890 & 0.689 & 0.523 & \textbf{\textcolor{mypurple2}{1.000}} & 0.915 \\
\midrule
\rowcolor{mypurple}
\multicolumn{17}{c}{\textbf{Multi Step with Review}} \\
\midrule
DeepSeek-V4.1-Flash     & \textbf{\textcolor{mypurple2}{0.513}} & 0.788 & \textbf{\textcolor{mypurple2}{0.737}} & 0.792 & \textbf{\textcolor{mypurple2}{0.773}} & 0.816 & \underline{0.751} & 0.830 & \textbf{\textcolor{mypurple2}{0.867}} & 0.835 & \textbf{\textcolor{mypurple2}{0.934}} & 0.851 & \underline{0.816} & \textbf{\textcolor{mypurple2}{0.849}} & 0.878 & 0.788 \\
Gemini 3.8 Flash        & \underline{0.436} & 0.764 & 0.400 & 0.778 & 0.544 & 0.769 & 0.694 & 0.748 & 0.733 & 0.765 & 0.869 & 0.748 & 0.806 & 0.747 & \textbf{\textcolor{mypurple2}{1.000}} & 0.869 \\
GPT-5.6 Terra           & 0.050 & 0.836 & 0.068 & 0.850 & 0.132 & 0.875 & 0.418 & 0.887 & 0.317 & 0.877 & 0.531 & 0.911 & 0.625 & 0.391 & \textbf{\textcolor{mypurple2}{1.000}} & 0.927 \\
Grok 4.6                & 0.218 & 0.805 & 0.184 & 0.797 & 0.348 & 0.814 & 0.555 & 0.856 & 0.564 & 0.826 & 0.655 & 0.842 & 0.707 & 0.579 & \underline{0.997} & 0.867 \\
Kimi K3                 & 0.177 & 0.738 & 0.240 & 0.721 & 0.322 & 0.780 & 0.493 & 0.790 & 0.509 & 0.816 & 0.795 & 0.818 & 0.713 & 0.580 & \textbf{\textcolor{mypurple2}{1.000}} & 0.902 \\
Qwen3.8-Max             & 0.389 & 0.841 & \underline{0.526} & 0.823 & \underline{0.619} & 0.850 & \textbf{\textcolor{mypurple2}{0.797}} & 0.879 & \underline{0.756} & 0.882 & \underline{0.883} & 0.905 & \textbf{\textcolor{mypurple2}{0.835}} & \underline{0.785} & \underline{0.997} & 0.903 \\
\bottomrule
\end{tabular}
}
\end{table*}

\paragraph{Damage Type Further Shapes Detection Difficulty.}
The damage-type breakdown in Tab.~\ref{tab:damage_type} shows that MLLM failures also depend on the visual form of the claimed defect. More structural or visually salient defects are generally easier to detect: averaged over MLLMs, cracked, leaking, shattered/broken, and melted samples obtain TPRs of 0.331, 0.288, 0.282, and 0.266. By contrast, packaging-damaged, torn/ripped, mouldy/spoiled, and dented evidence remains substantially harder, with average TPRs of 0.144, 0.152, 0.163, and 0.192. Packaging-damaged evidence is especially difficult, with several MLLMs close to zero TPR. This suggests that current MLLMs are more sensitive to obvious object-level breakage or fluid-like damage than to subtle deformation, packaging-level defects, or surface-level complaints. Further analysis is provided in Appx.~\ref{app:damage_type_sensitivity}.

\paragraph{Generator Effects Remain Strong Across Evaluation Settings.} Detection difficulty varies substantially across generators. 
In the single-image no-review setting, the average MLLM TPR is only 0.104 on GPT Image 2 and 0.144 on Grok Imagine Image, but rises to 0.387 on Qwen-Image-Edit-Max. This ordering remains visible under the review-conditioned single-image, multi-image, and multi-step settings, indicating that added context does not eliminate generator-dependent failure modes. This pattern is consistent with the difference between frontier text-to-image generation and image-editing pipelines: GPT Image 2 and Grok Imagine Image produce more visually integrated fake-damaged evidence, while our visual inspection suggests that Qwen-Image-Edit-Max more often leaves localized edit cues, such as sharper boundaries or material-transition artifacts.

\paragraph{Specialized Detectors Are Stronger, But Less Reliable.} Specialized AI-generated image detectors generally outperform MLLMs in the single-image no-review setting. ForgeLens, pre-trained on GenImage, achieves the highest overall balanced accuracy among the listed detectors, reaching 0.904 balanced accuracy. However, these detectors are highly checkpoint- and generator-dependent. Some methods achieve high TPR on generated images but low TNR on real evidence, indicating a tendency to over-reject authentic damaged evidence. For example, Effort pretrained on Chameleon obtains high TPR across several generator subsets, with an average fake-damage TPR of 0.854, but has very low TNR on real images at 0.051, making it unsuitable for refund verification, where falsely rejecting genuine customer evidence is costly.

\paragraph{Context Helps Only When Models Can Aggregate Evidence.} Additional context does not translate into uniform gains. Review text alone has a limited effect, with the average balanced accuracy of the six MLLMs in Tab.~\ref{tab:main_results_review} increasing only modestly from 0.591 to 0.619 in the single-image setting, and GPT-5.6 Terra even declining from 0.626 to 0.608, suggesting that models do not consistently ground customer comments in visual evidence. Multi-image evidence is more informative: all six models improve when related images from the same review are provided jointly, with Gemini 3.8 Flash rising from 0.626 to 0.742 over the single-image with-review setting. Sequential multi-step prompting further extends this gain for five of the six models, with Qwen3.8-Max improving from 0.689 to 0.835, whereas GPT-5.6 Terra falls from 0.688 to 0.625. Thus, \method{} exposes a distinction between having more context and reliably aggregating it: cross-image cues consistently help verification, but the benefit of stateful evidence integration remains model-dependent. To further probe whether review-conditioned performance reflects genuine cross-modal reasoning or weaker semantic shortcuts, we conduct an image-text mismatch experiment in Appx.~\ref{app:image_text_mismatch}.

\begin{wraptable}{r}{0.49\textwidth}
\centering
\vspace{-0.79cm}
\caption{Effect of prompt design on AI-generated evidence detection, measured by F1.}
\label{tab:prompt_ablation}
\scriptsize
\setlength{\tabcolsep}{2.5pt}
\resizebox{0.48\textwidth}{!}{%
\begin{tabular}{l cccccc}
\toprule
\textbf{Prompt} 
& \textbf{DeepSeek} 
& \textbf{Gemini} 
& \textbf{GPT} 
& \textbf{Grok} 
& \textbf{Kimi} 
& \textbf{Qwen} \\
\midrule
\rowcolor{mypurple} Baseline
& 0.402 
& \textbf{\textcolor{mypurple2}{0.150}} 
& \underline{0.400} 
& \textbf{\textcolor{mypurple2}{0.222}} 
& 0.171 
& 0.485 \\

Merged 
& \underline{0.526} 
& 0.106 
& 0.292 
& 0.107 
& \textbf{\textcolor{mypurple2}{0.263}} 
& \textbf{\textcolor{mypurple2}{0.514}} \\

NoChk. 
& 0.494 
& \underline{0.139} 
& \textbf{\textcolor{mypurple2}{0.416}} 
& \underline{0.170} 
& 0.159 
& \underline{0.489} \\

Gen. 
& 0.507 
& 0.098 
& 0.207 
& 0.095 
& 0.106 
& 0.433 \\

Min. 
& \textbf{\textcolor{mypurple2}{0.532}} 
& 0.094 
& 0.312 
& 0.119 
& \underline{0.207} 
& 0.452 \\
\bottomrule
\vspace{-0.6cm}
\end{tabular}}
\end{wraptable}

\subsection{Sensitivity to Prompt Design}
Prompting techniques are widely adopted to improve the reasoning and problem-solving capabilities of large models. We therefore examine whether prompt design can improve the visual and forensic reasoning abilities of MLLMs on \method{}. We compare five prompt variants with progressively reduced task-specific guidance: \textbf{(1) Baseline} (full forensic prompt), \textbf{(2) Merged} (system and user instructions combined), \textbf{(3) No Checklist} (artifact checklist removed), \textbf{(4) Generic} (generic assistant identity), and \textbf{(5) Minimal} (one-sentence instruction only).

As shown in Tab.~\ref{tab:prompt_ablation}, no single prompt is best for all models. The full forensic prompt ranks first or second in F1 for Gemini 3.8 Flash, Grok 4.6, and GPT-5.6 Terra, while a generic assistant identity lowers F1 for five of the six models. DeepSeek-V4.1-Flash gains F1 under Merged and Minimal only by sacrificing real-image TNR. This indicates that detailed forensic instructions can help MLLMs attend to relevant visual cues, yet prompt engineering alone remains insufficient for robust refund-evidence detection. Extended results and discussion appear in Appx.~\ref{app:prompt_sensitivity}.
\section{Conclusion}
AI-generated image editing has created a new platform-risk setting in which synthetic visuals can be tailored to concrete refund claims. To systematically study this threat, we introduced \method{}, a multimodal benchmark for detecting AI-generated fraudulent refund evidence, evaluating models along five practical dimensions. Our experiments expose a fundamental ``over-credulity vs.\ over-sensitivity'' dilemma: current MLLMs preserve genuine evidence reliably but fail to detect most synthetic damage claims, often falling below random guessing, while specialized detectors achieve stronger fake-image recall but vary widely across checkpoints, with some producing unacceptable false positive rates on authentic submissions, revealing a clear gap between generic AI-generated image detection and claim-conditioned refund-evidence verification. We hope \method{} will inspire more robust and transparent verification methods that can protect platform integrity without harming legitimate users.

\newpage
\subsection*{AI use statement}

In this work, we used generative AI tools for data curation and for producing the benchmark artifacts themselves. MLLMs are the object of evaluation in this study, and are also used for review-media screening, privacy-oriented filtering, damage-pattern mining, and prompt construction. Image editing and generation models synthesize the fake-damaged evidence, and LLMs generate the accompanying review text. Model outputs in every construction step are treated as candidates, and human annotators make the final inclusion decision. Generative AI tools were not used to originate the research problem or to design the benchmark and its evaluation protocol.

Additionally, we used generative AI tools for coding assistance, for language and structural editing of the manuscript, and for retrieving and verifying references. Generative AI tools were also used to propose initial outlines for sections of the manuscript. The authors revised all outlines and verified the edited text against the cited sources before rewriting it. All AI-assisted code was read and tested by the authors, the reported numbers were regenerated from the released evaluation harness rather than copied from model output, and every citation was checked against its primary source. We take responsibility for the final content of this work, including text, claims, and artifacts produced with the aid of generative AI.

\subsection*{Ethics statement}

The human reference in Tab.~\ref{tab:main} was produced by three evaluators. They had no role in generating or selecting the synthetic evidence and saw only the image, with labels, generator identities, model predictions, and source metadata withheld during evaluation. Under our institutional policy this evaluation did not require IRB approval, and we collected no personal data from the evaluators. 

The real evidence is derived from publicly accessible review data. We did not contact or attempt to identify the review authors, as the records were public and contacting them at scale was not feasible. Every item passed automated screening followed by human inspection, and items showing faces, license plates, addresses, or other directly identifying content were removed. The benchmark contains synthetic evidence of the kind used in the threat it studies, which could help an attacker refine such evidence. We release aggregate failure modes rather than guidance for constructing successful claims. Intended use, restrictions, and further discussion are set out in Appx.~\ref{sec:impact} and~\ref{sec:responsible_practice}.

\subsection*{Reproducibility statement}

A preliminary version of the benchmark is available at \url{https://huggingface.co/datasets/TristanYan/FraudBench}, and the evaluation code at \url{https://github.com/Tristan0318/FraudBench}. The construction pipeline and all prompts are specified in Appx.~\ref{app:construction}. Model identifiers, API endpoints, checkpoints, and the hardware and software environment are given in Appx.~\ref{appx:eval_mllm_settings}--\ref{appx:testbed}, and metric definitions in Sec.~\ref{sec:metrics} and Appx.~\ref{sec:evaluation_metrics}. Per-category, per-damage-type, per-prompt-variant, and mismatched-review results for the evaluated MLLMs are reported in Appx.~\ref{app:supp_eval}, from which every number in Tab.~\ref{tab:main}--\ref{tab:prompt_ablation} can be recomputed. Compute usage is reported in Appx.~\ref{sec:compute_reporting}.

\bibliography{iclr2027_conference}
\bibliographystyle{iclr2027_conference}

\newpage
\appendix
\appendix
\clearpage

\section{Appendix Overview}

This appendix provides supplementary material for \method{}, organized into the following sections.

\textbf{Appendix~\ref{appx:benchmark_schema}: Benchmark Schema.}
Describes the dataset structure and statistics of \method{}. 
Appendix~\ref{appx:method_overview} introduces the overall directory schema. Appendix~\ref{appx:dataset_stats} reports per-category review and image counts. Appendix~\ref{appx:image_lang_dist} covers image-count and language distribution. Appendix~\ref{appx:rating_stats} presents rating and purchase-verification statistics. Appendix~\ref{appx:qualitative_pairs} shows qualitative pairs of real-undamaged reference images and the fake-damaged evidence synthesized from them.

\textbf{Appendix~\ref{app:construction}: Benchmark Construction.} Details the full construction pipeline. Appendix~\ref{app:data_collection_details} describes data collection from Amazon, Trip.com, GrabFood, and self-collected sources. Appendix~\ref{app:preprocessing} covers data preprocessing, including MLLM-assisted image--review alignment cleaning, 
distribution-aware sampling, and human verification. Appendix~\ref{appx:synthetic_generation} documents the damage pattern mining, synthetic evidence prompt construction, and multi-model image generation pipeline. Appendix~\ref{appx:quality_control} describes the two-stage quality control procedure, and Appendix~\ref{appx:region_annotations} describes region-level annotations of the edited area.

\textbf{Appendix~\ref{appx:evaluation}: Benchmark Evaluation.}
Provides full evaluation details. Appendix~\ref{appx:eval_settings} documents implementation settings for MLLMs (Appx.~\ref{appx:eval_mllm_settings}), specialized detectors and image forgery detection and localization methods (Appx.~\ref{appx:specialized_detectors}), and human evaluations (Appx.~\ref{appx:human_eval}), together with the ablation study settings (Appx.~\ref{app:ablation_study}) and the localization evaluation (Appx.~\ref{appx:localization_settings}). Appendix~\ref{sec:evaluation_metrics} defines all reported metrics. Appendix~\ref{app:supp_eval} reports complete numerical results for the evaluated MLLMs in every setting (30 in the single-image review-free setting, 18 in the review-conditioned single-image setting, eight on all 29 categories and 18 on the 27 Amazon categories in the multi-image settings, and six in the multi-step settings). Appendix~\ref{app:supp_eval_explorations} presents supplementary analyses, including category-level detection difficulty (Appx.~\ref{app:category_difficulty}), damage-type sensitivity (Appx.~\ref{app:damage_type_sensitivity}), image-text mismatch (Appx.~\ref{app:image_text_mismatch}), prompt sensitivity (Appx.~\ref{app:prompt_sensitivity}), threshold-free evaluation (Appx.~\ref{app:threshold_free}), image forgery detection and localization baselines (Appx.~\ref{app:ifdl}), localization of the manipulated region (Appx.~\ref{app:localization}), and the complementarity of MLLMs and specialized detectors (Appx.~\ref{app:complementarity}).

\textbf{Appendix~\ref{app:positioning}: Scope of \method{} and the Role of MLLMs.}
Discusses the capabilities that \method{} measures, where current models fall short, and the value and risk of MLLM rationales.

\textbf{Appendix~\ref{app:limitation}: Limitations.} Discusses the limitations of \method{} and directions for future work.

\textbf{Appendix~\ref{sec:compute_reporting}: Compute Reporting.} Reports computational resources used for data collection, benchmark construction, and evaluation experiments.

\textbf{Appendix~\ref{sec:impact}: Impact Statement.} Discusses the broader societal impact and dual-use considerations of this work.

\textbf{Appendix~\ref{sec:responsible_practice}: Responsible Research 
Practices.} Documents data privacy measures, anonymization procedures, licensing considerations, and responsible-use constraints associated with \method{}.

\newpage
\section{Benchmark Schema}
\label{appx:benchmark_schema}

\subsection{Overview of \method{}}
\label{appx:method_overview}
\method{} is organized as a category-rooted directory tree. The top level 
contains one folder per product or service category, covering all 29 
categories: 27 Amazon product categories and two service categories 
(\textit{Hotels \& Accommodations} from Trip.com and 
\textit{Delivery, Pickup \& Dine-Out} from GrabFood). Within each 
category, three subdirectories partition the data by subset and purpose: 
\texttt{Positive}, \texttt{Negative}, and \texttt{DeepFake}. The overall structure is illustrated in 
Fig.~\ref{code:benchmark-schema}.

\begin{figure}[!htp]
\centering
\begin{minipage}{\textwidth}
\begin{nocontent}{Benchmark Directory Structure (Abbreviated)}
<Category>/
+-- Positive/                     # Real-undamaged images (synthesis sources)
|   +-- Review_XXX/
|       +-- Image_XXX_YY.jpg
|       +-- MetaReview_XXX.json
+-- Negative/                     # Real-damaged images (evaluation negatives)
|   +-- Review_XXX/
|       +-- Image_XXX_YY.jpg
|       +-- MetaReview_XXX.json
+-- DeepFake/                     # Synthetic fake-damaged images
    +-- <model-name>/             # One subfolder per generation model
    |   +-- Review_XXX/
    |       +-- Image_XXX_YY.jpg
    +-- Metadata/
       +-- Edit_XXX.json         # Generation metadata per review

\end{nocontent}
\end{minipage}
\caption{Top-level directory structure of \method{}, illustrated for a 
single product or service category. The structure is identical across 
all 29 categories.}
\label{code:benchmark-schema}
\end{figure}

\paragraph{Positive and Negative Subsets.}
The \texttt{Positive} directory contains real-undamaged review records, 
which serve as generation references for synthetic evidence and are 
excluded from evaluation. The \texttt{Negative} directory contains 
real-damaged review records, which form the negative class in the binary 
verification task. Both subsets follow the same review-level organization: 
each \texttt{Review\_XXX} folder contains one or more image files named 
\texttt{Image\_XXX\_YY.jpg} and a corresponding 
\texttt{MetaReview\_XXX.json} file. The metadata file stores the review 
identifier, rating, review text, local image filenames, product or service 
category, and enriched product or hotel metadata including titles, 
descriptions, and features, as well as the user identifier provided by the source platform and other platform-specific fields.

\paragraph{DeepFake Subset.}
The \texttt{DeepFake} directory is organized by generation model. Each of the 12 image editing and generation models has a dedicated subfolder, containing one \texttt{Review\_XXX} folder per source review with the corresponding synthesized images. Synthetic images retain the original filenames of their source images to facilitate alignment across models and with the source \texttt{Positive} records. Generation metadata is stored in the shared \texttt{Metadata} subfolder as \texttt{Edit\_XXX.json} files, one per review. Each metadata file records the selected damage type, the targeted surface or component, the full edit instruction submitted to the generation models, a short rationale, and the generated review comment used as the fake review text in review-conditioned evaluation settings.

\subsection{Dataset Statistics}
\label{appx:dataset_stats}

\method{} contains \textbf{822} real review samples and \textbf{13,856}
images across 29 product and service categories. The real-image pool
consists of 988 real-undamaged images from 422 positive reviews and
1,012 real-damaged images from 400 negative reviews. Each of the 988
real-undamaged images is processed by all 12 generation models,
yielding 11,856 synthetic fake-damaged images in total.
Table~\ref{tab:category_distribution} reports the per-category review
and image counts.

\begin{table}[!htp]
\centering
\caption{\textbf{Per-category review and image counts in \method{}.}
Pos.\ and Neg.\ denote positive (real-undamaged) and negative
(real-damaged) subsets respectively. DeepFake counts equal 12 times the corresponding Pos.\ image count, as each real-undamaged image is processed by all 12 generation models.}
\label{tab:category_distribution}
\resizebox{\textwidth}{!}{%
\begin{tabular}{l cc cc c}
\toprule
\textbf{Category}
& \textbf{Pos.\ Reviews} & \textbf{Neg.\ Reviews}
& \textbf{Pos.\ Images} & \textbf{Neg.\ Images}
& \textbf{DeepFake Images} \\
\midrule
All Beauty                     &  11 &  11 &  28 &  34 &   336 \\
Amazon Fashion                 &  11 &   9 &  35 &  17 &   420 \\
Appliances                     &  12 &  14 &  26 &  39 &   312 \\
Arts, Crafts \& Sewing         &   7 &  12 &  20 &  31 &   240 \\
Automotive                     &   8 &   9 &  29 &  30 &   348 \\
Baby Products                  &  11 &  17 &  31 &  56 &   372 \\
Beauty \& Personal Care        &  14 &   9 &  40 &  20 &   480 \\
Books                          &  10 &  13 &  15 &  33 &   180 \\
CDs \& Vinyl                   &  10 &  15 &  19 &  44 &   228 \\
Cell Phones \& Accessories     &  17 &  15 &  46 &  42 &   552 \\
Clothing, Shoes \& Jewelry     &  13 &  11 &  47 &  23 &   564 \\
Delivery, Pickup \& Dine-Out   &  78 &  63 & 100 & 100 &  1,200 \\
Electronics                    &  19 &  15 &  52 &  38 &   624 \\
Grocery \& Gourmet Food        &  15 &  20 &  34 &  65 &   408 \\
Handmade Products              &  16 &   8 &  50 &  36 &   600 \\
Health \& Household            &  15 &   6 &  35 &  15 &   420 \\
Health \& Personal Care        &  19 &   6 &  49 &  17 &   588 \\
Home \& Kitchen                &  11 &  18 &  29 &  48 &   348 \\
Hotels \& Accommodations       &  45 &  46 &  90 & 110 &  1,080 \\
Industrial \& Scientific       &  12 &  13 &  24 &  30 &   288 \\
Magazine Subscriptions         &   5 &   2 &   7 &   2 &    84 \\
Musical Instruments            &   7 &  10 &  16 &  25 &   192 \\
Office Products                &  11 &  11 &  37 &  27 &   444 \\
Patio, Lawn \& Garden          &   6 &   7 &  19 &  25 &   228 \\
Pet Supplies                   &   3 &   5 &  12 &   8 &   144 \\
Sports \& Outdoors             &  17 &   8 &  40 &  24 &   480 \\
Tools \& Home Improvement      &   4 &   9 &  13 &  27 &   156 \\
Toys \& Games                  &   6 &  10 &  16 &  28 &   192 \\
Video Games                    &   9 &   8 &  29 &  18 &   348 \\
\midrule
\textbf{Total}
& \textbf{422} & \textbf{400}
& \textbf{988} & \textbf{1,012} & \textbf{11,856} \\
\bottomrule
\end{tabular}}
\end{table}

\subsection{Image-Count and Language Distribution}
\label{appx:image_lang_dist}

Table~\ref{tab:image_count_dist} reports the distribution of reviews
by the number of attached images. Over 70\% of reviews contain one to
three images, while reviews with seven or more images account for fewer
than 4\%, reflecting the long-tail image-count structure typical of
real platform submissions. This distribution motivates the single-image
and multi-image evaluation dimensions in \method{}, as most reviews
provide limited visual context while a non-trivial minority supply
richer cross-image evidence.

\begin{table}[!htp]
\centering
\caption{\textbf{Review distribution by image-count bin.}
Counts are reported separately for positive and negative subsets.}
\label{tab:image_count_dist}
\begin{tabular}{c cc cc}
\toprule
\textbf{Image-Count Bin}
& \textbf{Pos.\ Reviews} & \textbf{Neg.\ Reviews}
& \textbf{Total} & \textbf{Proportion} \\
\midrule
1     & 183 & 166 & 349 & 42.5\% \\
2--3  & 137 & 120 & 257 & 31.3\% \\
4--6  &  90 &  95 & 185 & 22.5\% \\
7--10 &  12 &  19 &  31 &  3.8\% \\
\midrule
\textbf{Total}
& \textbf{422} & \textbf{400} & \textbf{822} & 100\% \\
\bottomrule
\end{tabular}
\end{table}

Table~\ref{tab:language_dist} reports the review language distribution.
English dominates at 88.9\%, consistent with the primary Amazon
e-commerce source. Chinese-language reviews account for 6.9\% of the
corpus: Simplified Chinese reviews derive from the GrabFood
food-delivery source in Singapore, while Traditional Chinese reviews
originate predominantly from Trip.com hotel submissions. A small
proportion of reviews are multilingual, with label counts summing to
837 across 822 reviews. \method{} retains multilingual text as
collected rather than normalizing to English, enabling robustness
evaluation under realistic multilingual user contexts.

\begin{table}[!htp]
\centering
\caption{\textbf{Review language distribution.}
Label counts exceed the total review count because a small number of
reviews contain text in more than one language.}
\label{tab:language_dist}
\begin{tabular}{c cc}
\toprule
\textbf{Language} & \textbf{Label Count} & \textbf{Proportion} \\
\midrule
English             & 744 & 88.9\% \\
Simplified Chinese  &  27 &  3.2\% \\
Traditional Chinese &  31 &  3.7\% \\
Other               &  35 &  4.2\% \\
\midrule
\textbf{Total label counts} & \textbf{837} & --- \\
\bottomrule
\end{tabular}
\end{table}

\subsection{Rating and Verification Statistics}
\label{appx:rating_stats}

Table~\ref{tab:rating_dist} reports the rating distribution across all
822 reviews. The distribution is strongly bimodal: 1-star reviews
account for 45.3\% of the corpus and 5-star reviews for 49.8\%, while
3-star neutral reviews are entirely absent. This U-shaped pattern
reflects the rating-derived subset assignment used during data
collection, where high-star reviews are grouped as positive
(real-undamaged) evidence and low-star reviews as negative
(real-damaged) evidence.

\begin{table}[!htp]
\centering
\caption{\textbf{Rating distribution across all 822 reviews.}
The bimodal concentration at 1-star and 5-star reflects the
rating-derived positive/negative subset assignment.}
\label{tab:rating_dist}
\begin{tabular}{c cc}
\toprule
\textbf{Star Rating} & \textbf{Count} & \textbf{Proportion} \\
\midrule
1 & 372 & 45.3\% \\
2 &  28 &  3.4\% \\
3 &   0 &  0.0\% \\
4 &  13 &  1.6\% \\
5 & 409 & 49.8\% \\
\midrule
\textbf{Total} & \textbf{822} & 100\% \\
\bottomrule
\end{tabular}
\end{table}

Among the 590 Amazon reviews, 390 (66.1\%) carry a verified-purchase
label and 200 (33.9\%) do not. Verified-purchase status is preserved
in the review metadata but is not used as an evaluation signal in
\method{}, as the benchmark focuses on visual and multimodal evidence
verification rather than purchase authenticity.

\subsection{Source--Synthesis Pairs}
\label{appx:qualitative_pairs}

\begin{figure}[!htp]
\centering
\includegraphics[width=\textwidth]{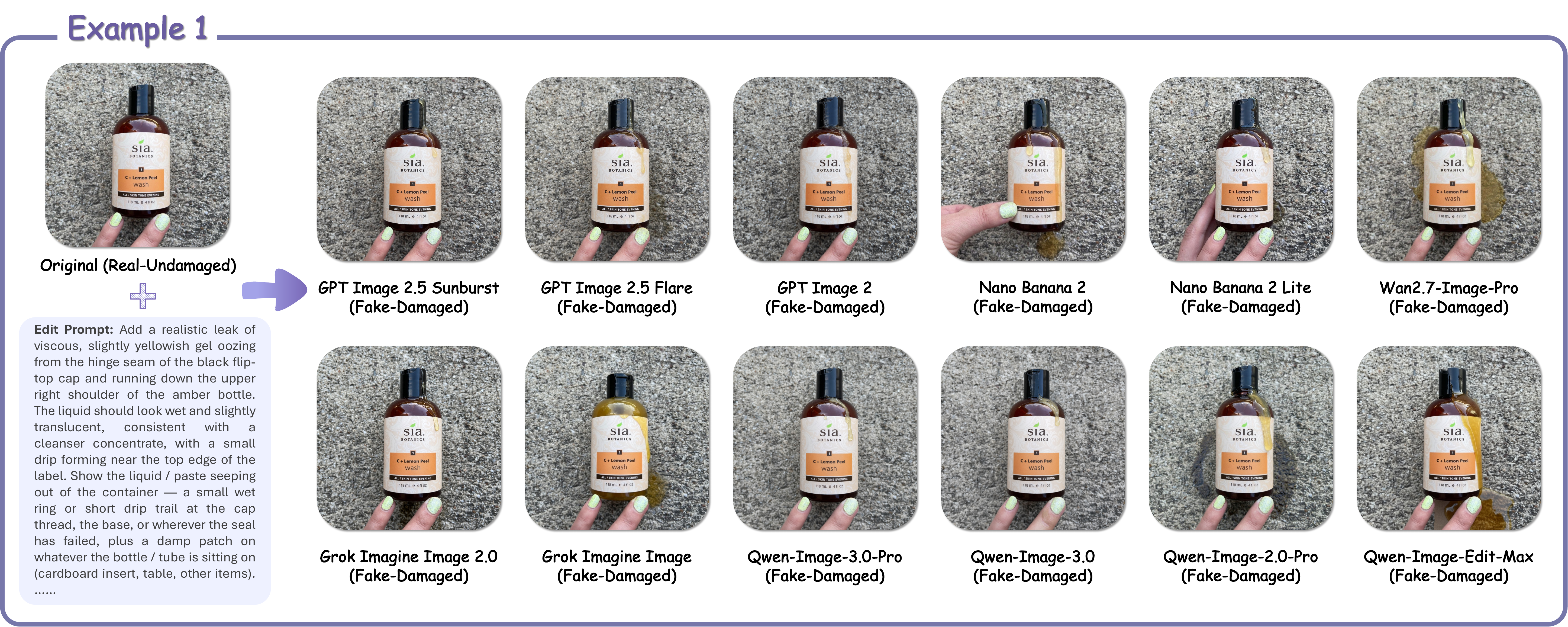}
\caption{\textbf{Source--synthesis example 1: All Beauty (leaking).} A leak at the flip-top cap and upper shoulder of a cleanser bottle.}
\label{fig:source_synthesis_pairs}
\end{figure}

\begin{figure}[!htp]
\centering
\includegraphics[width=\textwidth]{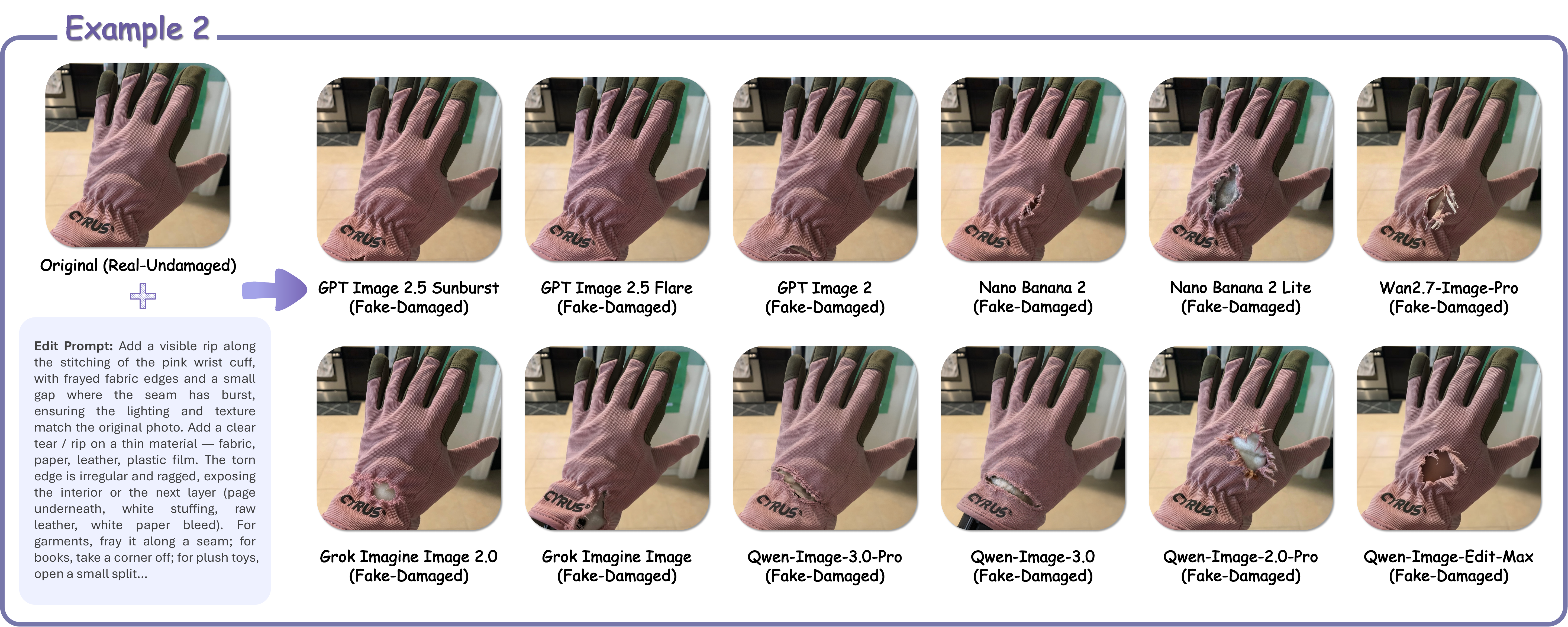}
\caption{\textbf{Source--synthesis example 2: Amazon Fashion (torn/ripped).} A tear along the wrist cuff seam of a gardening glove.}
\label{fig:source_synthesis_pairs_2}
\end{figure}

\begin{figure}[!htp]
\centering
\includegraphics[width=\textwidth]{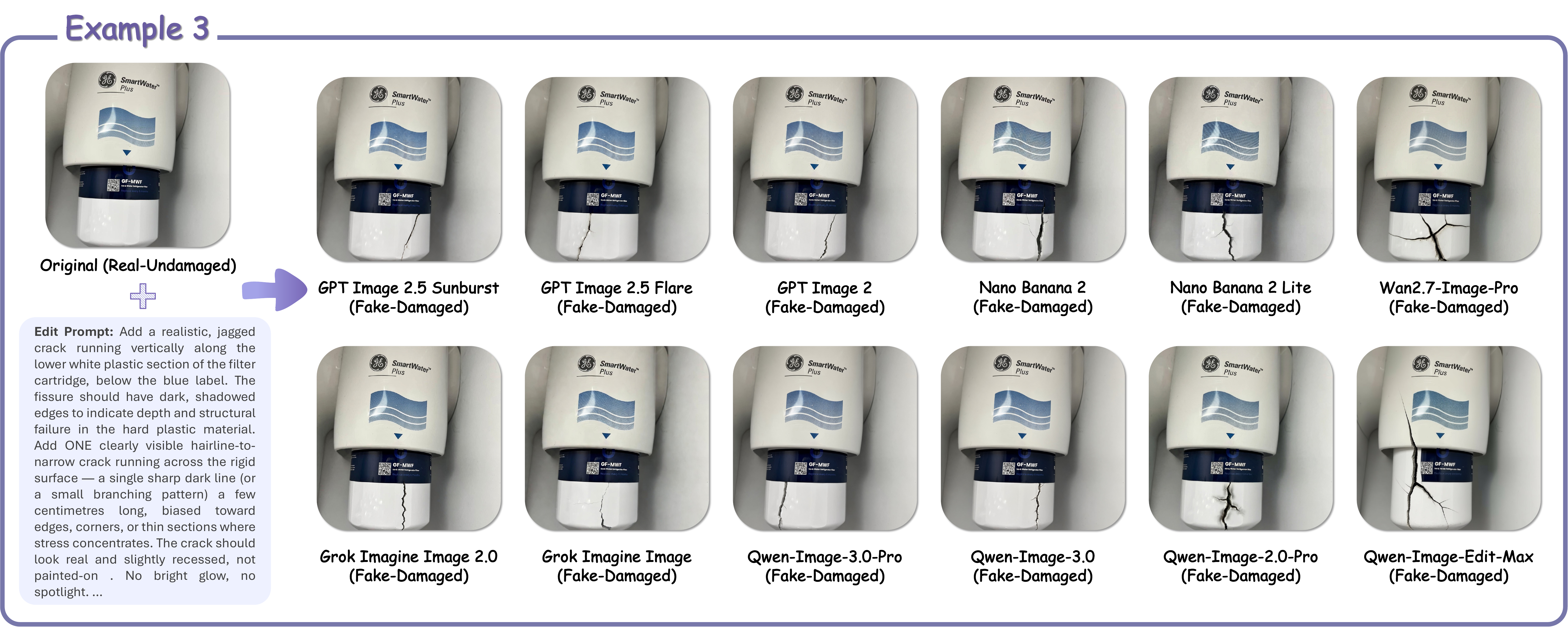}
\caption{\textbf{Source--synthesis example 3: Appliances (cracked).} A crack in the plastic body of a refrigerator water filter cartridge.}
\label{fig:source_synthesis_pairs_3}
\end{figure}

\begin{figure}[!htp]
\centering
\includegraphics[width=\textwidth]{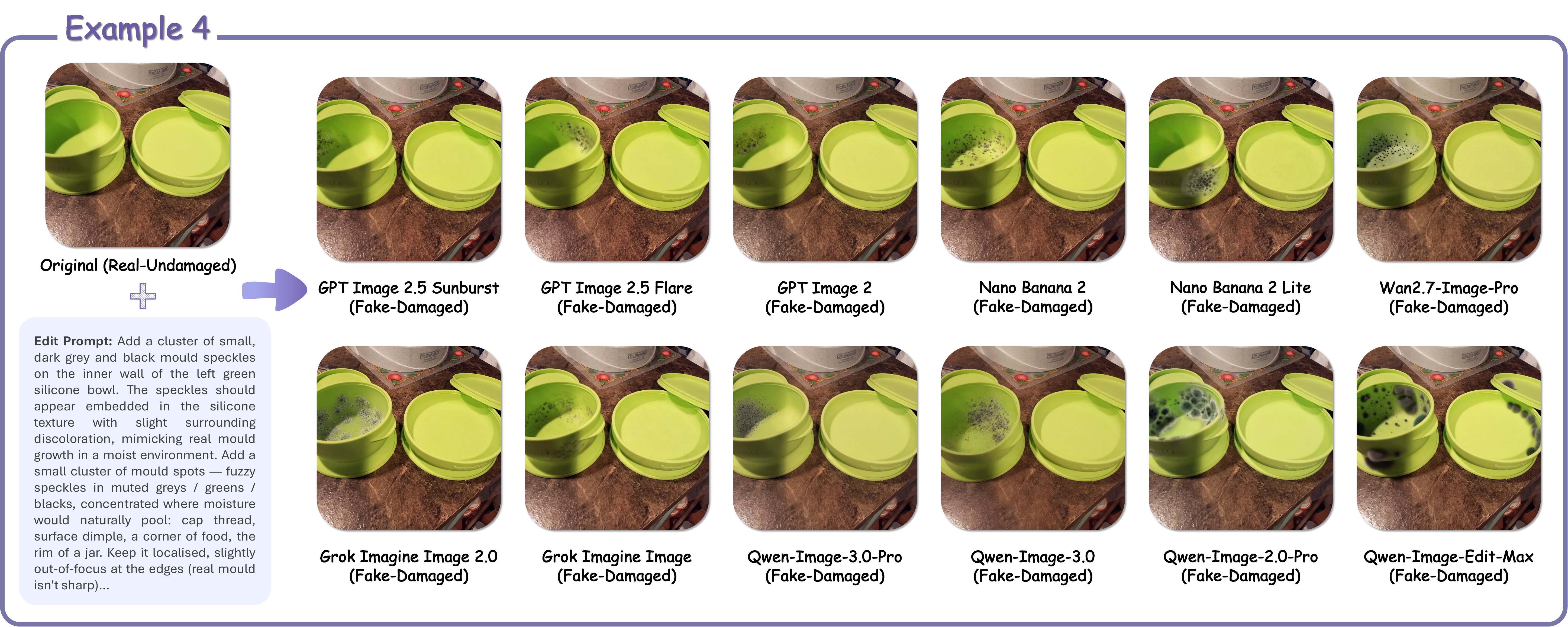}
\caption{\textbf{Source--synthesis example 4: Baby Products (mouldy/spoiled).} Mould on the inner surface of a silicone baby bowl.}
\label{fig:source_synthesis_pairs_4}
\end{figure}

\begin{figure}[!htp]
\centering
\includegraphics[width=\textwidth]{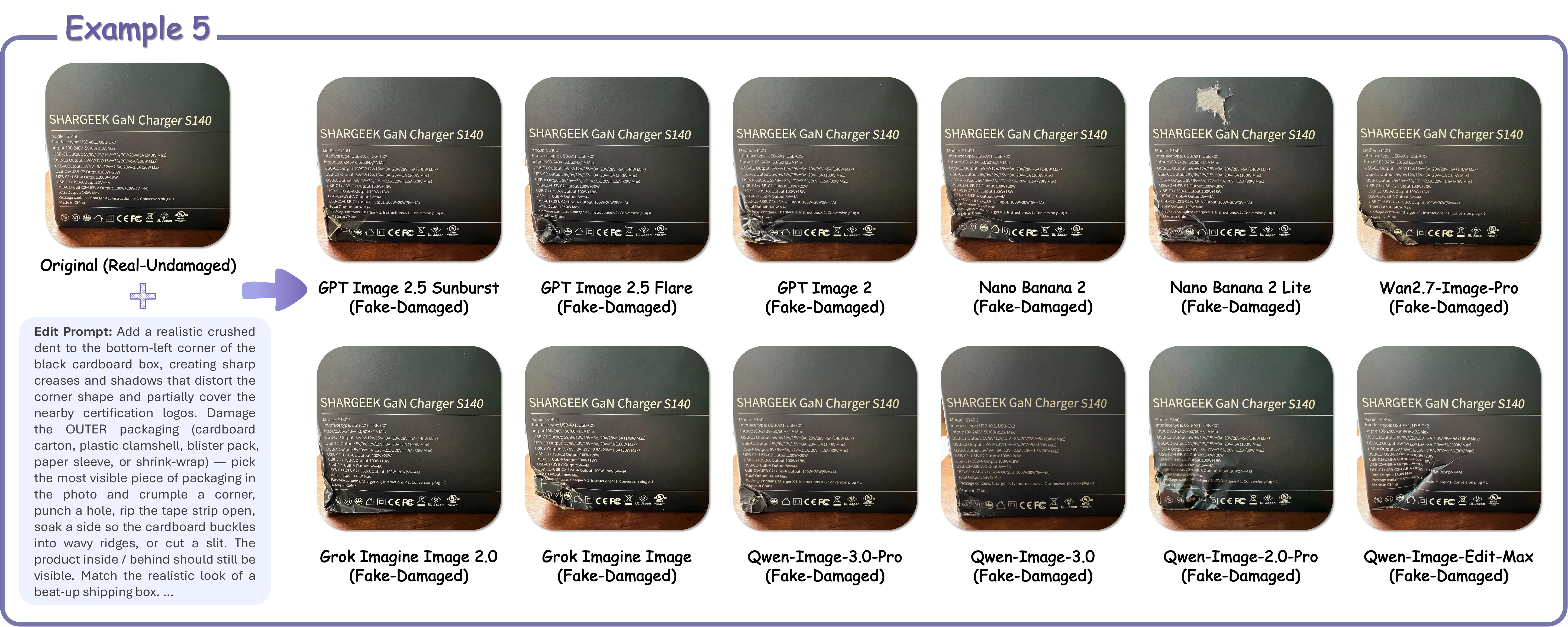}
\caption{\textbf{Source--synthesis example 5: Cell Phones \& Accessories (packaging damaged).} Damage to a corner of the retail box of a charger.}
\label{fig:source_synthesis_pairs_5}
\end{figure}

\begin{figure}[!htp]
\centering
\includegraphics[width=\textwidth]{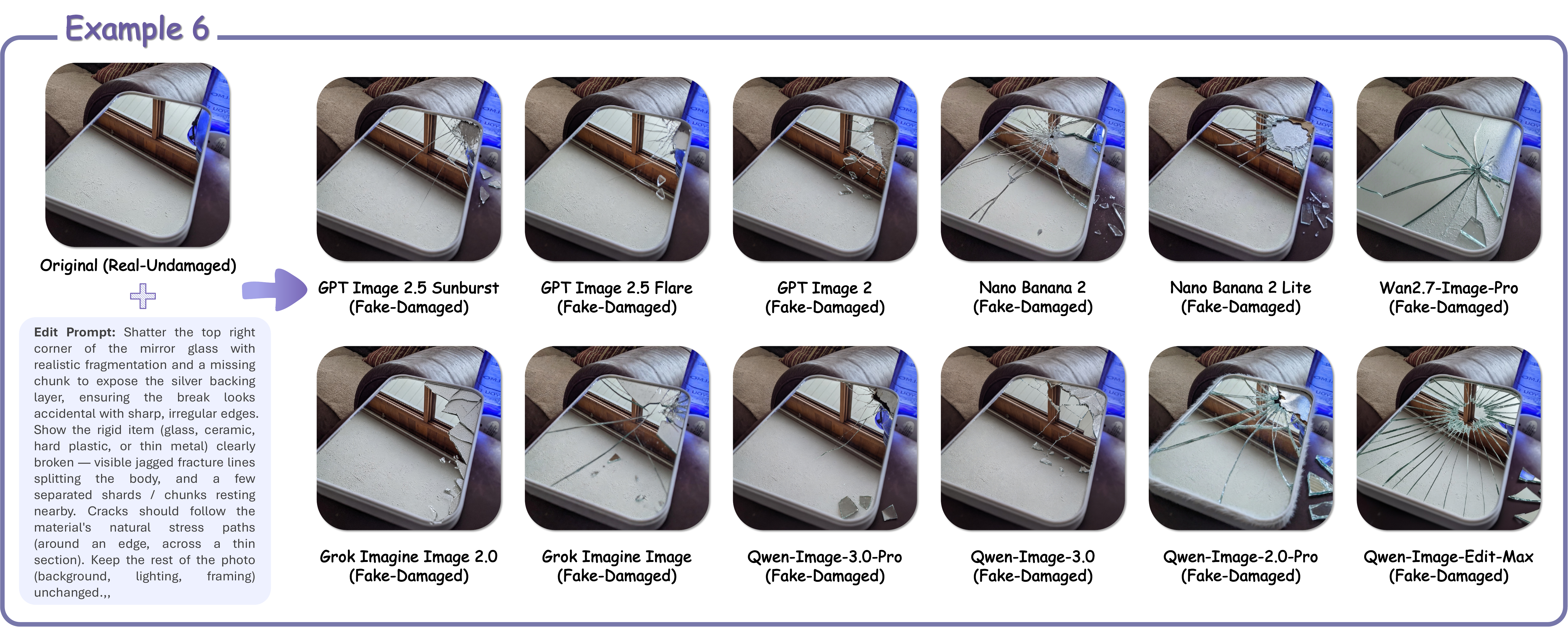}
\caption{\textbf{Source--synthesis example 6: Beauty \& Personal Care (shattered/broken).} A shattered corner of the glass of a makeup mirror.}
\label{fig:source_synthesis_pairs_6}
\end{figure}

\begin{figure}[!htp]
\centering
\includegraphics[width=\textwidth]{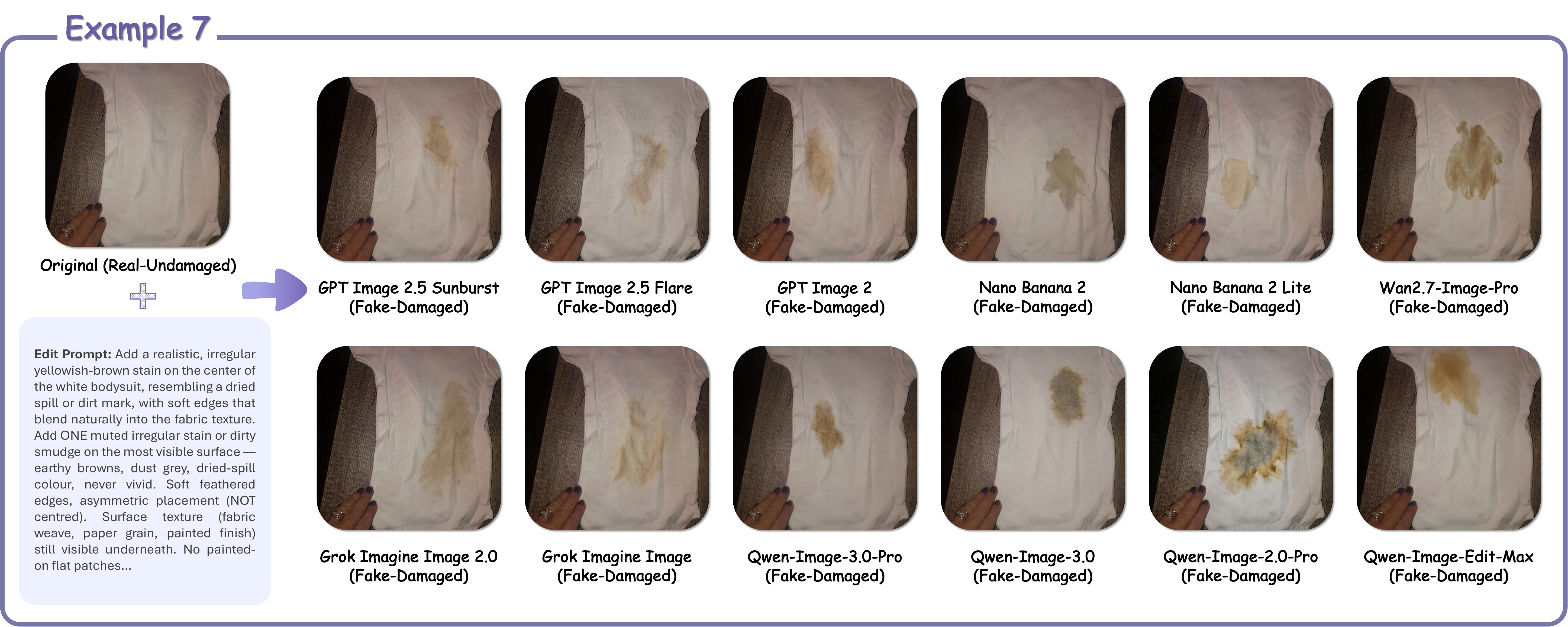}
\caption{\textbf{Source--synthesis example 7: Clothing, Shoes \& Jewelry (stained/dirty).} A stain on the front of a white infant bodysuit.}
\label{fig:source_synthesis_pairs_7}
\end{figure}

\begin{figure}[!htp]
\centering
\includegraphics[width=\textwidth]{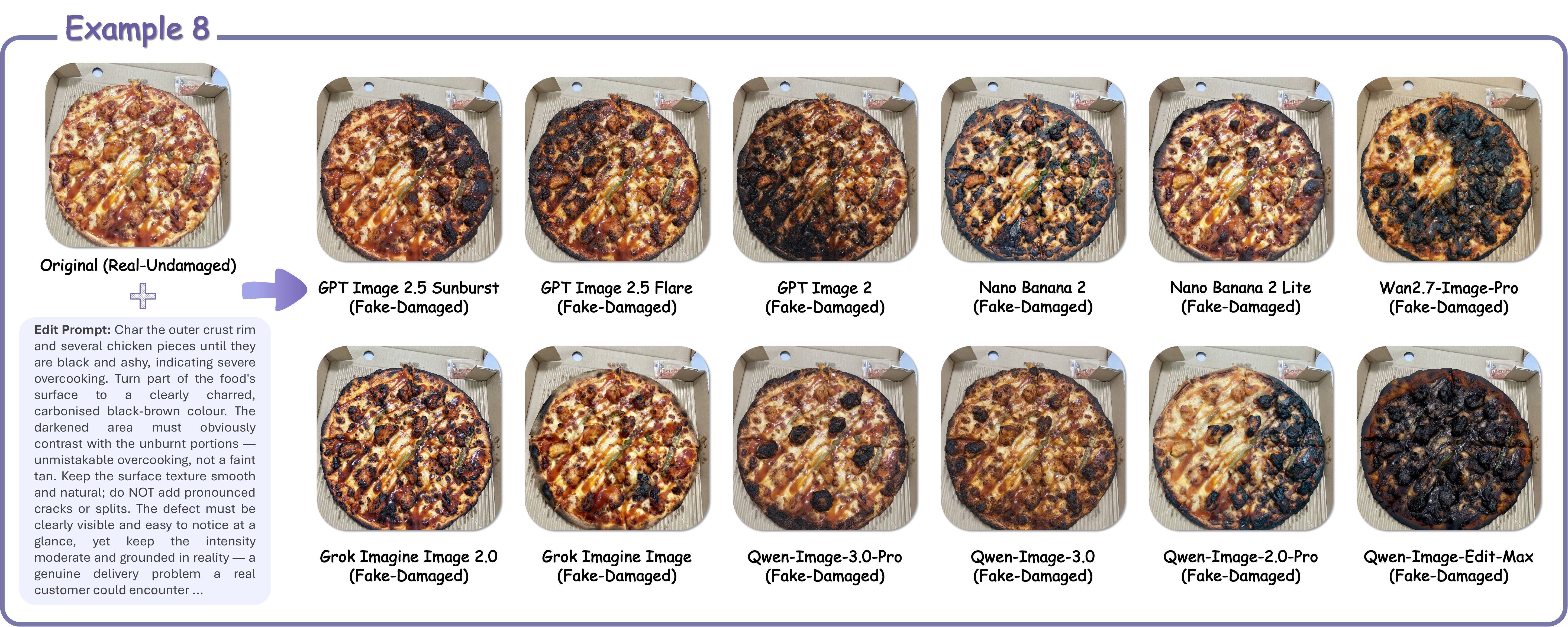}
\caption{\textbf{Source--synthesis example 8: Delivery, Pickup \& Dine-Out (burnt).} A burnt chicken pizza.}
\label{fig:source_synthesis_pairs_8}
\end{figure}

\begin{figure}[!htp]
\centering
\includegraphics[width=\textwidth]{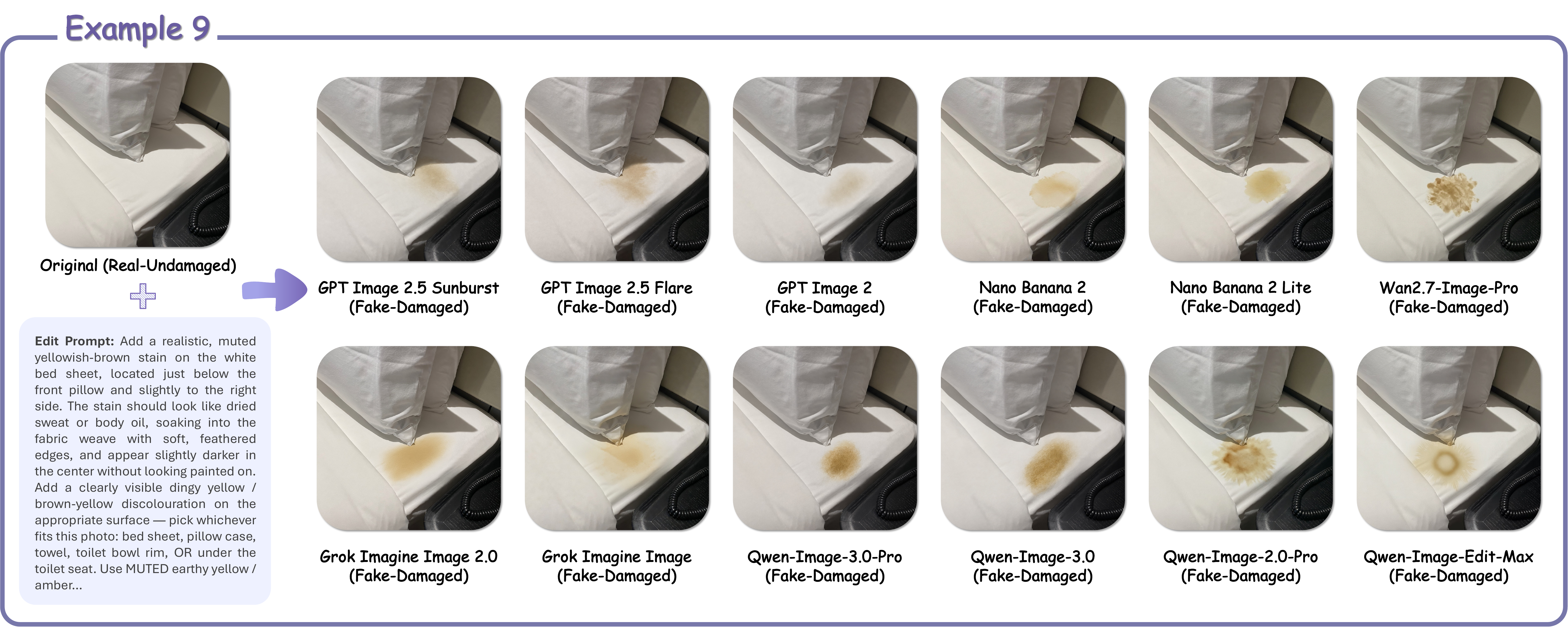}
\caption{\textbf{Source--synthesis example 9: Hotels \& Accommodations (yellow stains).} Yellow stains on a white bed sheet.}
\label{fig:source_synthesis_pairs_last}
\end{figure}

Figures~\ref{fig:source_synthesis_pairs}--\ref{fig:source_synthesis_pairs_last} show examples of source--synthesis pairs. Each example shows a real-undamaged reference image, the edit prompt constructed for it, and the fake-damaged evidence synthesized from it by each of the 12 image editing and generation models, and the caption names the product or service category, the damage type, and the edited object. Edit prompts are truncated for brevity. In these examples, the selected damage is localized to the target object, while the original product identity, viewpoint, lighting, and background are preserved.

\newpage
\section{Benchmark Construction}
\label{app:construction}

\subsection{Data Collection}
\label{app:data_collection_details}

\subsubsection{Amazon}
For the e-commerce source, we use the User Reviews and Item Metadata files from Amazon Reviews 2023~\citep{hou2024bridging}, a large-scale corpus containing user reviews, item metadata, and user--item links. The User Reviews files provide ratings, review text, helpfulness votes, timestamps, user and item identifiers, verified-purchase labels, and attached review images, while the Item Metadata files provide product titles, category hierarchies, descriptions, prices, store information, product images, and other product-level details. Representative records are shown in Fig.~\ref{code:Amazon_User_Review_Record} and~\ref{code:Amazon_Item_Metadata_Record}.

To collect the raw data, we parse the User Reviews JSONL files and retain records with image attachments. For each attached image, we extract the \texttt{large\_image\_url} field and download the corresponding high-resolution image. We keep only images satisfying a minimum resolution requirement of $1024 \times 1024$ pixels, since low-resolution images are less suitable for fine-grained evidence verification and subsequent synthetic damage generation.

For each retained review, we preserve the original review metadata, including rating, title, review text, timestamp, helpfulness votes, verified-purchase status, \texttt{asin}, \texttt{parent\_asin}, and anonymized user identifier. We then enrich each review by matching \texttt{asin} and \texttt{parent\_asin} to the Item Metadata files. The enriched metadata includes the product title, main category, category hierarchy, description, features, price, store information, product images, and available product details. This enrichment provides product-level context for later image--review alignment filtering and image-specific fake-damage prompt construction.

\begin{figure}[!htp]
\centering
\begin{minipage}{\textwidth}
\begin{example}{Amazon User Review Record}
{
  "sort_timestamp": 1634275259292,
  "rating": 3.0,
  "helpful_votes": 0,
  "title": "Meh",
  "text": "These were lightweight and soft but much too small for my liking. I would have preferred two of these together to make one loc. For that reason I will not be repurchasing.",
  "images": [
    {
      "small_image_url": "https://m.media-amazon.com/images/I/81FN4c0VHzL._SL256_.jpg",
      "medium_image_url": "https://m.media-amazon.com/images/I/81FN4c0VHzL._SL800_.jpg",
      "large_image_url": "https://m.media-amazon.com/images/I/81FN4c0VHzL._SL1600_.jpg",
      "attachment_type": "IMAGE"
    }
  ],
  "asin": "B088SZDGXG",
  "verified_purchase": true,
  "parent_asin": "B08BBQ29N5",
  "user_id": "AEYORY2AVPMCPDV57CE337YU5LXA"
}
\end{example}
\end{minipage}
\caption{Structure of an Amazon user review record.}
\label{code:Amazon_User_Review_Record}
\end{figure}

After downloading and enriching the review records, we organize the retained Amazon data by category and rating-derived subset. Each product category is stored as a top-level directory. Within each category, reviews are divided into \texttt{Positive} and \texttt{Negative} subsets. The \texttt{Positive} subset contains high-rating reviews with no apparent damage evidence, corresponding to genuine undamaged evidence. The \texttt{Negative} subset contains low-rating reviews with damage- or complaint-related evidence, corresponding to genuine damaged or problematic evidence. The rating is used only as an initial grouping signal; subsequent textual and visual filtering with human verification remove ambiguous or misaligned cases.

We then normalize the storage format at the review level. Each retained review is assigned a sequential identifier and stored as \texttt{Review\_ID} under its corresponding category and subset. The associated review images are renamed as \texttt{Image\_ID\_SUBID.jpg}, where \texttt{ID} denotes the review index and \texttt{SUBID} denotes the image index within that review. The enriched metadata are saved as \texttt{MetaReview\_ID.json}, and the local image filenames are written back into the JSON file under the \texttt{image\_files} field. The resulting structure is illustrated in Fig.~\ref{code:enriched-review-level-record}.

\begin{figure}[!htp]
\centering
\begin{minipage}{\textwidth}
\begin{example}{Amazon Item Metadata Record}
{
  "main_category": "All Beauty",
  "title": "Lurrose 100Pcs Full Cover Fake Toenails Artificial Transparent Nail Tips Nail Art for DIY",
  "average_rating": 3.7,
  "rating_number": 35,
  "features": [
    "The false toenails are durable with perfect length. You have the option to wear them long or clip them short, easy to trim and file them to in any length and shape you like.",
    "ABS is kind of green enviromental material, and makes the nails durable, breathable, light even no pressure on your own nails.",
    ......
  ],
  "description": [
    "Description",
    "The false toenails are durable with perfect length. You have the option to wear them long or clip them short, easy to trim and file them to in any length and shape you like. Plus, ABS is kind of green enviromental material, and makes the nails durable, breathable, light even no pressure on your own toenails. Fit well to your natural toenails. Non toxic, no smell, no harm to your health.",
    "Feature",
    "- Color: As Shown.- Material: ABS.- Size: 14.3 x 7.2 x 1cm.",
    "Package Including",
    "100 x Pieces fake toenails"
  ],
  "price": 6.99,
  "images": [
    {
      "hi_res": "https://m.media-amazon.com/images/I/41a1Sj7Q20L._SL1005_.jpg",
      "thumb": "https://m.media-amazon.com/images/I/31dlCd7tHSL._SS40_.jpg",
      "large": "https://m.media-amazon.com/images/I/31dlCd7tHSL.jpg",
      "variant": "MAIN"
    },
    ......
  ],
  "videos": [],
  "bought_together": null,
  "store": "Lurrose",
  "categories": [],
  "details": {
    "Color": "As Shown",
    "Size": "Large",
    "Material": "Acrylonitrile Butadiene Styrene (ABS)",
    "Brand": "Lurrose",
    "Style": "French",
    "Product Dimensions": "5.63 x 2.83 x 0.39 inches; 1.9 Ounces",
    "UPC": "799768026253",
    "Manufacturer": "Lurrose"
  },
  "parent_asin": "B07G9GWFSM"
}
\end{example}
\end{minipage}
\caption{Structure of an Amazon item metadata record.}
\label{code:Amazon_Item_Metadata_Record}
\end{figure}

\begin{figure}[!htp]
\centering
\begin{minipage}{\textwidth}
\begin{example}{Amazon Enriched Review-Level Record}
{
  "rating": 1.0,
  "title": "Damaged",
  "text": "Product arrived shattered",
  "images": [
    {
      "small_image_url": "https://m.media-amazon.com/images/I/71+Jx6Xx2GL._SL256_.jpg",
      "medium_image_url": "https://m.media-amazon.com/images/I/71+Jx6Xx2GL._SL800_.jpg",
      "large_image_url": "https://m.media-amazon.com/images/I/71+Jx6Xx2GL._SL1600_.jpg",
      "attachment_type": "IMAGE"
    }
  ],
  "asin": "B07QLN67ZB",
  "parent_asin": "B07QLN67ZB",
  "user_id": "AEYT2WPKKFXMRKEBZ27KV445J5ZA",
  "timestamp": 1632690103586,
  "helpful_vote": 0,
  "verified_purchase": true,
  "review_id": "1632690103586_AEYT2WPKKFXMRKEBZ27KV445J5ZA",
  "date": "2021-09-27",
  "image_files": [
    "Image_001_01.jpg"
  ],
  "product_meta": {
    "main_category": "All Beauty",
    "title": "Bzbuy 4 Nail Art Acrylic Liquid Powder Dappen Dish Glass Crystal Cup Glassware Tools",
    "average_rating": 4.2,
    "rating_number": 217,
    "features": [],
    "description": [
      "Bzbuy 4 Pieces Nail Art Acrylic Liquid Powder Dappen Dish Glass Crystal Cup Glassware Tools"
    ],
    "price": 6.0,
    "images": [
      {
        "thumb": "https://m.media-amazon.com/images/I/41iGFLpfJeL._SS40_.jpg",
        "large": "https://m.media-amazon.com/images/I/41iGFLpfJeL.jpg",
        "variant": "MAIN",
        "hi_res": "https://m.media-amazon.com/images/I/81TaeLP9CfL._SL1500_.jpg"
      },
      ......
    ],
    "videos": [],
    "store": "Bzbuy",
    "categories": [],
    "details": {
      "Product Dimensions": "0.89 x 0.55 x 0.55 inches; 2.88 Ounces",
      "UPC": "619317532250"
    },
    "parent_asin": "B07QLN67ZB",
    "bought_together": null
  }
}
\end{example}
\end{minipage}
\caption{Structure of an enriched Amazon review-level metadata record after matching a retained user review with its corresponding item metadata.}
\label{code:enriched-review-level-record}
\end{figure}

\subsubsection{Trip.com}
For the travel and hospitality source, we collect hotel review data from Trip.com, a large-scale online travel platform that provides hotel listings, hotel-level information, guest ratings, textual reviews, and user-uploaded review media. We use Trip.com as a complementary non-e-commerce source because hotel reviews contain multimodal evidence about room conditions, facilities, cleanliness, service quality, and other stay-related experiences. These reviews include both normal-condition evidence and genuine problematic evidence, making them suitable for constructing fraudulent refund or compensation evidence scenarios in the travel domain.

To collect hotel candidates, we define target cities across different geographic regions. In our implementation, the selected cities include Tokyo, Seoul, Paris, London, Cairo, Rio de Janeiro, and Melbourne, which \textit{\textbf{increase geographic and cultural diversity}}. For each city, we use the corresponding Trip.com city hotel-list page, where the URL is constructed from the city slug and city identifier. We parse the listing page to extract hotel identifiers, hotel names, hotel URLs, representative hotel images, rating values, review counts, and other available listing-level fields. Hotels are retained according to the default recommended order returned by Trip.com, and the top-ranked hotels in each city are selected as candidate hotels. This preserves the platform's ranking signal while yielding a diverse set of hotel properties.

For each selected hotel, we collect guest reviews by programmatically querying the Trip.com hotel comment interface. Each request specifies the hotel identifier, page index, page size, locale, currency, check-in date, check-out date, and review filter options. The interface returns structured review records, including review text, rating information, user-related fields, hotel reply information, and attached image or video URLs. Since our benchmark focuses on multimodal evidence rather than text-only complaints, we retain only reviews containing at least one image or video attachment. We use the image-video review filter to prioritize reviews with visual evidence, and for negative reviews we additionally apply the negative-review filter provided by the interface. Pagination continues until the target number of reviews is reached or the maximum page limit is exhausted.

We organize the retained Trip.com reviews into two rating-derived subsets. The \texttt{positive} subset contains 5-star reviews and represents normal or satisfactory hotel conditions. The \texttt{negative} subset contains low-rating reviews, with ratings from 1.0 to 2.0, and represents genuine problematic hotel evidence, such as dirty rooms, broken facilities, poor maintenance, pests, stains, mold, plumbing issues, or unclean bathrooms. Ratings are used only as an initial grouping signal; subsequent image--review relevance filtering and manual inspection remove irrelevant, ambiguous, or visually uninformative cases.

For each retained review, we preserve the available review-level metadata, including the review identifier, hotel identifier, hotel name, city, review URL, rating, star value, username, user avatar URL, travel type, room type, review text, creation date, check-in date, useful-vote count, hotel reply, and the original URLs of attached images and videos. Fig.~\ref{code:Tripcom_Review_Record} shows a representative retained Trip.com review-level record after metadata parsing and media URL extraction.

For hotel-level context, we also preserve the corresponding hotel metadata including the hotel identifier, hotel name, hotel URL, rating value, review count, representative image, city, and the source field indicating how the hotel was discovered. For each attached media item, we download the corresponding image or video into the review folder. Since Trip.com image URLs may contain thumbnail-resolution suffixes, we normalize image URLs where possible before downloading the media. Each downloaded file is assigned a deterministic local filename based on the original URL and written back into the review metadata. This keeps each review folder self-contained and aligns the review text, rating information, hotel context, and visual evidence for later image--text consistency checking, fake-evidence prompt construction, and benchmark evaluation.

Finally, we normalize the Trip.com storage format at the city, hotel, and review levels. Each city is stored as a top-level directory, and each selected hotel is stored under a directory named by its hotel identifier and sanitized hotel name. Hotel-level metadata are saved as \texttt{Hotel\_Info.json}. Under each hotel directory, retained reviews are divided into \texttt{Positive} and \texttt{Negative} folders. Each review is stored as \texttt{Review\_ID}, containing the downloaded media files and a \texttt{MetaReview\_ID.json} file with parsed metadata, original media URLs, and local media paths. Reviews requiring additional inspection are saved into \texttt{Manual\_MetaReview.jsonl}.

\begin{figure}[!htp]
\centering
\begin{minipage}{\textwidth}
\begin{example}{Trip.com User Review Record}
{
  "review_id": "1547772078",
  "hotel_id": 2197402,
  "hotel_name": "Travelodge Hotel Melbourne Docklands",
  "city": "Melbourne",
  "review_url": "https://www.trip.com/hotels/detail/?hotelId=2197402",
  "rating": 1.0,
  "stars": 1,
  "username": "[anonymized]",
  "user_avatar": "[omitted]",
  "travel_type": "Family",
  "room_type": "Guest King Room",
  "content": "Terrible experience - not worth \$250 at all!\n\nThis was one of the worst stays I've ever had. The room is small, dark, and outdated. They joined two single mattresses and called it a king - horrible idea. It was so uncomfortable I woke up with back pain. The blanket was old, both bedside lamps didn't work, and there was thick dust under the bed, clearly not cleaned properly.\n\nThe bathroom was tiny with cheap fittings, and the toilet paper holder kept falling off during use. They also charged me \$1 extra for a drink compared to the price on the menu - sneaky and dishonest.\n\nThis place seriously needs an upgrade. I won't come back, and I absolutely do not recommend it. Even budget motels are better than this!",
  "create_date": "2025-06-08 21:47:31",
  "check_in_date": "2025-06-01 00:00:00",
  "images": [
    {
      "original_url": "https://ak-d.tripcdn.com/images/0232w224x9227yyg2ED86.jpg"
    }
  ],
  "videos": [],
  "useful_count": 0,
  "reply": ""
}
\end{example}
\end{minipage}
\caption{Structure of a retained Trip.com user review record. User names and avatar links are redacted in this figure.}
\label{code:Tripcom_Review_Record}
\end{figure}

\subsubsection{GrabFood}

For the food delivery source, we collect restaurant review data from GrabFood, a large-scale food delivery platform that provides restaurant listings, merchant-level information, ratings, textual reviews, and user-uploaded review media. We use GrabFood as a complementary food-service source because food delivery reviews contain multimodal evidence about food quality, packaging condition, hygiene, freshness, and order presentation. Such reviews include both normal-condition food evidence and genuine problematic food evidence, making them suitable for constructing fraudulent refund or complaint evidence scenarios in the food delivery domain.

To collect restaurant candidates, we use Singapore as the target city, with the corresponding latitude, longitude, country code, GrabFood country path, and language configuration. We query the GrabFood guest category interface using the city coordinates and the food-category shortcut identifier. The returned listing records are parsed to extract restaurant identifiers, restaurant names, cuisine tags, ratings, addresses, coordinates, representative image URLs, price levels, opening status, and raw listing metadata. We further filter out non-restaurant businesses, such as grocery stores, convenience shops, supermarkets, pharmacies, pet stores, flower shops, health and beauty stores, and electronics shops. The retained stores are used as restaurant-level collection targets.

For each selected restaurant, we collect user reviews through an Apify GrabFood review actor, which returns structured review records from GrabFood. Reviews are fetched page by page using the restaurant identifier, language setting, and pagination token. Pagination continues until the target numbers of \texttt{positive} and \texttt{negative} reviews are reached, the maximum page limit is exhausted, or no further review token is returned. Since our benchmark focuses on multimodal evidence rather than text-only feedback, we retain only reviews with non-empty review text and at least one image or video attachment.

We organize the retained GrabFood reviews into two rating-derived subsets. The \texttt{positive} subset contains high-rating reviews, with 5-star reviews used as the primary source and 4-star reviews used as a fallback when needed. This subset represents normal or satisfactory delivered food evidence. The \texttt{negative} subset contains low-rating reviews, with 1-star reviews used as the primary source and 2-star reviews used as a fallback. This subset represents genuine problematic food delivery evidence. Ratings are used only as an initial grouping signal; subsequent multimodal review filtering removes irrelevant, subjective, or visually uninformative cases.

For each retained review, we preserve the available review-level metadata, including the review identifier, restaurant identifier, restaurant name, city, reviewer name, rating, review type, review text, posting time, scraping time, order identifier, and the original URLs of attached photos and videos. For restaurant-level context, we also preserve store metadata, including the store identifier, store name, city, cuisine tags, rating, address, latitude, longitude, representative image URL, price level, and opening status. This aligns the visual evidence with restaurant context for later image--text consistency checking and fake-evidence prompt construction.

For each attached media item, we download the corresponding image or video into the review folder. The downloader supports both image and video content types, assigns deterministic local filenames, and writes these filenames back into the review metadata. Failed downloads are skipped after exhausting retry attempts, while successfully downloaded files are stored together with the corresponding review JSON. This makes each review folder self-contained for benchmark construction and evaluation.

\begin{figure}[!htp]
\centering
\begin{minipage}{\textwidth}
\begin{example}{GrabFood User Review Record}
{
  "review_id": "297f4d91c9528a6a0f4719af3c2d80b0",
  "store_name": "Alt Pizza - Balestier",
  "store_id": "4-CYVAPGJZJ63GEX",
  "city": "Singapore",
  "reviewer_name": "[anonymized]",
  "rating": 5,
  "review_type": "positive",
  "review_text": "Love the pizza so much! Taste great!",
  "posted_at": "2025-08-24T09:57:50Z",
  "scraped_at": "2026-04-09T08:54:10.427802+00:00",
  "order_id": "",
  "photos": [
    {
      "id": "2czN2QzMxMTN1IjN1kTOykTM",
      ......
    },
    ......
  ],
  "videos": [],
  "has_media": true,
  "media_count": 3
}
\end{example}
\end{minipage}
\caption{Structure of a retained GrabFood user review record. User name is redacted in this figure.}
\label{code:GrabFood_Review_Record}
\end{figure}

Finally, we normalize the GrabFood storage format at the city, restaurant, and review levels. Each city is stored as a top-level directory, and each selected restaurant is stored under a directory named by its sanitized restaurant name and restaurant identifier. Restaurant-level metadata are saved as \texttt{Store\_Info.json}. Under each restaurant directory, retained reviews are divided into \texttt{Positive} and \texttt{Negative} folders. Each review is stored as \texttt{Review\_ID}, containing the downloaded media files and a \texttt{MetaReview\_ID.json} file with parsed metadata, original media URLs, and local media paths.

\subsubsection{Self-Collected Real-World Evidence}
In addition to the platform-derived data, we include a small set of self-collected images to improve coverage of common real-world refund-evidence scenarios that are underrepresented in the raw sources. This subset contains 24 images from seven review-level records (22 real-undamaged and two real-damaged images), all assigned to the Hotels \& Accommodations category according to their visible content; the real-undamaged images account for about 2.2\% of the \texttt{Positive} images. For each selected image, we constructed a small number of review-level metadata records by following the schema of existing \texttt{MetaReview} JSON files from the same category. These manually constructed records include category information, review text, rating-derived subset label, local image filename, and other fields required by the downstream data-loading and prompt-generation pipeline. The self-collected subset is used only as a minor supplement to the platform-derived data, providing additional coverage of realistic user-submitted evidence patterns and is separately documented for provenance tracking.

\subsubsection{Data Usage and Licensing}
The benchmark is constructed from publicly accessible review data across the above sources. Amazon Reviews 2023 is used in accordance with its publicly documented access terms~\citep{hou2024bridging}. Trip.com and GrabFood review records were collected from publicly accessible platform pages via API-based requests and direct collection. Because the copyright and licensing status of user-generated review media is not uniform across source platforms, we do not assert that all third-party review text, images, or metadata are covered by a single standardized open license. The released benchmark is restricted to research use, and users are responsible for complying with the applicable terms of the original platforms and data sources. Further details on privacy handling and responsible-use constraints are 
provided in Appx.~\ref{sec:responsible_practice}.

\subsection{Data Preprocessing}
\label{app:preprocessing}

We apply a multimodal preprocessing pipeline before constructing the benchmark. The pipeline covers MLLM-assisted image--review alignment filtering, human-conducted text and image cleaning, privacy-oriented anonymization, and human-verified quality control.

\paragraph{Image--Review Alignment Cleaning.}
To improve data quality, we apply platform-specific MLLM-based image relevance filters before finalizing each review record. In our implementation, each filter is built with \texttt{Qwen3.6-Plus}, accessed through the official API with default inference parameters, and returns a JSON object with a binary label and a short reason. The MLLM serves only as a preliminary screener; all final inclusion decisions are made by human annotators.

For Trip.com hotel reviews, the system prompt (Fig.~\ref{code:tripcom-relevance-filter-prompt}) defines the model as a hotel review image relevance classifier. For \texttt{positive} reviews, the filter retains images showing hotel-related content, such as rooms, bathrooms, beds, lobbies, restaurants, pools, gyms, or amenities, and rejects images dominated by unrelated scenery, tourist attractions, selfies, or unrecognizable content. For \texttt{negative} reviews, the filter retains images showing visible hotel problems, such as stains, dirt, damaged facilities, pests, mold, poor maintenance, plumbing issues, or unclean bathrooms.

\begin{figure}[!htp]
\centering
\begin{minipage}{\textwidth}
\begin{prompt}{Trip.com Image Relevance Filtering Prompt}
You are a hotel review image relevance classifier. You will receive one or more images
from a hotel review together with the review text and the sentiment label (positive or
negative). Your job is to decide whether the images are RELEVANT to the hotel review
context.

Respond ONLY with a JSON object: {"relevant": true/false, "reason": "..."}
Do NOT include any other text.
\end{prompt}
\end{minipage}
\caption{System prompt used for MLLM-based image relevance filtering of Trip.com hotel review media.}
\label{code:tripcom-relevance-filter-prompt}
\end{figure}

For GrabFood food delivery reviews, the system prompts (Figs.~\ref{code:grabfood-postive-review-filter-prompts} and~\ref{code:grabfood-negative-review-filter-prompts}) define separate criteria for each sentiment. For \texttt{positive} reviews, the filter retains samples whose images show the purchased food or drink item and removes non-food content such as app screenshots, delivery-rider photos, packaging-only photos, selfies, or unrecognizable images. For \texttt{negative} reviews, the filter retains images showing visible food-quality or packaging problems, such as spoilage, mold, foreign objects, undercooked food, spilled or crushed packaging, leaking or dirty containers, or hygiene concerns. Samples describing only subjective opinions, price complaints, delivery speed, or vague dissatisfaction are removed unless the attached media shows a verifiable quality issue.

\begin{figure}[!htp]
\centering
\begin{minipage}{\textwidth}
\begin{prompt}{GrabFood Positive Review Filtering Prompts}
You are a food delivery review classifier. You will receive the review text AND the
attached photo(s).

Given a single positive review with its photo(s), decide whether to **keep** or
**remove** it.

## KEEP (label = "keep")
The photo shows the actual purchased food or drink item.

## REMOVE (label = "remove")
The photo does NOT show food/drink, for example:
- Screenshot of the app or order page
- Photo of the delivery rider or packaging only
- Selfie, receipt, or any non-food image
- Blurry or unrecognizable image

Respond with a JSON object:
{"label": "keep" or "remove", "reason": "one-sentence explanation, 15 words max"}

Return ONLY the JSON object, no markdown fences, no extra text.
\end{prompt}
\end{minipage}
\caption{System prompts used for MLLM-based filtering of GrabFood positive review media, retaining reviews whose images show the purchased food or drink.}
\label{code:grabfood-postive-review-filter-prompts}
\end{figure}

\begin{figure}[!htp]
\centering
\begin{minipage}{\textwidth}
\begin{prompt}{GrabFood Negative Review Filtering Prompts}
You are a food delivery review classifier. You will receive the review text AND the
attached photo(s).

Given a single negative review with its photo(s), decide whether to **keep** or
**remove** it.

## KEEP (label = "keep")
The photo shows a visually observable food quality issue:
- Food spoilage or mold
- Foreign objects found in food (hair, insects, plastic, metal, etc.)
- Undercooked or raw
- Packaging problems: spilled, crushed, leaking, or dirty container
- Hygiene or cleanliness concerns

## REMOVE (label = "remove")
- The photo does not show any visible food quality issue
- False advertising: item received looks significantly different from the menu photo, or advertised ingredients are missing
- Purely subjective taste opinions ("not tasty", "too sweet", "too salty", "bland", "I didn't like the flavor")
- Price or value complaints only ("too expensive", "small portion for the price")
- Delivery speed complaints only ("took too long"), unless the delay caused food quality degradation
- Missing items or wrong items delivered
- Allergic reactions caused by undisclosed ingredients
- Vague dissatisfaction with no specific quality issue ("disappointing", "not worth it", "won't order again") unless combined with a keep-worthy reason
- Rating-only reviews with no actionable quality detail

Respond with a JSON object:
{"label": "keep" or "remove", "reason": "one-sentence explanation, 15 words max"}

Return ONLY the JSON object, no markdown fences, no extra text.
\end{prompt}
\end{minipage}
\caption{System prompts used for MLLM-based filtering of GrabFood negative review media, retaining reviews whose images show visible food-quality or packaging issues.}
\label{code:grabfood-negative-review-filter-prompts}
\end{figure}

\paragraph{Distribution-Aware Sampling.}
After filtering, we apply distribution-aware sampling to preserve structural diversity and evaluation usability across three dimensions. First, all 29 product and service categories are retained in the final benchmark. At the category level, the sampled distribution is kept as close as possible to the post-filtering distribution, avoiding over-representation of frequent categories while preserving the natural long-tail structure of the collected data. Second, multilingual review text is retained as collected rather than normalized to English, reflecting the natural language distribution of the source platforms and enabling evaluation under realistic multilingual user contexts. Third, within each category, reviews are stratified by the number of attached images into four bins: single-image (1), small multi-image (2--3), medium multi-image (4--6), and large multi-image (7--10). Sampling targets are assigned to maintain approximately balanced coverage across bins, with single-image reviews allocated a slightly larger share to reflect their higher prevalence in real platform submissions. This stratification ensures that the benchmark supports both single-image evaluation and multiple levels of multi-image context.

\paragraph{Human Verification.}
\label{appx:human_generate}
Human verification is conducted jointly by two annotators, the first two authors. The first author performs a systematic review of each retained record by inspecting the review text and attached images as stored in the \texttt{MetaReview} JSON files. This inspection serves two purposes: content verification and privacy-oriented anonymization. For content verification, each sample is assessed for task relevance, image--review alignment, and evidence usability; ambiguous, misleading, or insufficiently informative cases are flagged for exclusion. For anonymization, images and metadata are inspected for privacy-sensitive content, including visible human faces, personal identifiers such as names, addresses, order numbers, or contact information, and other directly identifying content; samples containing such information that cannot be reliably de-identified are excluded from the benchmark. The second author independently cross-examines the processed records to verify the consistency of inclusion decisions and the completeness of the anonymization procedure. Disagreements are resolved through joint discussion, and the final inclusion decision requires agreement from both annotators.

\subsection{Synthetic Evidence Generation}
\label{appx:synthetic_generation}

\paragraph{Models.}
\label{appx:generate_mllm_settings}
We use 12 state-of-the-art image editing and generation models to synthesize fake-damaged evidence: GPT Image 2, Grok Imagine Image, Nano Banana 2, Qwen-Image-2.0-Pro, Qwen-Image-Edit-Max and Wan2.7-Image-Pro, later extended with GPT Image 2.5 Flare, GPT Image 2.5 Sunburst, Grok Imagine Image 2.0, Nano Banana 2 Lite, Qwen-Image-3.0 and Qwen-Image-3.0-Pro to keep the threat model current as stronger generators reached general availability. The selection spans both image-editing and image-generation pipelines, covering a range of frontier capabilities across four vendors. GPT Image 2.5 represents a recent milestone in photorealistic image synthesis, while the remaining models offer competitive quality and serve as meaningful points of comparison.

\begin{table*}[!htp]
\centering
\caption{\textbf{Version and access details for image editing and generation models.} The table lists the display name used in the paper, the exact model identifier, and the API endpoint for each model.}
\label{tab:generate_mllm_access_version}
\resizebox{1\textwidth}{!}{%
\begin{tabular}{l l l}
\toprule
\textbf{Display Name} & \textbf{Exact Model ID} & \textbf{API Endpoint} \\
\midrule
GPT Image 2 & \texttt{gpt-image-2} & \texttt{api.openai.com/v1} \\
GPT Image 2.5 Flare & \texttt{gpt-image-2.5-flare-2026-09-08} & \texttt{api.openai.com/v1} \\
GPT Image 2.5 Sunburst & \texttt{gpt-image-2.5-sunburst-2026-09-08} & \texttt{api.openai.com/v1} \\
Grok Imagine Image & \texttt{grok-imagine-image} & \texttt{api.x.ai/v1} \\
Grok Imagine Image 2.0 & \texttt{grok-imagine-image-2.0} & \texttt{api.x.ai/v1} \\
Nano Banana 2 & \texttt{gemini-3.1-flash-image-preview} & \texttt{generativelanguage.googleapis.com/v1beta} \\
Nano Banana 2 Lite & \texttt{gemini-3.1-flash-lite-image} & \texttt{generativelanguage.googleapis.com/v1beta} \\
Qwen-Image-2.0-Pro & \texttt{qwen-image-2.0-pro} & \texttt{dashscope.aliyuncs.com/compatible-mode/v1} \\
Qwen-Image-3.0 & \texttt{qwen-image-3.0} & \texttt{dashscope.aliyuncs.com/compatible-mode/v1} \\
Qwen-Image-3.0-Pro & \texttt{qwen-image-3.0-pro} & \texttt{dashscope.aliyuncs.com/compatible-mode/v1} \\
Qwen-Image-Edit-Max & \texttt{qwen-image-edit-max} & \texttt{dashscope.aliyuncs.com/compatible-mode/v1} \\
Wan2.7-Image-Pro & \texttt{wan2.7-image-pro} & \texttt{dashscope.aliyuncs.com/compatible-mode/v1} \\
\bottomrule
\end{tabular}}
\end{table*}

Table~\ref{tab:generate_mllm_access_version} lists the exact model identifiers and API endpoints. Critically, all 12 are accessible through mainstream AI platforms or official API services with minimal configuration, reflecting a realistic threat model in which low-barrier tools can already be leveraged for synthetic evidence creation. All models are accessed through their respective official or authorized API services using provider-default inference settings, and we do not manually configure sampling parameters such as temperature, top-$p$, or maximum output tokens.

The two GPT Image 2.5 variants are sibling models split by workload rather than by capability, and both are pinned to the \texttt{2026-09-08} snapshot so that the benchmark remains reproducible after the bare alias moves. Nano Banana 2 is likewise pinned to \texttt{gemini-3.1-flash-image-preview}, the only variant exposed when the corresponding images were generated. A stable \texttt{gemini-3.1-flash-image} has since been released, but we retain the preview identifier so that the released images match the endpoint that produced them.

\paragraph{Damage Pattern Mining.}
Before constructing image-specific editing prompts, we first characterize 
the damage landscape within each data source by systematically analyzing 
real negative-review evidence. For each platform, we deploy a 
\texttt{Qwen3.6-Plus}-based multimodal analysis script that processes 
every retained negative-review record, jointly inspecting the attached 
images and the associated review text. The model is prompted to identify 
the product or scene category, the most prominent damage or quality issue 
visible in the images, and whether the visual evidence is consistent with 
the reviewer's written complaint. The analysis is designed to be source-specific: 
for Amazon e-commerce reviews, the prompt additionally leverages the 
product title, top-level category, and listing description to ground the 
damage identification within the known product context 
(Fig.~\ref{code:analyze-amazon-prompt}); for Trip.com hotel reviews, 
the prompt focuses on in-room locations and surface-level defects 
(Fig.~\ref{code:analyze-trip-prompt}); and for GrabFood food delivery 
reviews, the prompt targets food category identification and condition 
assessment (Fig.~\ref{code:analyze-grabfood-prompt}). Each analysis call 
returns a structured JSON object containing the damage type, a short 
issue description, a review-consistency judgment, and a confidence level. 
The per-review outputs are aggregated by product or service category to 
produce category-specific candidate damage lists, which enumerate the 
most frequently observed and visually verifiable defect types for each 
category. These lists form the basis of the subsequent image-specific 
prompt construction pipeline.

\begin{figure}[!htp]
\centering
\begin{minipage}{\textwidth}
\begin{prompt}{Amazon Damage Analysis Prompt}
You are a senior product-quality auditor specialising in e-commerce review 
images on Amazon. The user will tell you the *known* Amazon top-level 
category for the listing (e.g. "Electronics", "All_Beauty", "Books") AND 
the listing's product name and short description so you know exactly what 
the buyer ordered. Your task is to analyse the review image(s) together 
with the reviewer's written comment and identify:

  1. product_subtype   - a more specific product label within the known 
category (e.g. for "Electronics": "Wireless Earbuds", "USB Cable", 
"Laptop Charger"; for "All_Beauty": "Lipstick", "Shampoo Bottle", 
"Perfume Bottle"; for "Books": "Hardcover Book", "Paperback Book"). 
Prefer wording consistent with the supplied product name.
  2. damage_type       - ONE concise label describing the most prominent 
issue, drawn from this controlled vocabulary where possible:
        normal, shattered/broken, cracked, dented, scratched, leaking, 
        stained/dirty, mouldy/spoiled, melted, bent/warped, torn/ripped, 
        missing_parts, wrong_item, used/opened, packaging_damaged, 
        expired, defective_function, color_or_print_wrong, size_wrong, 
        counterfeit_suspected, other_defect.
     If the item looks fine, use "normal".
  3. issue_detail      - one or two sentences describing what you actually 
see in the image(s).
  4. review_consistent - true if the visual evidence matches what the 
reviewer wrote, otherwise false.
  5. confidence        - high | medium | low.

Respond ONLY with a valid JSON object matching this schema, no prose 
outside JSON:
{
  "product_subtype":   "<string>",
  "damage_type":       "<string - one label from the vocabulary above>",
  "issue_detail":      "<string>",
  "review_consistent": <true | false>,
  "confidence":        "<high | medium | low>"
}
\end{prompt}
\end{minipage}
\caption{System prompt used for MLLM-based damage pattern analysis of Amazon 
e-commerce review images.}
\label{code:analyze-amazon-prompt}
\end{figure}

\begin{figure}[!htp]
\centering
\begin{minipage}{\textwidth}
\begin{prompt}{Trip.com Damage Analysis Prompt}
You are a senior hospitality quality auditor for online travel platforms 
(e.g. Trip.com). Every image you analyse has been uploaded with a guest 
review and is assumed to show something INSIDE the guest's hotel room. 
Public/shared areas (lobby, reception, corridor, elevator, pool, gym, 
restaurant, breakfast area, exterior of the building) are OUT OF SCOPE. 
If an image clearly shows such a public area, label its room_location as 
"Not In-Room".

Your task is to identify:
  1. The specific IN-ROOM location shown. Pick the single most appropriate 
label from this closed list:
     - "Bed / Bedding"            (mattress, sheets, pillows, headboard)
     - "Bathroom - Shower / Tub"
     - "Bathroom - Toilet"
     - "Bathroom - Sink / Vanity"
     - "Bathroom - General"       (bathroom floor, whole-bathroom shot)
     - "Floor / Carpet"           (room floor, rug, carpet)
     - "Wall / Ceiling"           (paint, wallpaper, stains on wall/ceiling)
     - "Window / Curtains"        (window glass, drapes, blinds)
     - "Desk / Work Area"
     - "Closet / Wardrobe"
     - "TV / Entertainment Area"
     - "Minibar / Fridge / Kettle"
     - "Air Conditioner / Heater"
     - "Door / Entry"
     - "Balcony (private, part of room)"
     - "Room - General / Overall" (wide shot of the whole room)
     - "Other In-Room"
     - "Not In-Room"              (use only if clearly a public area)
  2. The specific problem visible at that in-room location (e.g. 
"stained sheets", "hair on pillow", "mouldy grout", "clogged drain", 
"cracked tile", "broken remote", "dust on surface", "cockroach / insect", 
"worn / outdated furniture", "damaged wallpaper", "dirty carpet", 
"leaking / water damage", "clean / normal"). Keep it short and concrete.
  3. Whether the visual evidence is consistent with the reviewer's written 
comment.

Respond ONLY with a valid JSON object matching this schema -- no prose 
outside JSON:
{
  "room_location":     "<one label from the list above>",
  "issue":             "<short concrete label, e.g. 'stained sheets', 'mould', 'clean / normal'>",
  "issue_detail":      "<one or two sentences describing what you observe>",
  "review_consistent": <true | false>,
  "confidence":        "<high | medium | low>"
}
\end{prompt}
\end{minipage}
\caption{System prompt used for MLLM-based damage pattern analysis of 
Trip.com hotel review images.}
\label{code:analyze-trip-prompt}
\end{figure}

\begin{figure}[!htp]
\centering
\begin{minipage}{\textwidth}
\begin{prompt}{GrabFood Damage Analysis Prompt}
You are a senior food-quality auditor specialising in food-delivery 
platforms. Your task is to analyse food images from customer reviews 
and identify:
  1. The precise food category (e.g. "Pizza", "Fried Chicken", 
"Noodles / Ramen", "Burger", "Rice Dish", "Sushi / Japanese", "Dessert", 
"Beverage", "Other").
  2. The quality or condition of the food shown (e.g. normal/acceptable, 
burnt, undercooked, foreign object present, incorrect order, packaging 
damaged, portion too small, spillage, mould/spoilage, other defect).
  3. Whether the visual evidence is consistent with the reviewer's 
written complaint.

Respond ONLY with a valid JSON object matching this schema -- no prose 
outside JSON:
{
  "food_category":   "<string>",
  "food_condition":  "<string -- one concise label, e.g. 'burnt', 'foreign object', 'normal'>",
  "issue_detail":    "<string -- one or two sentences describing what you observe>",
  "review_consistent": <true | false>,
  "confidence":      "<high | medium | low>"
}
\end{prompt}
\end{minipage}
\caption{System prompt used for MLLM-based damage pattern analysis of 
GrabFood food delivery review images.}
\label{code:analyze-grabfood-prompt}
\end{figure}

\paragraph{Image-Specific Prompt Construction.}

The damage pattern mining yields prioritized category-specific damage 
shortlists that form the structured defect vocabulary for prompt 
construction. For Amazon e-commerce data, the mining covers all 27 product categories and identifies 12 visually verifiable damage types: 
\textit{shattered/broken}, \textit{cracked}, \textit{dented}, 
\textit{scratched}, \textit{leaking}, \textit{stained/dirty}, 
\textit{mouldy/spoiled}, \textit{melted}, \textit{bent/warped}, 
\textit{torn/ripped}, \textit{packaging-damaged}, and 
\textit{other\_defect}. The per-category shortlists are shown in 
Tab.~\ref{tab:category_damage_shortlist}; within 
each row, damage types are ordered by their empirical frequency in real 
negative reviews, so the first entry reflects the most commonly observed 
defect in that category. Damage types that are inherently non-visual or 
inference-heavy, including \textit{missing parts}, 
\textit{defective function}, \textit{used/opened}, and 
\textit{wrong item}, are excluded from the vocabulary because they cannot 
be reliably synthesized as visually verifiable evidence. For Trip.com 
hotel reviews, the analysis produces location-aware damage lists organized 
by in-room location (e.g., Bed/Bedding, Bathroom, Floor/Carpet), covering 
defect types such as stained linens, mould, broken fixtures, and surface 
contamination. For GrabFood food delivery reviews, the analysis produces 
food-category-level damage lists covering quality issues such as spoilage, 
foreign objects, spillage, and packaging damage.

\begin{table}[!htp]
\centering
\caption{\textbf{Per-category damage shortlists for Amazon product 
categories.} Damage types are ordered by empirical frequency in real 
negative reviews; the first entry in each row is the most commonly 
observed defect. These shortlists are used as the primary candidate 
vocabulary during image-specific prompt construction.}
\label{tab:category_damage_shortlist}
\small
\setlength{\tabcolsep}{4pt}
\begin{tabular}{l p{9cm}}
\toprule
\textbf{Category} & \textbf{Damage Shortlist (in priority order)} \\
\midrule
All Beauty 
  & shattered/broken, leaking, dented, packaging-damaged, cracked \\
Amazon Fashion 
  & torn/ripped, stained/dirty, scratched \\
Appliances 
  & shattered/broken, cracked, dented, scratched, packaging-damaged, bent/warped \\
Arts, Crafts \& Sewing 
  & leaking, stained/dirty, packaging-damaged, bent/warped, torn/ripped, dented, shattered/broken \\
Automotive 
  & cracked, scratched, stained/dirty, dented, bent/warped, packaging-damaged \\
Baby Products 
  & shattered/broken, torn/ripped, leaking, packaging-damaged, stained/dirty, mouldy/spoiled, cracked \\
Beauty \& Personal Care 
  & shattered/broken, leaking, packaging-damaged, cracked, stained/dirty, mouldy/spoiled \\
Books 
  & torn/ripped, stained/dirty, bent/warped, scratched, packaging-damaged \\
CDs \& Vinyl 
  & scratched, cracked, bent/warped, packaging-damaged, stained/dirty, shattered/broken \\
Cell Phones \& Accessories 
  & torn/ripped, scratched, cracked, shattered/broken, packaging-damaged \\
Clothing, Shoes \& Jewelry 
  & torn/ripped, stained/dirty, scratched, shattered/broken \\
Electronics 
  & shattered/broken, cracked, scratched, bent/warped, packaging-damaged, dented \\
Grocery \& Gourmet Food 
  & melted, mouldy/spoiled, dented, leaking, packaging-damaged, stained/dirty \\
Handmade Products 
  & stained/dirty, cracked, melted, shattered/broken, scratched, dented, packaging-damaged \\
Health \& Household 
  & leaking, shattered/broken, packaging-damaged, mouldy/spoiled \\
Health \& Personal Care 
  & mouldy/spoiled, packaging-damaged, leaking \\
Home \& Kitchen 
  & cracked, shattered/broken, stained/dirty, dented, scratched, melted, packaging-damaged, bent/warped \\
Industrial \& Scientific 
  & shattered/broken, cracked, leaking, scratched, packaging-damaged, bent/warped \\
Magazine Subscriptions 
  & torn/ripped, bent/warped, stained/dirty, packaging-damaged \\
Musical Instruments 
  & shattered/broken, cracked, bent/warped, scratched, dented, packaging-damaged \\
Office Products 
  & bent/warped, leaking, cracked, dented, packaging-damaged, torn/ripped, scratched, stained/dirty \\
Patio, Lawn \& Garden 
  & shattered/broken, cracked, scratched, packaging-damaged, bent/warped, torn/ripped, mouldy/spoiled, dented \\
Pet Supplies 
  & torn/ripped, leaking, bent/warped, stained/dirty, mouldy/spoiled, packaging-damaged \\
Sports \& Outdoors 
  & torn/ripped, stained/dirty, cracked, scratched, dented, bent/warped, packaging-damaged \\
Tools \& Home Improvement 
  & cracked, leaking, dented, packaging-damaged, scratched, bent/warped \\
Toys \& Games 
  & packaging-damaged, cracked, shattered/broken, scratched, torn/ripped, stained/dirty \\
Video Games 
  & cracked, shattered/broken, torn/ripped, scratched, packaging-damaged \\
\bottomrule
\end{tabular}
\end{table}

Building on the category-specific damage shortlists characterized above, 
we construct synthetic evidence prompts through a two-stage pipeline. 
In the first stage, the shortlists serve as a structured vocabulary of 
visually verifiable defects for each product category. In the second 
stage, for each target image, an MLLM analyzes the visible product, 
packaging, material, and physical attributes, and combines this visual 
analysis with product metadata including the title, category, description, 
and features. The corresponding category-specific shortlist is then 
retrieved, and the most plausible damage pattern is selected according to 
the object, material, scene, and metadata. When no category-specific 
candidate is suitable, the selection falls back to the universal damage 
vocabulary. The selected damage pattern is then converted into an 
image-specific editing prompt. The prompt specifies that the modification 
should be visible, photorealistic, physically plausible, and localized to 
the target object, while preserving the original product identity, 
viewpoint, lighting, background, surrounding objects, and overall 
composition. Each generated sample is also paired with a fake review 
comment, which is later used in review-conditioned evaluation.

\paragraph{Image Generation.}
The constructed edit prompt, together with the original real-undamaged image, is submitted to all 12 image editing and generation models. To ensure a fair comparison across models, the visual analysis and damage selection in the two-stage pipeline are executed once per image, and the resulting edit instruction is shared across all 12 backends without modification, so that any differences in generated output are attributable to model behavior rather than prompt variation. Each model independently produces a synthetic fake-damaged image by applying the specified modification to the original. The generated image, the selected damage type, the full edit instruction, and the associated review comment are stored together as a structured metadata record for each synthetic sample. The generated review comment, produced as part of Stage 2, serves as the fake review text in review-conditioned evaluation settings.

\subsection{Region-Level Annotations}
\label{appx:region_annotations}

\paragraph{Derivation.} Each fake-damaged image in \method{} is an edit of a single real-undamaged image, so the edited area can be recovered from the residual between the two images. Since the generators re-render the whole frame, the raw difference also contains global tone and resampling shifts. We therefore resize both images to a common $1024 \times 1024$ canvas, apply a light Gaussian blur, and match the per-channel gain and offset of the edited image to its source before computing the pixel-wise color difference, which is then smoothed with a second Gaussian filter. The difference map is normalized by its median and median absolute deviation, and pixels whose \textit{\textbf{robust z-score exceeds 4}} are marked as edited. The resulting mask is cleaned by morphological opening, closing, and hole filling, and connected components covering less than 0.15\% of the frame are discarded. The remaining components form the annotated regions, ordered by area. Since the annotations are derived automatically rather than drawn by hand, every record is marked as machine-derived.

\paragraph{Annotation Record.} For every region, the annotation provides a bounding box in normalized and pixel coordinates, a polygon traced along the region boundary and simplified with the Ramer--Douglas--Peucker algorithm, the area fraction, and the peak and mean z-scores as a measure of edit strength. A binary mask stores the pixel-level extent of all regions, and each region references its connected component in this mask. The largest region is marked as the primary damage region and inherits the damage type and target part of the edit metadata, while the remaining regions are marked as secondary evidence, such as spatter, debris, or incidental re-rendering. A \texttt{claim\_grounding} field links the generated claim to the primary region through the claimed damage type and target part, and the whole annotation is attached as a \texttt{localization} block to the entry of the image in the edit metadata. Fig.~\ref{code:region-annotation-record} shows a representative record.

\begin{figure}[!htp]
\centering
\begin{minipage}{\textwidth}
\begin{example}{Region-Level Annotation}
"localization": {
  "annotation_schema": "fraudbench-grounding-v1",
  "generator": "qwen-image-edit-max",
  "image_size": {"width": 1536, "height": 1024},
  "source_image": "Electronics/Positive/Review_013/Image_013_02.jpg",
  "edited_image": "Electronics/DeepFake/qwen-image-edit-max/Review_013/Image_013_02.jpg",
  "manipulated_area_fraction": 0.08908,
  "n_regions": 1,
  "edit_is_local": true,
  "regions": [
    {
      "region_id": "r1",
      "role": "primary_damage",
      "damage_type": "cracked",
      "target_part": "matte black plastic headband arm just above the hinge mechanism",
      "bbox_normalized": [0.3203, 0.1309, 0.8086, 0.7734],
      "bbox_pixels": [491, 134, 1242, 791],
      "area_fraction": 0.08867,
      "area_pixels": 139473,
      "polygon_normalized": [[0.6616, 0.1309], [0.6685, 0.1309], [0.6685, 0.1396], ...],
      "mask_ref": "masks/Electronics__Review_013__Image_013_02__mask.png#component=1",
      "residual_strength": {"peak_z": 49.8, "mean_z": 11.8}
    }
  ],
  "claim_grounding": {
    "claim_text": "The headband snapped right at the hinge the first time I tried to fold them up, completely useless now.",
    "claimed_damage_type": "cracked",
    "claimed_target_part": "matte black plastic headband arm just above the hinge mechanism",
    "primary_region_id": "r1"
  },
  "provenance": {
    "geometry_source": "source-vs-edit residual, MAD-normalised",
    "params": {"canvas": 1024, "mad_k": 4.0, "min_area": 0.0015},
    "human_verified": false
  }
}
\end{example}
\end{minipage}
\caption{Structure of the region-level annotation attached to an image entry of the edit metadata. Most polygon vertices are omitted for brevity.}
\label{code:region-annotation-record}
\end{figure}

\subsection{Quality Control}
\label{appx:quality_control}
Quality control is applied after both raw-data preprocessing and synthetic 
image generation. In both stages, an MLLM serves as a preliminary screener 
to surface candidate quality issues; the specific screening criteria and 
prompts for each stage are described in 
Appx.~\ref{app:preprocessing} and~\ref{appx:synthetic_generation}. Screened issues include weak image--review alignment, 
privacy-sensitive content, low-quality files, implausible edits, prompt 
misalignment, and trivial generation artifacts. The MLLM does not assign 
final labels or determine sample validity; human annotators examine all 
flagged cases and make the final inclusion decision according to predefined 
quality-control criteria. Separating automated screening from human 
adjudication reduces the risk that MLLM-assisted filtering introduces 
systematic bias into the benchmark.
\newpage
\section{Benchmark Evaluation}
\label{appx:evaluation}

\subsection{Evaluation Settings}
\label{appx:eval_settings}

\subsubsection{Multimodal Language Models}
\label{appx:eval_mllm_settings}
\paragraph{Models.}
We evaluate 30 state-of-the-art MLLMs, including DeepSeek-V4.1-Flash~\citep{xu2026deepseek}, Gemini 3.5 Flash-Lite~\citep{google2026gemini35flashlite}, Gemini 3 Flash~\citep{google2025gemini3flash}, Gemini 3.8 Flash~\citep{google2026gemini38flash}, Gemini 3.1 Pro~\citep{google2026gemini31pro}, GPT-5.4 mini~\citep{openai2026gpt54mini}, GPT-5.6 Luna, GPT-5.6 Terra and GPT-5.6 Sol~\citep{openai2026gpt56}, GPT-6 Luna, GPT-6 Sol~\citep{openai2026gpt6solluna} and GPT-6 Astra~\citep{openai2026gpt6astra}, Grok 4.1 Fast Reasoning~\citep{xai2025grok41fast}, Grok 4.20 Reasoning~\citep{xai2026grok420}, Grok 4.6~\citep{xai2026grok46}, Kimi K2.6~\citep{moonshot2026kimi26}, Kimi K3~\citep{team2026kimi}, QVQ-Max-Latest~\citep{qwen2024qvq}, Qwen3-VL-Flash and Qwen3-VL-Plus~\citep{bai2025qwen3}, Qwen3.5-Omni-Flash and Qwen3.5-Omni-Plus~\citep{team2026qwen3}, Qwen3.6-35B-A3B~\citep{qwen2026qwen36flash}, Qwen3.6-Plus~\citep{qwen2026qwen36plus}, Qwen3.7-Flash~\citep{qwen2026qwen37}, Qwen3.7-Plus~\citep{qwen2026qwen37plus}, Qwen3.7-Max~\citep{qwen2026qwen37}, Qwen3.8-27B~\citep{qwen38}, Qwen3.8-Flash~\citep{qiu2026design}, and Qwen3.8-Max~\citep{qwen38}. This model set covers proprietary and open-weight models, spanning general-purpose multimodal models, vision-language models, omni-modal models, visual-reasoning models, and multimodal Mixture-of-Experts models, and pairs several models with a same-vendor successor so that detector behavior can be compared across model generations as well as across vendors.

Table~\ref{tab:mllm_access_version} reports the exact model identifiers and API endpoints used in our experiments. Table~\ref{tab:mllm_overview} further summarizes the evaluated MLLMs and their providers and model categories. All models are accessed through their respective official or authorized API access, and evaluated using provider-default inference settings unless otherwise specified.

\begin{table*}[!htp]
\centering
\caption{\textbf{Version and access details for evaluated MLLMs.} The table lists the display name used in the paper, the exact model identifier, and the API endpoint used for each evaluated model.}
\label{tab:mllm_access_version}
\resizebox{\textwidth}{!}{%
\begin{tabular}{l l l}
\toprule
\textbf{Display Name} & \textbf{Exact Model ID} & \textbf{API Endpoint} \\
\midrule
Gemini 3.5 Flash-Lite & \texttt{gemini-3.5-flash-lite} & \texttt{generativelanguage.googleapis.com/v1beta} \\
Gemini 3 Flash & \texttt{gemini-3-flash-preview} & \texttt{generativelanguage.googleapis.com/v1beta} \\
Gemini 3.8 Flash & \texttt{gemini-3.8-flash} & \texttt{generativelanguage.googleapis.com/v1beta} \\
Gemini 3.1 Pro & \texttt{gemini-3.1-pro-preview} & \texttt{generativelanguage.googleapis.com/v1beta} \\
GPT-5.4 mini & \texttt{gpt-5.4-mini} & \texttt{api.openai.com/v1} \\
GPT-5.6 Luna & \texttt{gpt-5.6-luna} & \texttt{api.openai.com/v1} \\
GPT-5.6 Terra & \texttt{gpt-5.6-terra} & \texttt{api.openai.com/v1} \\
GPT-5.6 Sol & \texttt{gpt-5.6-sol} & \texttt{api.openai.com/v1} \\
GPT-6 Luna & \texttt{gpt-6-luna} & \texttt{api.openai.com/v1} \\
GPT-6 Sol & \texttt{gpt-6-sol} & \texttt{api.openai.com/v1} \\
GPT-6 Astra & \texttt{gpt-6-astra} & \texttt{api.openai.com/v1} \\
Grok 4.1 Fast Reasoning & \texttt{grok-4-1-fast-reasoning} & \texttt{api.x.ai/v1} \\
Grok 4.20 Reasoning & \texttt{grok-4.20-reasoning-latest} & \texttt{api.x.ai/v1} \\
Grok 4.6 & \texttt{grok-4.6} & \texttt{api.x.ai/v1} \\
QVQ-Max-Latest & \texttt{qvq-max-latest} & \texttt{dashscope.aliyuncs.com/compatible-mode/v1} \\
Qwen3-VL-Flash & \texttt{qwen3-vl-flash} & \texttt{dashscope.aliyuncs.com/compatible-mode/v1} \\
Qwen3-VL-Plus & \texttt{qwen3-vl-plus} & \texttt{dashscope.aliyuncs.com/compatible-mode/v1} \\
Qwen3.5-Omni-Flash & \texttt{qwen3.5-omni-flash-2026-03-15} & \texttt{dashscope.aliyuncs.com/compatible-mode/v1} \\
Qwen3.5-Omni-Plus & \texttt{qwen3.5-omni-plus-2026-03-15} & \texttt{dashscope.aliyuncs.com/compatible-mode/v1} \\
Qwen3.6-Plus & \texttt{qwen3.6-plus} & \texttt{dashscope.aliyuncs.com/compatible-mode/v1} \\
Qwen3.7-Flash & \texttt{qwen3.7-flash-2026-07-15} & \texttt{dashscope.aliyuncs.com/compatible-mode/v1} \\
Qwen3.7-Plus & \texttt{qwen3.7-plus-2026-05-26} & \texttt{dashscope.aliyuncs.com/compatible-mode/v1} \\
Qwen3.7-Max & \texttt{qwen3.7-max-2026-06-08} & \texttt{dashscope.aliyuncs.com/compatible-mode/v1} \\
Qwen3.8-Max & \texttt{qwen3.8-max-0902} & \texttt{dashscope.aliyuncs.com/compatible-mode/v1} \\
DeepSeek-V4.1-Flash & \texttt{deepseek-v4.1-flash} & \texttt{dashscope.aliyuncs.com/compatible-mode/v1} \\
Kimi K2.6 & \texttt{kimi-k2.6} & \texttt{dashscope.aliyuncs.com/compatible-mode/v1} \\
Kimi K3 & \texttt{kimi-k3} & \texttt{dashscope.aliyuncs.com/compatible-mode/v1} \\
Qwen3.6-35B-A3B & \texttt{qwen3.6-flash} & \texttt{dashscope.aliyuncs.com/compatible-mode/v1} \\
Qwen3.8-27B & \texttt{qwen3.8-27b} & \texttt{dashscope.aliyuncs.com/compatible-mode/v1} \\
Qwen3.8-Flash & \texttt{qwen3.8-flash} & \texttt{dashscope.aliyuncs.com/compatible-mode/v1} \\
\bottomrule
\end{tabular}}
\end{table*}

\begin{table*}[!htp]
\centering
\label{appx:eval_full_list}
\caption{\textbf{Overview of MLLMs evaluated in \method{}.} We summarize the evaluated model set and the input modalities supported by each model. \checkmark indicates that the model accepts the corresponding input modality, while $\times$ indicates that it does not. All models produce text output.}
\label{tab:mllm_overview}
\small
\setlength{\tabcolsep}{4.5pt}
\resizebox{\textwidth}{!}{%
\begin{tabular}{l l c c c c l}
\toprule
\textbf{Model} & \textbf{Provider} & \textbf{Text} & \textbf{Image} & \textbf{Audio} & \textbf{Video} & \textbf{Model Type} \\
\midrule
\rowcolor{mypurple}
\multicolumn{7}{c}{\textbf{Proprietary MLLMs}} \\
\midrule
Gemini 3.5 Flash-Lite & Google & \checkmark & \checkmark & \checkmark & \checkmark & Lightweight multimodal \\
Gemini 3 Flash & Google & \checkmark & \checkmark & \checkmark & \checkmark & Fast multimodal \\
Gemini 3.8 Flash & Google & \checkmark & \checkmark & \checkmark & \checkmark & Fast multimodal \\
Gemini 3.1 Pro & Google & \checkmark & \checkmark & \checkmark & \checkmark & Reasoning multimodal \\
GPT-5.4 mini & OpenAI & \checkmark & \checkmark & $\times$ & $\times$ & General-purpose multimodal \\
GPT-5.6 Luna & OpenAI & \checkmark & \checkmark & $\times$ & $\times$ & Lightweight multimodal \\
GPT-5.6 Terra & OpenAI & \checkmark & \checkmark & $\times$ & $\times$ & General-purpose multimodal \\
GPT-5.6 Sol & OpenAI & \checkmark & \checkmark & $\times$ & $\times$ & General-purpose multimodal \\
GPT-6 Luna & OpenAI & \checkmark & \checkmark & $\times$ & $\times$ & Lightweight multimodal \\
GPT-6 Sol & OpenAI & \checkmark & \checkmark & $\times$ & $\times$ & General-purpose multimodal \\
GPT-6 Astra & OpenAI & \checkmark & \checkmark & $\times$ & $\times$ & General-purpose multimodal \\
Grok 4.1 Fast Reasoning & xAI & \checkmark & \checkmark & $\times$ & $\times$ & Reasoning multimodal \\
Grok 4.20 Reasoning & xAI & \checkmark & \checkmark & $\times$ & $\times$ & Reasoning multimodal \\
Grok 4.6 & xAI & \checkmark & \checkmark & $\times$ & $\times$ & Reasoning multimodal \\
QVQ-Max-Latest & DashScope & \checkmark & \checkmark & $\times$ & $\times$ & Visual reasoning \\
Qwen3-VL-Flash & DashScope & \checkmark & \checkmark & $\times$ & \checkmark & Vision-language \\
Qwen3-VL-Plus & DashScope & \checkmark & \checkmark & $\times$ & \checkmark & Vision-language \\
Qwen3.5-Omni-Flash & DashScope & \checkmark & \checkmark & \checkmark & \checkmark & Omni-modal \\
Qwen3.5-Omni-Plus & DashScope & \checkmark & \checkmark & \checkmark & \checkmark & Omni-modal \\
Qwen3.6-Plus & DashScope & \checkmark & \checkmark & $\times$ & \checkmark & General-purpose multimodal \\
Qwen3.7-Flash & DashScope & \checkmark & \checkmark & $\times$ & \checkmark & Fast multimodal \\
Qwen3.7-Plus & DashScope & \checkmark & \checkmark & $\times$ & \checkmark & General-purpose multimodal \\
Qwen3.7-Max & DashScope & \checkmark & \checkmark & $\times$ & \checkmark & General-purpose multimodal \\
Qwen3.8-Max & DashScope & \checkmark & \checkmark & $\times$ & \checkmark & General-purpose multimodal \\
\midrule
\rowcolor{mypurple}
\multicolumn{7}{c}{\textbf{Open-Weight MLLMs}} \\
\midrule
DeepSeek-V4.1-Flash & DashScope & \checkmark & \checkmark & $\times$ & $\times$ & Open-weight multimodal \\
Kimi K2.6 & DashScope & \checkmark & \checkmark & $\times$ & $\times$ & Open-weight multimodal \\
Kimi K3 & DashScope & \checkmark & \checkmark & $\times$ & $\times$ & Open-weight multimodal \\
Qwen3.6-35B-A3B & DashScope & \checkmark & \checkmark & $\times$ & \checkmark & Multimodal MoE \\
Qwen3.8-27B & DashScope & \checkmark & \checkmark & $\times$ & \checkmark & Open-weight multimodal \\
Qwen3.8-Flash & DashScope & \checkmark & \checkmark & $\times$ & \checkmark & Fast multimodal \\
\bottomrule
\end{tabular}}
\end{table*}

\newpage

\paragraph{Evaluation Prompt Design.}
We use a unified prompt template across all MLLM-based evaluation settings, with setting-specific user instructions for single-image, multi-image, and review-augmented inputs. The same review-text template, which refers to a purchase on Amazon, is also used for the Trip.com and GrabFood categories.
A shared system prompt defines the model as a forensic image analyst and specifies the expected JSON-only response format, as shown in Fig.~\ref{code:shared-system-prompt}.

\begin{figure}[!htp]
\centering
\begin{minipage}{\textwidth}
\begin{prompt}{Shared System Prompt for MLLM-based Detection}
You are a forensic image analyst specialising in detecting images that have been produced or modified by AI -- either through AI image-editing LLMs or through AI image-generation LLMs. You look for characteristic traces such as local patch inconsistencies inside an otherwise plausible scene, object insertion or removal seams, mismatched shadows or reflections on specific objects, blurred or resampled regions around modified areas, copy-paste texture repetition, broken or re-drawn text and logos, anatomical or geometric implausibilities, over-smoothed textures, and diffusion-style artefacts. You never refuse the task and you always return valid JSON.
\end{prompt}
\end{minipage}
\caption{Shared system prompt used to establish the forensic-analysis role for MLLM-based AI-generated evidence detection.}
\label{code:shared-system-prompt}
\end{figure}

For single-image evaluation, each image is submitted independently using the prompt shown in Fig.~\ref{code:single-image-detection-prompt}.

\begin{figure}[!htp]
\centering
\begin{minipage}{\textwidth}
\begin{prompt}{Single-Image Detection Prompt}
Analyse the provided image and decide whether it has been produced or modified by AI -- either through AI image-editing LLMs or through AI image-generation LLMs -- or whether it is an unmodified genuine camera photograph of a real scene.

Return ONLY a single JSON object -- no markdown fences, no prose outside the object -- with this exact schema:
{
  "is_ai_modified": <true or false>,
  "confidence":     <number between 0 and 1>,
  "reason":         "<1-3 concise sentences naming the specific visual evidence you relied on>"
}
\end{prompt}
\end{minipage}
\caption{Single-image prompt used for review-free MLLM-based detection.}
\label{code:single-image-detection-prompt}
\end{figure}

When review text is available, the review-text instruction is inserted between the image-analysis instruction and the JSON schema.

\begin{figure}[!htp]
\centering
\begin{minipage}{\textwidth}
\begin{prompt}{Review-Text Augmentation Prompt}
The customer who purchased this product on Amazon left the following review:
"{review text}"
Consider this customer review when making your assessment.
\end{prompt}
\end{minipage}
\caption{Review-text augmentation inserted between the image-analysis instruction and the JSON output schema when review text is available.}
\label{code:review-text-augmentation-prompt}
\end{figure}

For multi-image evaluation, we consider two settings. The single-turn setting packs all images from the same review into one user message, using the prompt shown in Fig.~\ref{code:single-turn-multi-image-prompt}. The multi-step setting holds a multi-turn conversation in which the images of a review are submitted one per turn and the model's replies are kept in the conversation history. The shared system prompt is extended with the sentence in Fig.~\ref{code:multi-step-system-prompt}. For each non-final image, the model gives a short forensic observation without a verdict (Figs.~\ref{code:multi-step-first-prompt} and~\ref{code:multi-step-continuation-prompt}), and the final turn requests the consolidated verdict (Fig.~\ref{code:multi-step-final-prompt}). In review-conditioned runs, the review-text block of Fig.~\ref{code:review-text-augmentation-prompt} is appended to the turn of the first image it refers to, and each distinct review text is shown only once. Inline comments (\texttt{//}) present in the JSON schema of the actual prompts are omitted from the figures for brevity.

\begin{figure}[!htp]
\centering
\begin{minipage}{\textwidth}
\begin{prompt}{Multi-Image Detection Prompt}
The {n} images above all come from the same product listing. Examine all {n} images together and decide whether they have been produced or modified by AI -- either through AI image-editing LLMs or through AI image-generation LLMs -- or whether they are unmodified genuine camera photographs of a real product.

Return ONLY a single JSON object -- no markdown fences, no prose outside the object -- with this exact schema:
{
  "is_ai_modified": <true or false>,
  "confidence":     <number between 0 and 1>,
  "reason":         "<1-3 sentences explaining the combined verdict over all {n} images>",
  "per_image": [
    {
      "index":          <1-indexed position, 1..{n}>,
      "is_ai_modified": <true or false>,
      "confidence":     <number between 0 and 1>,
      "notes":          "<1-2 sentences on visual evidence from THIS image>"
    },
    ...
  ]
}
\end{prompt}
\end{minipage}
\caption{Multi-image prompt used when all images from the same review are submitted in one user message.}
\label{code:single-turn-multi-image-prompt}
\end{figure}

\begin{figure}[!htp]
\centering
\begin{minipage}{\textwidth}
\begin{prompt}{Multi-Step System Prompt Extension}
[Shared system prompt] You are reviewing a customer refund claim in which the customer uploads their evidence photos one message at a time. Answer each message as it arrives, and carry your earlier observations forward.
\end{prompt}
\end{minipage}
\caption{Sentence appended to the shared system prompt in the multi-step setting.}
\label{code:multi-step-system-prompt}
\end{figure}

\begin{figure}[!htp]
\centering
\begin{minipage}{\textwidth}
\begin{prompt}{Multi-Step First-Turn Prompt}
A customer has opened a refund claim for this product and is uploading their evidence photos one at a time. This is image {i} of {n}. Give a short forensic observation (2-3 sentences) of anything relevant to whether this image was produced or modified by AI -- either through AI image-editing LLMs or AI image-generation LLMs. Do not commit to a final verdict yet; more images from this claim are coming.
\end{prompt}
\end{minipage}
\caption{Prompt sent with the first image of a review in the multi-step setting.}
\label{code:multi-step-first-prompt}
\end{figure}

\begin{figure}[!htp]
\centering
\begin{minipage}{\textwidth}
\begin{prompt}{Multi-Step Intermediate-Turn Prompt}
Here is image {i} of {n} from the same refund claim. Taking your earlier observations into account, give a short forensic observation (2-3 sentences) of this image and say whether it changes your reading of the set so far. Do not commit to a final verdict yet; more images from this claim are coming.
\end{prompt}
\end{minipage}
\caption{Prompt sent with each subsequent non-final image in the multi-step setting.}
\label{code:multi-step-continuation-prompt}
\end{figure}

\begin{figure}[!htp]
\centering
\begin{minipage}{\textwidth}
\begin{prompt}{Multi-Step Final-Turn Prompt}
Here is image {n} of {n} -- the final image from this refund claim. Considering this image together with every image you have seen and every observation you made earlier in this conversation, decide whether these {n} images have been produced or modified by AI -- either through AI image-editing LLMs or through AI image-generation LLMs -- or whether they are unmodified genuine camera photographs of real scenes.

Return ONLY a single JSON object -- no markdown fences, no prose outside the object -- with this exact schema:
{
  "is_ai_modified": <true or false>,
  "confidence":     <number between 0 and 1>,
  "reason":         "<1-3 sentences explaining the combined verdict over all {n} images>",
  "per_image": [
    {
      "index":          <1-indexed position, 1..{n}>,
      "is_ai_modified": <true or false>,
      "confidence":     <number between 0 and 1>,
      "notes":          "<1-2 sentences on visual evidence from THIS image>"
    },
    ...
  ]
}
\end{prompt}
\end{minipage}
\caption{Prompt sent with the final image to obtain the consolidated verdict in the multi-step setting.}
\label{code:multi-step-final-prompt}
\end{figure}

Specifically, for Alibaba DashScope-hosted models, including the Qwen and QVQ models, the system prompt is prepended to the user turn as plain text rather than passed as a separate system-role message, following provider-specific usage constraints. 
All model responses are parsed against the required JSON schema. Responses that cannot be parsed against the schema are recorded as failed and re-queried under the same prompt template in a later pass, until every reported record has a valid response.

\newpage
\paragraph{Other Details.}
\label{appx:other_detail}
For some providers, multi-image and multi-step requests exceeded the request size limit, because high-resolution images inflate once base64-encoded. We handle these cases using a \textit{\textbf{deterministic image downscaling protocol}} rather than treating them as immediate failures. A request is downscaled when the provider rejects it as too large (HTTP~413, or HTTP~400 with a size-related message), when the connection of a request carrying at least 10~MB of image data is dropped twice, or when an Alibaba DashScope endpoint reports a transient failure to fetch the images (HTTP~400). Each such event re-encodes the images at the next of ten fixed tiers, progressively reducing the long-edge pixel limit and JPEG quality in tandem, from 2,560~px/Q95 down to 512~px/Q72. Other transient errors, such as rate limits and network failures, are retried with exponential back-off, so each request is attempted at most 13 times: three back-off retries plus ten downscale steps. Re-encoding is performed with Pillow using aspect-ratio-preserving resizing, and the resulting image is serialized as a base64 data URI before resubmission.

\subsubsection{Specialized Detectors}
\label{appx:specialized_detectors}
\paragraph{Models.} In addition to MLLM-based detectors, we evaluate four specialized AI-generated image detectors: CO-SPY~\citep{cheng2025co}, ForgeLens~\citep{chen2025forgelens}, Effort~\citep{yan2025orthogonal}, and IAPL~\citep{li2026towards}. For each method, we use the authors' official open-source implementation, linked as \url{https://github.com/Megum1/CO-SPY}, \url{https://github.com/Yingjian-Chen/ForgeLens}, \url{https://github.com/YZY-stack/Effort-AIGI-Detection}, and \url{https://github.com/liyih/IAPL}. We evaluate the released pretrained checkpoints provided by the corresponding repositories, including checkpoints trained on standard AI-generated image detection datasets such as GenImage, UniversalFakeDetect, Chameleon, and ProGAN. 

As these specialized detectors operate on individual images rather than interactive multimodal inputs, we report their results only in the single-image, review-free setting, excluding multi-image context, review text, multi-step interaction, and prompt-based reasoning instructions.

\paragraph{Testbed.}
\label{appx:testbed}
Experiments are conducted on a single Elastic Compute Service (ECS) instance. We employ an instance equipped with an Intel Xeon Platinum 8369B CPU with 16 available vCPUs, 125~GB RAM, 2048~GB of available disk space, and one NVIDIA A100 SXM GPU with 80~GB memory. This instance type is used for all specialized-detector evaluations to provide a consistent hardware environment across methods. For reproducibility and consistent performance, we recommend the \texttt{ecs.gn7e-c16g1.4xlarge} instance types on Alibaba Cloud.

We performed all experiments on Ubuntu~24.04.4 LTS. For the specialized detectors, we follow the official repository instructions and create a separate Conda environment for each method to configure the required Python, PyTorch, CUDA, and dependency versions. We therefore do not enforce a single shared software environment across repositories; instead, each detector is evaluated under the environment recommended by its released implementation. This setup avoids package-version conflicts while minimizing compatibility changes to the original codebases.

\paragraph{Image Forgery Detection and Localization Methods.} Since every fake-damaged image in \method{} is a local edit of a genuine photograph, we further evaluate image forgery detection and localization (IFDL) methods, which are designed to detect and localize spliced, copy-moved, and inpainted regions. We evaluate MVSS-Net~\citep{dong2022mvss} and the ResFCN baseline released in the same repository, each with its checkpoints trained on CASIA and DEFACTO, together with TruFor~\citep{guillaro2023trufor} and FakeShield~\citep{xu2025fakeshield}. For each method, we use the authors' official open-source implementation and released weights, linked as \url{https://github.com/dong03/MVSS-Net}, \url{https://github.com/grip-unina/TruFor}, and \url{https://github.com/zhipeixu/FakeShield}. Like the AI-generated image detectors above, IFDL methods operate on individual images, so we report their results only in the single-image, review-free setting on the same 6,940 images.

As for the AI-generated image detectors, we run the official inference scripts with \textit{\textbf{released checkpoints and default parameters}}, without fine-tuning or threshold calibration on our benchmark. For MVSS-Net and ResFCN, the image-level score is the maximum of the predicted manipulation map, following the official evaluation script, and for TruFor it is the image-level detection score. Both scores are thresholded at 0.5. FakeShield returns a textual verdict together with a manipulation mask for images it judges tampered, and we use its verdict directly.

\paragraph{Hyperparameters.} We run the official inference or evaluation scripts with \textit{\textbf{original configuration files and default parameters}}, without additional hyperparameter tuning or threshold calibration on our benchmark, to ensure that the comparison reflects each detector's standard off-the-shelf performance under the authors' recommended evaluation settings.

\subsubsection{Human Evaluations}
\label{appx:human_eval}

\paragraph{Evaluators.}
The human evaluation was conducted by three evaluators. The evaluators were not involved in curating the real evidence, generating the synthetic images, constructing the generation prompts, or selecting the generated samples used in the human evaluation subset. Their educational backgrounds include law, computer science, and engineering, with one evaluator holding a bachelor's degree in law and the other two holding undergraduate or master's degrees in computer science.

\paragraph{Task Design.}
Evaluators were asked to judge whether a product-review evidence image was AI-generated or manipulated. Each evaluation instance was presented through our self-built web-based evaluation interface, whose source code is available at~\url{https://huggingface.co/datasets/TristanYan/FraudBench}. The interface displayed a single image only, with no review text shown, even when such text was available in the original review. To maintain a blinded evaluation, the interface did not display the ground-truth label, generator identity, model predictions, source metadata, or prompt information. Evaluators provided a binary decision for each image.

\paragraph{Quality Control and Aggregation.}
Each image was independently labeled by three evaluators. We report human accuracy as the average of the three individual evaluator accuracies. Disagreements were not adjudicated, since the aggregate human baseline is computed from individual evaluator performance rather than from a consensus label.

\paragraph{Ethics.}
No external participants were recruited, and no compensation was involved. The evaluation required only binary judgments of product-review evidence images. It did not involve collecting personal information, demographic attributes, behavioral data, or sensitive information from the evaluators. Following our institutional policy, we determined that IRB approval was not required.

\subsubsection{Ablation Studies}
\label{app:ablation_study}

\paragraph{Prompt Sensitivity.}
We compare five prompt variants with progressively reduced task-specific guidance. 
The variants differ in the system-role framing, whether the forensic artifact checklist is retained, and whether the role instruction is separated from the task instruction. 
\textbf{Baseline} uses the full forensic prompt with an explicit artifact checklist. 
\textbf{Merged} uses the same content as Baseline but merges the system and user instructions into a single user message. 
\textbf{No Checklist} retains the forensic role while removing the artifact checklist. 
\textbf{Generic} replaces the forensic expert role with a generic assistant identity. 
\textbf{Minimal} uses only a one-sentence instruction with no system message.

All variants require a structured JSON response containing \texttt{is\_ai\_modified}, \texttt{confidence}, and \texttt{reason}. 
This keeps the output format comparable across variants and isolates the effect of prompt framing and task-specific guidance. 
Experiments are conducted in the single-image, review-free setting on \textit{\textbf{a stratified 10\% random sample}} of the Amazon subset, covering all 27 product categories and excluding the Trip.com and GrabFood service-review domains, using a fixed random seed of \texttt{42}.
\begin{figure}[!htp]
\centering
\begin{minipage}{\textwidth}
\begin{prompt}{Baseline Prompt}
System:

You are a forensic image analyst specialising in detecting images that have been produced or modified by AI -- either through AI image-editing LLMs or through AI image-generation LLMs. You look for characteristic traces such as local patch inconsistencies inside an otherwise plausible scene, object insertion or removal seams, mismatched shadows or reflections on specific objects, blurred or resampled regions around modified areas, copy-paste texture repetition, broken or re-drawn text and logos, anatomical or geometric implausibilities, over-smoothed textures, and diffusion-style artefacts. You never refuse the task and you always return valid JSON.

User:

Analyse the provided image and decide whether it has been produced or modified by AI -- either through AI image-editing LLMs or through AI image-generation LLMs -- or whether it is an unmodified genuine camera photograph of a real scene.

Return ONLY a single JSON object -- no markdown fences, no prose outside the object -- with this exact schema:
{
  "is_ai_modified": <true or false>,
  "confidence":     <number between 0 and 1>,
  "reason":         "<1-3 concise sentences naming the specific visual evidence you relied on>"
}
\end{prompt}
\end{minipage}
\caption{Baseline prompt used in the prompt-sensitivity study.}
\label{code:prompt-ablation-base}
\end{figure}

\begin{figure}[!htp]
\centering
\begin{minipage}{\textwidth}
\begin{prompt}{Merged Prompt}
System:

None

User:

You are a forensic image analyst specialising in detecting images that have been produced or modified by AI -- either through AI image-editing LLMs or through AI image-generation LLMs. You look for characteristic traces such as local patch inconsistencies inside an otherwise plausible scene, object insertion or removal seams, mismatched shadows or reflections on specific objects, blurred or resampled regions around modified areas, copy-paste texture repetition, broken or re-drawn text and logos, anatomical or geometric implausibilities, over-smoothed textures, and diffusion-style artefacts. You never refuse the task and you always return valid JSON.

Analyse the provided image and decide whether it has been produced or modified by AI -- either through AI image-editing LLMs or through AI image-generation LLMs -- or whether it is an unmodified genuine camera photograph of a real scene.

Return ONLY a single JSON object -- no markdown fences, no prose outside the object -- with this exact schema:
{
  "is_ai_modified": <true or false>,
  "confidence":     <number between 0 and 1>,
  "reason":         "<1-3 concise sentences naming the specific visual evidence you relied on>"
}
\end{prompt}
\end{minipage}
\caption{Merged prompt used in the prompt-sensitivity study. The system and user instructions are combined into a single user turn.}
\label{code:prompt-ablation-merged}
\end{figure}

\begin{figure}[!htp]
\centering
\begin{minipage}{\textwidth}
\begin{prompt}{No Checklist Prompt}
System:

You are a forensic image analyst specialising in detecting images that have been produced or modified by AI -- either through AI image-editing LLMs or through AI image-generation LLMs. You never refuse the task and you always return valid JSON.

User:

Analyse the provided image and decide whether it has been produced or modified by AI -- either through AI image-editing LLMs or through AI image-generation LLMs -- or whether it is an unmodified genuine camera photograph of a real scene.

Return ONLY a single JSON object -- no markdown fences, no prose outside the object -- with this exact schema:
{
  "is_ai_modified": <true or false>,
  "confidence":     <number between 0 and 1>,
  "reason":         "<1-3 concise sentences naming the specific visual evidence you relied on>"
}
\end{prompt}
\end{minipage}
\caption{No Checklist prompt used in the prompt-sensitivity study.}
\label{code:prompt-ablation-nochk}
\end{figure}

\begin{figure}[!htp]
\centering
\begin{minipage}{\textwidth}
\begin{prompt}{Generic Prompt}
System:

You are a helpful AI assistant. You always respond with valid JSON when asked to analyse an image.

User:

Analyse the provided image and decide whether it has been produced or modified by AI -- either through AI image-editing LLMs or through AI image-generation LLMs -- or whether it is an unmodified genuine camera photograph of a real scene.

Return ONLY a single JSON object -- no markdown fences, no prose outside the object -- with this exact schema:
{
  "is_ai_modified": <true or false>,
  "confidence":     <number between 0 and 1>,
  "reason":         "<1-3 concise sentences naming the specific visual evidence you relied on>"
}
\end{prompt}
\end{minipage}
\caption{Generic prompt used in the prompt-sensitivity study.}
\label{code:prompt-ablation-generic}
\end{figure}

\begin{figure}[!htp]
\centering
\begin{minipage}{\textwidth}
\begin{prompt}{Minimal Prompt}
System:

None

User:

Is this image a genuine photograph or was it produced / modified by an AI model? Return JSON only -- no markdown, no extra text:
{"is_ai_modified": <true or false>, "confidence": <0.0-1.0>, "reason": "<1-2 sentences>"}
\end{prompt}
\end{minipage}
\caption{Minimal prompt used in the prompt-sensitivity study.}
\label{code:prompt-ablation-minimal}
\end{figure}

\paragraph{Mismatched Review.}
The standard review-conditioned setting pairs each image with the customer review associated with the same product or service instance, allowing models to use both visual forensic cues and text-image consistency. To isolate the contribution of the review text, we introduce a mismatched-review condition on \textit{\textbf{a stratified 10\% random sample}} of the Amazon subset, covering all 27 product categories and excluding the Trip.com and GrabFood service-review domains, with a fixed random seed of \texttt{42}. For each image in this sample, we replace the matched review with a review sampled uniformly at random from a different product category.

For example, an AI-manipulated electronics image may be paired with a review originally written for a beauty product. 
The mismatched review is inserted at the same position in the prompt as in the standard review-conditioned setting, while the image input and all other prompt components are held fixed. Comparing the matched and mismatched conditions on the same image set, therefore, provides a controlled diagnostic of whether models rely on semantic coherence between the review text and visual content, rather than on visual forensic cues alone.

\subsubsection{Localization Evaluation}
\label{appx:localization_settings}

\paragraph{Ground Truth.} To score all methods on the same images, we use a ground truth that assigns a manipulated area to every fake-damaged image. After Gaussian smoothing of the absolute difference between a fake-damaged image and its real-undamaged source, we take the \textit{\textbf{smallest set of pixels that carries half of the total difference}} as the manipulated area. This area covers 17.8\% of the frame on average, more than the annotated regions in Appx.~\ref{appx:region_annotations}, since it also absorbs part of the diffuse residual left by re-rendering. To examine the effect of this choice, we additionally vary the fraction of the difference from 30\% to 70\%.

\paragraph{Models.} We compare methods that return a region. For Gemini 3 Flash, we issue a separate localization query for every image of the six evaluated generator subsets and the real-damaged subset, using the prompt in Fig.~\ref{code:localization-prompt}, which asks for a verdict, a confidence score, the tightest bounding box around the AI-modified content, and a short rationale. FakeShield returns a manipulation mask for every image it judges tampered, and MVSS-Net and ResFCN return a manipulation map, which we threshold at 0.5 (Appx.~\ref{appx:specialized_detectors}). TruFor retains only summary statistics of its localization map in our runs, so we do not score its localization. The localization metrics are defined in Appx.~\ref{sec:evaluation_metrics}.

\begin{figure}[!htp]
\centering
\begin{minipage}{\textwidth}
\begin{prompt}{Localization Prompt}
System:

You are a forensic image analyst. You localise regions of an image that have been produced or modified by AI image-editing or image-generation models. You never refuse the task and you always return valid JSON.

User:

Decide whether this image has been modified by AI, and if so, where.

Return ONLY a single JSON object with this exact schema:
{
  "is_ai_modified": <true or false>,
  "confidence": <number between 0 and 1>,
  "box_2d": [ymin, xmin, ymax, xmax],
  "reason": "<1-2 sentences naming the visual evidence>"
}

"box_2d" uses the standardS bounding-box convention: integers in [0,1000] normalised to the image size, ordered [ymin, xmin, ymax, xmax], with the origin at the top-left corner. Give the single tightest box containing the AI-modified content. If the image is not AI-modified, set "box_2d" to null.
\end{prompt}
\end{minipage}
\caption{Prompt used to query Gemini 3 Flash for the location of AI-modified content.}
\label{code:localization-prompt}
\end{figure}

\subsection{Evaluation Metrics}
\label{sec:evaluation_metrics}

\paragraph{Binary Verification.}
We formulate the task as binary verification. Fake-damaged images are treated as the positive class, while genuine, real-damaged images are treated as the negative class. For each evaluated model, the predicted label is obtained from the returned binary field \texttt{is\_ai\_modified}. The model-reported confidence score is used only for reporting confidence statistics and is not used to threshold or calibrate the binary prediction.

Let TP, FP, TN, and FN denote true positives, false positives, true negatives, and false negatives, respectively. In our setting, TP denotes a fake-damaged image correctly predicted as AI-generated or AI-modified, while TN denotes a genuine, real-damaged image correctly predicted as real. 
We define the base metrics as follows:
\begin{equation}
    \mathrm{Precision} = \frac{\mathrm{TP}}{\mathrm{TP}+\mathrm{FP}},
\end{equation}
\begin{equation}
    \mathrm{TPR} = \mathrm{Recall} = \frac{\mathrm{TP}}{\mathrm{TP}+\mathrm{FN}},
\end{equation}
\begin{equation}
    \mathrm{TNR} = \frac{\mathrm{TN}}{\mathrm{TN}+\mathrm{FP}}.
\end{equation}

\paragraph{Subset-Level Metrics.}
For each generator subset, we report TPR to measure the fraction of fake-damaged images correctly detected as AI-generated or AI-modified. 
The shared real-damaged subset is used as the negative reference set when computing metrics that require negative examples. 
On the real-damaged subset, we report TNR to measure the fraction of authentic damaged evidence correctly retained as genuine rather than falsely rejected as synthetic.

The reported \textbf{Conf.} value is the mean model-reported confidence over correctly classified images only. 
Thus, for generated subsets, \textbf{Conf.} is averaged over fake-damaged images that are correctly detected as AI-generated or AI-modified; for the real-damaged subset, it is averaged over genuine damaged images that are correctly classified as real.

\paragraph{Overall Aggregation.}
For overall comparison, we report Balanced Accuracy (Bal.Acc.) and F1. Both are macro-averaged over product categories: for each category, Bal.Acc. is the average of the category-level TPR, computed over all six generator subsets, and TNR, and F1 is the F1 score of the fake-damaged class. The category-level values are then averaged across categories, which prevents larger categories from dominating the overall scores.

\begin{equation}
    \mathrm{F1} =
    \frac{2 \cdot \mathrm{Precision} \cdot \mathrm{TPR}}
    {\mathrm{Precision}+\mathrm{TPR}},
\end{equation}
\begin{equation}
    \mathrm{Balanced\ Accuracy} =
    \frac{\mathrm{TPR}+\mathrm{TNR}}{2}.
\end{equation}

\paragraph{Threshold-Free Metrics.}
For the specialized detectors, which output a continuous score, we additionally report threshold-free metrics in Appx.~\ref{app:threshold_free}. AUROC measures how well the score ranks fake-damaged images above real-damaged images, and the partial AUROC for FPR $\leq 0.1$ is standardized so that 0.5 corresponds to chance. TPR@$k$\% denotes the TPR at the threshold whose false-positive rate on real-damaged evidence is $k$\%, which is equivalent to fixing the TNR at $1-k$\%. Unlike the other metrics, these are computed on the scores pooled over all categories and generators.

\paragraph{Localization Metrics.}
For the localization evaluation in Appx.~\ref{appx:localization_settings}, we rasterize every predicted region and the ground-truth manipulated area onto a common canvas, filling a bounding box as a rectangle. Let $P$ and $G$ denote the predicted and ground-truth pixel sets of a fake-damaged image. We compute
\begin{equation}
    \mathrm{IoU} = \frac{|P \cap G|}{|P \cup G|}, \qquad
    \mathrm{F1}_{\mathrm{pix}} = \frac{2\,|P \cap G|}{|P| + |G|}.
\end{equation}
FakeShield is scored on the images for which it returns a mask, and Gemini 3 Flash on those of them that it also flags with a box, whereas the thresholded maps of MVSS-Net and ResFCN are scored on every fake-damaged image. We report the mean and median IoU, the mean pixel F1, and the mean fraction of the frame covered by the prediction. On real-damaged evidence, where nothing was edited, any returned region counts as a false localization.

\subsection{Supplementary Evaluation Results}
\label{app:supp_eval}

Tab.~\ref{tab:T2} and~\ref{tab:T2b} report the single-image results of all MLLMs evaluated without and with review text, Tab.~\ref{tab:T3}--\ref{tab:T6} the multi-image and multi-step results, Tab.~\ref{tab:T3b} and~\ref{tab:T4b} the multi-image results of all 18 MLLMs evaluated on the 27 Amazon categories, Tab.~\ref{tab:T7}--\ref{tab:T35} the per-category results of all 30 MLLMs and the eight detector checkpoints, and Tab.~\ref{tab:human_eval} the human evaluation. Tables of the prompt-sensitivity and mismatch studies list the MLLMs of the corresponding study first and additional MLLMs evaluated with the same protocol in a separate block.

Before presenting the supplementary results, we clarify the metric aggregation and missing-value conventions used in the following tables. Unless otherwise specified, metrics are macro-averaged over product categories rather than computed from pooled predictions. For each category, generator-specific TPR is computed on the corresponding AI-generated subset, real-image TNR is computed on the shared real-damaged subset, and category-level F1 is computed for the fake-damaged class. The reported F1 is then averaged across categories.

TNR measures the fraction of genuinely damaged images correctly classified as unmodified, whereas TPR measures the fraction of AI-generated or AI-modified images correctly flagged as manipulated. 
F1 is the harmonic mean of precision and recall, Balanced Accuracy is the arithmetic mean of TPR and TNR, and Conf. denotes the mean model-reported confidence over correctly classified samples.

A cell is left blank, denoted by ``--'', when \textit{\textbf{the corresponding metric is undefined for a model-category combination}}. 
This occurs when a sampled category contains no images from a particular generator or when all API calls for that model and category fail. In practice, such blanks occur almost exclusively in generator-specific TPR columns for sparse 10\% ablation or mismatch-review subsets and do not affect the overall F1 or Balanced Accuracy columns, which aggregate valid predictions across generators within each category.

\begin{table}[!htp]
\centering
\caption{\textbf{Performance comparison of MLLMs on \method{}.} Results are reported across six AI image editing and generation subsets under the single-image setting without review context for all 30 evaluated MLLMs.}
\label{tab:T2}
\resizebox{\textwidth}{!}{
\begin{tabular}{c|cc cc cc cc cc cc|cc|cc}
\toprule
\multirow{2}[2]{*}{\textbf{Models}}
& \multicolumn{2}{c}{\textbf{GPT Image 2}}
& \multicolumn{2}{c}{\textbf{Grok Imagine}}
& \multicolumn{2}{c}{\textbf{Nano Banana 2}}
& \multicolumn{2}{c}{\textbf{Wan2.7-Image}}
& \multicolumn{2}{c}{\textbf{Qwen-Image-2.0}}
& \multicolumn{2}{c|}{\textbf{Qwen-Image-Edit}}
& \multicolumn{2}{c|}{\textbf{Overall}}
& \multicolumn{2}{c}{\textbf{Real-Damaged}} \\
\cmidrule(lr){2-3}
\cmidrule(lr){4-5}
\cmidrule(lr){6-7}
\cmidrule(lr){8-9}
\cmidrule(lr){10-11}
\cmidrule(lr){12-13}
\cmidrule(lr){14-15}
\cmidrule(lr){16-17}
& \textbf{TPR} & \textbf{Conf.}
& \textbf{TPR} & \textbf{Conf.}
& \textbf{TPR} & \textbf{Conf.}
& \textbf{TPR} & \textbf{Conf.}
& \textbf{TPR} & \textbf{Conf.}
& \textbf{TPR} & \textbf{Conf.}
& \textbf{Bal.Acc.} & \textbf{F1}
& \textbf{TNR} & \textbf{Conf.} \\
\midrule

\rowcolor{mypurple}
\multicolumn{17}{c}{\textbf{Multimodal Large Language Models}} \\
\midrule
DeepSeek-V4.1-Flash & 0.084 & 0.693 & 0.171 & 0.701 & 0.213 & 0.690 & 0.255 & 0.700 & 0.357 & 0.710 & 0.423 & 0.721 & 0.601 & 0.391 & 0.952 & 0.764\\
Gemini 3 Flash & 0.158 & 0.961 & 0.199 & 0.954 & 0.258 & 0.958 & 0.398 & 0.959 & \underline{0.491} & 0.957 & \underline{0.692} & 0.964 & \textbf{\textcolor{mypurple2}{0.674}} & \underline{0.526} & 0.982 & 0.968\\
Gemini 3.1 Pro & 0.091 & 0.714 & 0.122 & 0.712 & 0.167 & 0.712 & 0.260 & 0.719 & 0.327 & 0.715 & 0.519 & 0.717 & 0.619 & 0.391 & 0.990 & 0.704\\
Gemini 3.5 Flash-Lite & 0.050 & 0.965 & 0.066 & 0.958 & 0.085 & 0.964 & 0.119 & 0.968 & 0.178 & 0.965 & 0.244 & 0.968 & 0.560 & 0.212 & 0.996 & 0.950\\
Gemini 3.8 Flash & 0.025 & 0.703 & 0.013 & 0.708 & 0.066 & 0.703 & 0.102 & 0.708 & 0.157 & 0.707 & 0.199 & 0.710 & 0.547 & 0.165 & \textbf{\textcolor{mypurple2}{1.000}} & 0.658\\
\bottomrule
\end{tabular}
}
\end{table}
\begin{table*}[!htp]
\centering
\caption{\textbf{Performance comparison of MLLMs on \method{}.} Results are reported across six AI image editing and generation subsets under the single-image setting with review context for the 18 MLLMs evaluated in this setting.}
\label{tab:T2b}
\resizebox{\textwidth}{!}{
\begin{tabular}{c|cc cc cc cc cc cc|cc|cc}
\toprule
\multirow{2}[2]{*}{\textbf{Models}}
& \multicolumn{2}{c}{\textbf{GPT Image 2}}
& \multicolumn{2}{c}{\textbf{Grok Imagine}}
& \multicolumn{2}{c}{\textbf{Nano Banana 2}}
& \multicolumn{2}{c}{\textbf{Wan2.7-Image}}
& \multicolumn{2}{c}{\textbf{Qwen-Image-2.0}}
& \multicolumn{2}{c|}{\textbf{Qwen-Image-Edit}}
& \multicolumn{2}{c|}{\textbf{Overall}}
& \multicolumn{2}{c}{\textbf{Real-Damaged}} \\
\cmidrule(lr){2-3}
\cmidrule(lr){4-5}
\cmidrule(lr){6-7}
\cmidrule(lr){8-9}
\cmidrule(lr){10-11}
\cmidrule(lr){12-13}
\cmidrule(lr){14-15}
\cmidrule(lr){16-17}
& \textbf{TPR} & \textbf{Conf.}
& \textbf{TPR} & \textbf{Conf.}
& \textbf{TPR} & \textbf{Conf.}
& \textbf{TPR} & \textbf{Conf.}
& \textbf{TPR} & \textbf{Conf.}
& \textbf{TPR} & \textbf{Conf.}
& \textbf{Bal.Acc.} & \textbf{F1}
& \textbf{TNR} & \textbf{Conf.} \\
\midrule

\rowcolor{mypurple}
\multicolumn{17}{c}{\textbf{Multimodal Large Language Models}} \\
\midrule
DeepSeek-V4.1-Flash & 0.261 & 0.727 & 0.334 & 0.689 & 0.397 & 0.689 & 0.454 & 0.724 & 0.519 & 0.720 & 0.612 & 0.721 & 0.658 & 0.588 & 0.887 & 0.782\\
Gemini 3 Flash & 0.211 & 0.950 & 0.281 & 0.936 & 0.370 & 0.946 & 0.488 & 0.957 & 0.576 & 0.954 & 0.776 & 0.963 & 0.720 & 0.614 & 0.990 & 0.975\\
Gemini 3.8 Flash & 0.119 & 0.697 & 0.106 & 0.703 & 0.201 & 0.703 & 0.274 & 0.710 & 0.349 & 0.710 & 0.467 & 0.712 & 0.626 & 0.385 & 1.000 & 0.673\\
GPT-5.4 mini & 0.011 & 0.953 & 0.016 & 0.934 & 0.026 & 0.951 & 0.040 & 0.948 & 0.043 & 0.951 & 0.112 & 0.961 & 0.520 & 0.078 & 0.999 & 0.935\\
GPT-5.6 Terra & 0.069 & 0.780 & 0.086 & 0.801 & 0.119 & 0.807 & 0.295 & 0.870 & 0.275 & 0.843 & 0.458 & 0.864 & 0.608 & 0.346 & 0.999 & 0.964\\
\bottomrule
\end{tabular}
}
\end{table*}
\begin{table*}[!htp]
\centering
\caption{\textbf{Performance comparison of MLLMs on \method{}.} Results are reported across six AI image editing and generation subsets under the multi-image setting without review context for the eight MLLMs evaluated on all 29 categories in both multi-image settings.}
\label{tab:T3}
\resizebox{\textwidth}{!}{
\begin{tabular}{c|cc cc cc cc cc cc|cc|cc}
\toprule
\multirow{2}[2]{*}{\textbf{Models}}
& \multicolumn{2}{c}{\textbf{GPT Image 2}}
& \multicolumn{2}{c}{\textbf{Grok Imagine}}
& \multicolumn{2}{c}{\textbf{Nano Banana 2}}
& \multicolumn{2}{c}{\textbf{Wan2.7-Image}}
& \multicolumn{2}{c}{\textbf{Qwen-Image-2.0}}
& \multicolumn{2}{c|}{\textbf{Qwen-Image-Edit}}
& \multicolumn{2}{c|}{\textbf{Overall}}
& \multicolumn{2}{c}{\textbf{Real-Damaged}} \\
\cmidrule(lr){2-3}
\cmidrule(lr){4-5}
\cmidrule(lr){6-7}
\cmidrule(lr){8-9}
\cmidrule(lr){10-11}
\cmidrule(lr){12-13}
\cmidrule(lr){14-15}
\cmidrule(lr){16-17}
& \textbf{TPR} & \textbf{Conf.}
& \textbf{TPR} & \textbf{Conf.}
& \textbf{TPR} & \textbf{Conf.}
& \textbf{TPR} & \textbf{Conf.}
& \textbf{TPR} & \textbf{Conf.}
& \textbf{TPR} & \textbf{Conf.}
& \textbf{Bal.Acc.} & \textbf{F1}
& \textbf{TNR} & \textbf{Conf.} \\
\midrule

\rowcolor{mypurple}
\multicolumn{17}{c}{\textbf{Multimodal Large Language Models}} \\
\midrule
DeepSeek-V4.1-Flash & 0.067 & 0.680 & 0.216 & 0.695 & 0.271 & 0.721 & 0.279 & 0.716 & 0.429 & 0.739 & 0.621 & 0.749 & 0.637 & 0.463 & 0.960 & 0.808\\
Gemini 3.8 Flash & 0.050 & 0.717 & 0.065 & 0.714 & 0.153 & 0.718 & 0.268 & 0.718 & 0.367 & 0.721 & 0.467 & 0.721 & 0.612 & 0.353 & 1.000 & 0.685\\
GPT-5.6 Terra & 0.048 & 0.867 & 0.103 & 0.886 & 0.131 & 0.907 & 0.560 & 0.942 & 0.361 & 0.929 & 0.587 & 0.946 & 0.649 & 0.446 & 1.000 & 0.954\\
GPT-6 Astra & 0.279 & 0.879 & 0.294 & 0.878 & 0.418 & 0.902 & 0.684 & 0.909 & 0.692 & 0.921 & 0.833 & 0.927 & 0.773 & 0.679 & 1.000 & 0.959\\
Grok 4.6 & 0.063 & 0.843 & 0.025 & 0.828 & 0.102 & 0.827 & 0.221 & 0.849 & 0.195 & 0.838 & 0.311 & 0.841 & 0.570 & 0.253 & 1.000 & 0.874\\
\bottomrule
\end{tabular}
}
\end{table*}
\begin{table*}[!htp]
\centering
\caption{\textbf{Performance comparison of MLLMs on \method{}.} Results are reported across six AI image editing and generation subsets under the multi-image setting with review context for the eight MLLMs evaluated on all 29 categories in both multi-image settings.}
\label{tab:T4}
\resizebox{\textwidth}{!}{
\begin{tabular}{c|cc cc cc cc cc cc|cc|cc}
\toprule
\multirow{2}[2]{*}{\textbf{Models}}
& \multicolumn{2}{c}{\textbf{GPT Image 2}}
& \multicolumn{2}{c}{\textbf{Grok Imagine}}
& \multicolumn{2}{c}{\textbf{Nano Banana 2}}
& \multicolumn{2}{c}{\textbf{Wan2.7-Image}}
& \multicolumn{2}{c}{\textbf{Qwen-Image-2.0}}
& \multicolumn{2}{c|}{\textbf{Qwen-Image-Edit}}
& \multicolumn{2}{c|}{\textbf{Overall}}
& \multicolumn{2}{c}{\textbf{Real-Damaged}} \\
\cmidrule(lr){2-3}
\cmidrule(lr){4-5}
\cmidrule(lr){6-7}
\cmidrule(lr){8-9}
\cmidrule(lr){10-11}
\cmidrule(lr){12-13}
\cmidrule(lr){14-15}
\cmidrule(lr){16-17}
& \textbf{TPR} & \textbf{Conf.}
& \textbf{TPR} & \textbf{Conf.}
& \textbf{TPR} & \textbf{Conf.}
& \textbf{TPR} & \textbf{Conf.}
& \textbf{TPR} & \textbf{Conf.}
& \textbf{TPR} & \textbf{Conf.}
& \textbf{Bal.Acc.} & \textbf{F1}
& \textbf{TNR} & \textbf{Conf.} \\
\midrule

\rowcolor{mypurple}
\multicolumn{17}{c}{\textbf{Multimodal Large Language Models}} \\
\midrule
DeepSeek-V4.1-Flash & 0.327 & 0.707 & 0.364 & 0.735 & 0.438 & 0.727 & 0.572 & 0.739 & 0.633 & 0.759 & 0.789 & 0.759 & 0.748 & 0.671 & 0.980 & 0.803\\
Gemini 3.8 Flash & 0.321 & 0.719 & 0.278 & 0.717 & 0.411 & 0.720 & 0.550 & 0.715 & 0.561 & 0.718 & 0.782 & 0.721 & 0.742 & 0.633 & 1.000 & 0.687\\
GPT-5.6 Terra & 0.152 & 0.899 & 0.170 & 0.897 & 0.276 & 0.902 & 0.591 & 0.944 & 0.434 & 0.929 & 0.699 & 0.945 & 0.688 & 0.545 & 1.000 & 0.967\\
GPT-6 Astra & 0.746 & 0.858 & 0.768 & 0.862 & 0.824 & 0.881 & 0.917 & 0.909 & 0.954 & 0.921 & 0.983 & 0.938 & 0.930 & 0.925 & 1.000 & 0.957\\
Grok 4.6 & 0.068 & 0.851 & 0.037 & 0.814 & 0.124 & 0.841 & 0.294 & 0.855 & 0.232 & 0.862 & 0.366 & 0.849 & 0.591 & 0.303 & 1.000 & 0.898\\
Kimi K3 & 0.094 & 0.746 & 0.154 & 0.731 & 0.222 & 0.778 & 0.372 & 0.769 & 0.361 & 0.780 & 0.582 & 0.789 & 0.648 & 0.444 & 1.000 & 0.900\\
Qwen3.8-Flash & 0.127 & 0.844 & 0.225 & 0.837 & 0.304 & 0.844 & 0.475 & 0.865 & 0.517 & 0.861 & 0.597 & 0.881 & 0.680 & 0.531 & 0.985 & 0.922\\
Qwen3.8-Max & 0.108 & 0.824 & 0.180 & 0.817 & 0.238 & 0.828 & 0.472 & 0.853 & 0.521 & 0.861 & 0.709 & 0.890 & 0.689 & 0.523 & 1.000 & 0.915\\
\bottomrule
\end{tabular}
}
\end{table*}
\begin{table*}[!htp]
\centering
\caption{\textbf{Performance comparison of MLLMs on the 27 Amazon categories of \method{}.} Results are reported across six AI image editing and generation subsets under the multi-image setting without review context, macro-averaged over the 27 Amazon categories, for the 18 MLLMs evaluated on these categories in both multi-image settings. The first block repeats the MLLMs of Tab.~\ref{tab:T3} restricted to the same categories.}
\label{tab:T3b}
\resizebox{\textwidth}{!}{
\begin{tabular}{c|cc cc cc cc cc cc|cc|cc}
\toprule
\multirow{2}[2]{*}{\textbf{Models}}
& \multicolumn{2}{c}{\textbf{GPT Image 2}}
& \multicolumn{2}{c}{\textbf{Grok Imagine}}
& \multicolumn{2}{c}{\textbf{Nano Banana 2}}
& \multicolumn{2}{c}{\textbf{Wan2.7-Image}}
& \multicolumn{2}{c}{\textbf{Qwen-Image-2.0}}
& \multicolumn{2}{c|}{\textbf{Qwen-Image-Edit}}
& \multicolumn{2}{c|}{\textbf{Overall}}
& \multicolumn{2}{c}{\textbf{Real-Damaged}} \\
\cmidrule(lr){2-3}
\cmidrule(lr){4-5}
\cmidrule(lr){6-7}
\cmidrule(lr){8-9}
\cmidrule(lr){10-11}
\cmidrule(lr){12-13}
\cmidrule(lr){14-15}
\cmidrule(lr){16-17}
& \textbf{TPR} & \textbf{Conf.}
& \textbf{TPR} & \textbf{Conf.}
& \textbf{TPR} & \textbf{Conf.}
& \textbf{TPR} & \textbf{Conf.}
& \textbf{TPR} & \textbf{Conf.}
& \textbf{TPR} & \textbf{Conf.}
& \textbf{Bal.Acc.} & \textbf{F1}
& \textbf{TNR} & \textbf{Conf.} \\
\midrule

\rowcolor{mypurple}
\multicolumn{17}{c}{\textbf{Multimodal Large Language Models}} \\
\midrule
DeepSeek-V4.1-Flash & 0.067 & 0.681 & 0.215 & 0.695 & 0.278 & 0.723 & 0.281 & 0.709 & 0.423 & 0.740 & 0.621 & 0.751 & 0.636 & 0.464 & 0.958 & 0.808\\
Gemini 3.8 Flash & 0.049 & 0.720 & 0.054 & 0.713 & 0.144 & 0.718 & 0.260 & 0.718 & 0.359 & 0.722 & 0.461 & 0.721 & 0.608 & 0.347 & 1.000 & 0.685\\
GPT-5.6 Terra & 0.049 & 0.856 & 0.093 & 0.895 & 0.129 & 0.908 & 0.554 & 0.946 & 0.345 & 0.931 & 0.580 & 0.948 & 0.645 & 0.440 & 1.000 & 0.955\\
GPT-6 Astra & 0.278 & 0.884 & 0.271 & 0.882 & 0.408 & 0.907 & 0.673 & 0.910 & 0.692 & 0.922 & 0.824 & 0.929 & 0.769 & 0.671 & 1.000 & 0.960\\
Grok 4.6 & 0.065 & 0.843 & 0.027 & 0.828 & 0.104 & 0.827 & 0.224 & 0.851 & 0.198 & 0.839 & 0.309 & 0.845 & 0.571 & 0.255 & 1.000 & 0.875\\
\bottomrule
\end{tabular}
}
\end{table*}
\begin{table*}[!htp]
\centering
\caption{\textbf{Performance comparison of MLLMs on the 27 Amazon categories of \method{}.} Results are reported across six AI image editing and generation subsets under the multi-image setting with review context, macro-averaged over the 27 Amazon categories, for the 18 MLLMs evaluated on these categories in both multi-image settings. The first block repeats the MLLMs of Tab.~\ref{tab:T4} restricted to the same categories.}
\label{tab:T4b}
\resizebox{\textwidth}{!}{
\begin{tabular}{c|cc cc cc cc cc cc|cc|cc}
\toprule
\multirow{2}[2]{*}{\textbf{Models}}
& \multicolumn{2}{c}{\textbf{GPT Image 2}}
& \multicolumn{2}{c}{\textbf{Grok Imagine}}
& \multicolumn{2}{c}{\textbf{Nano Banana 2}}
& \multicolumn{2}{c}{\textbf{Wan2.7-Image}}
& \multicolumn{2}{c}{\textbf{Qwen-Image-2.0}}
& \multicolumn{2}{c|}{\textbf{Qwen-Image-Edit}}
& \multicolumn{2}{c|}{\textbf{Overall}}
& \multicolumn{2}{c}{\textbf{Real-Damaged}} \\
\cmidrule(lr){2-3}
\cmidrule(lr){4-5}
\cmidrule(lr){6-7}
\cmidrule(lr){8-9}
\cmidrule(lr){10-11}
\cmidrule(lr){12-13}
\cmidrule(lr){14-15}
\cmidrule(lr){16-17}
& \textbf{TPR} & \textbf{Conf.}
& \textbf{TPR} & \textbf{Conf.}
& \textbf{TPR} & \textbf{Conf.}
& \textbf{TPR} & \textbf{Conf.}
& \textbf{TPR} & \textbf{Conf.}
& \textbf{TPR} & \textbf{Conf.}
& \textbf{Bal.Acc.} & \textbf{F1}
& \textbf{TNR} & \textbf{Conf.} \\
\midrule

\rowcolor{mypurple}
\multicolumn{17}{c}{\textbf{Multimodal Large Language Models}} \\
\midrule
DeepSeek-V4.1-Flash & 0.335 & 0.707 & 0.365 & 0.734 & 0.440 & 0.728 & 0.576 & 0.739 & 0.631 & 0.758 & 0.788 & 0.759 & 0.748 & 0.674 & 0.978 & 0.804\\
Gemini 3.8 Flash & 0.323 & 0.719 & 0.263 & 0.718 & 0.405 & 0.720 & 0.544 & 0.716 & 0.559 & 0.718 & 0.781 & 0.721 & 0.739 & 0.630 & 1.000 & 0.688\\
GPT-5.6 Terra & 0.145 & 0.907 & 0.152 & 0.902 & 0.270 & 0.904 & 0.586 & 0.944 & 0.428 & 0.932 & 0.698 & 0.949 & 0.684 & 0.540 & 1.000 & 0.968\\
GPT-6 Astra & 0.748 & 0.863 & 0.760 & 0.865 & 0.828 & 0.885 & 0.911 & 0.911 & 0.959 & 0.923 & 0.982 & 0.939 & 0.930 & 0.924 & 1.000 & 0.958\\
Grok 4.6 & 0.070 & 0.849 & 0.038 & 0.824 & 0.128 & 0.841 & 0.302 & 0.855 & 0.230 & 0.863 & 0.366 & 0.850 & 0.592 & 0.307 & 1.000 & 0.898\\
\bottomrule
\end{tabular}
}
\end{table*}
\begin{table*}[!htp]
\centering
\caption{\textbf{Performance comparison of MLLMs on \method{}.} Results are reported across six AI image editing and generation subsets under the multi-step setting without review context, in which the images of a review are presented one per turn in a multi-turn conversation. Results are shown for the six MLLMs evaluated in this setting.}
\label{tab:T5}
\resizebox{\textwidth}{!}{
\begin{tabular}{c|cc cc cc cc cc cc|cc|cc}
\toprule
\multirow{2}[2]{*}{\textbf{Models}}
& \multicolumn{2}{c}{\textbf{GPT Image 2}}
& \multicolumn{2}{c}{\textbf{Grok Imagine}}
& \multicolumn{2}{c}{\textbf{Nano Banana 2}}
& \multicolumn{2}{c}{\textbf{Wan2.7-Image}}
& \multicolumn{2}{c}{\textbf{Qwen-Image-2.0}}
& \multicolumn{2}{c|}{\textbf{Qwen-Image-Edit}}
& \multicolumn{2}{c|}{\textbf{Overall}}
& \multicolumn{2}{c}{\textbf{Real-Damaged}} \\
\cmidrule(lr){2-3}
\cmidrule(lr){4-5}
\cmidrule(lr){6-7}
\cmidrule(lr){8-9}
\cmidrule(lr){10-11}
\cmidrule(lr){12-13}
\cmidrule(lr){14-15}
\cmidrule(lr){16-17}
& \textbf{TPR} & \textbf{Conf.}
& \textbf{TPR} & \textbf{Conf.}
& \textbf{TPR} & \textbf{Conf.}
& \textbf{TPR} & \textbf{Conf.}
& \textbf{TPR} & \textbf{Conf.}
& \textbf{TPR} & \textbf{Conf.}
& \textbf{Bal.Acc.} & \textbf{F1}
& \textbf{TNR} & \textbf{Conf.} \\
\midrule

\rowcolor{mypurple}
\multicolumn{17}{c}{\textbf{Multimodal Large Language Models}} \\
\midrule
DeepSeek-V4.1-Flash & 0.371 & 0.785 & 0.549 & 0.790 & 0.629 & 0.789 & 0.634 & 0.818 & 0.739 & 0.840 & 0.870 & 0.853 & 0.762 & 0.753 & 0.905 & 0.768\\
Gemini 3.8 Flash & 0.221 & 0.735 & 0.245 & 0.734 & 0.399 & 0.740 & 0.533 & 0.733 & 0.608 & 0.741 & 0.759 & 0.745 & 0.724 & 0.613 & 0.995 & 0.820\\
GPT-5.6 Terra & 0.044 & 0.824 & 0.072 & 0.854 & 0.104 & 0.859 & 0.365 & 0.901 & 0.275 & 0.856 & 0.469 & 0.922 & 0.612 & 0.348 & 1.000 & 0.908\\
Grok 4.6 & 0.139 & 0.782 & 0.116 & 0.774 & 0.255 & 0.799 & 0.380 & 0.829 & 0.378 & 0.826 & 0.481 & 0.834 & 0.634 & 0.429 & 0.996 & 0.868\\
Kimi K3 & 0.056 & 0.758 & 0.126 & 0.731 & 0.209 & 0.799 & 0.419 & 0.780 & 0.350 & 0.804 & 0.612 & 0.818 & 0.649 & 0.441 & 0.999 & 0.881\\
Qwen3.8-Max & 0.339 & 0.823 & 0.469 & 0.837 & 0.624 & 0.839 & 0.778 & 0.878 & 0.723 & 0.875 & 0.930 & 0.896 & 0.818 & 0.772 & 0.987 & 0.893\\
\bottomrule
\end{tabular}
}
\end{table*}
\begin{table*}[!htp]
\centering
\caption{\textbf{Performance comparison of MLLMs on \method{}.} Results are reported across six AI image editing and generation subsets under the multi-step setting with review context, in which the images of a review are presented one per turn in a multi-turn conversation. Results are shown for the six MLLMs evaluated in this setting.}
\label{tab:T6}
\resizebox{\textwidth}{!}{
\begin{tabular}{c|cc cc cc cc cc cc|cc|cc}
\toprule
\multirow{2}[2]{*}{\textbf{Models}}
& \multicolumn{2}{c}{\textbf{GPT Image 2}}
& \multicolumn{2}{c}{\textbf{Grok Imagine}}
& \multicolumn{2}{c}{\textbf{Nano Banana 2}}
& \multicolumn{2}{c}{\textbf{Wan2.7-Image}}
& \multicolumn{2}{c}{\textbf{Qwen-Image-2.0}}
& \multicolumn{2}{c|}{\textbf{Qwen-Image-Edit}}
& \multicolumn{2}{c|}{\textbf{Overall}}
& \multicolumn{2}{c}{\textbf{Real-Damaged}} \\
\cmidrule(lr){2-3}
\cmidrule(lr){4-5}
\cmidrule(lr){6-7}
\cmidrule(lr){8-9}
\cmidrule(lr){10-11}
\cmidrule(lr){12-13}
\cmidrule(lr){14-15}
\cmidrule(lr){16-17}
& \textbf{TPR} & \textbf{Conf.}
& \textbf{TPR} & \textbf{Conf.}
& \textbf{TPR} & \textbf{Conf.}
& \textbf{TPR} & \textbf{Conf.}
& \textbf{TPR} & \textbf{Conf.}
& \textbf{TPR} & \textbf{Conf.}
& \textbf{Bal.Acc.} & \textbf{F1}
& \textbf{TNR} & \textbf{Conf.} \\
\midrule

\rowcolor{mypurple}
\multicolumn{17}{c}{\textbf{Multimodal Large Language Models}} \\
\midrule
DeepSeek-V4.1-Flash & 0.513 & 0.788 & 0.737 & 0.792 & 0.773 & 0.816 & 0.751 & 0.830 & 0.867 & 0.835 & 0.934 & 0.851 & 0.816 & 0.849 & 0.878 & 0.788\\
Gemini 3.8 Flash & 0.436 & 0.764 & 0.400 & 0.778 & 0.544 & 0.769 & 0.694 & 0.748 & 0.733 & 0.765 & 0.869 & 0.748 & 0.806 & 0.747 & 1.000 & 0.869\\
GPT-5.6 Terra & 0.050 & 0.836 & 0.068 & 0.850 & 0.132 & 0.875 & 0.418 & 0.887 & 0.317 & 0.877 & 0.531 & 0.911 & 0.625 & 0.391 & 1.000 & 0.927\\
Grok 4.6 & 0.218 & 0.805 & 0.184 & 0.797 & 0.348 & 0.814 & 0.555 & 0.856 & 0.564 & 0.826 & 0.655 & 0.842 & 0.707 & 0.579 & 0.997 & 0.867\\
Kimi K3 & 0.177 & 0.738 & 0.240 & 0.721 & 0.322 & 0.780 & 0.493 & 0.790 & 0.509 & 0.816 & 0.795 & 0.818 & 0.713 & 0.580 & 1.000 & 0.902\\
Qwen3.8-Max & 0.389 & 0.841 & 0.526 & 0.823 & 0.619 & 0.850 & 0.797 & 0.879 & 0.756 & 0.882 & 0.883 & 0.905 & 0.835 & 0.785 & 0.997 & 0.903\\
\bottomrule
\end{tabular}
}
\end{table*}
\begin{table*}[!htp]
\centering
\caption{\textbf{Performance comparison of MLLMs on the All Beauty subset of \method{}.} Results are reported across six AI image editing and generation subsets under the single-image setting without review context.}
\label{tab:T7}
\resizebox{\textwidth}{!}{
\begin{tabular}{c|cc cc cc cc cc cc|cc|cc}
\toprule
\multirow{2}[2]{*}{\textbf{Models}}
& \multicolumn{2}{c}{\textbf{GPT Image 2}}
& \multicolumn{2}{c}{\textbf{Grok Imagine}}
& \multicolumn{2}{c}{\textbf{Nano Banana 2}}
& \multicolumn{2}{c}{\textbf{Wan2.7-Image}}
& \multicolumn{2}{c}{\textbf{Qwen-Image-2.0}}
& \multicolumn{2}{c|}{\textbf{Qwen-Image-Edit}}
& \multicolumn{2}{c|}{\textbf{Overall}}
& \multicolumn{2}{c}{\textbf{Real-Damaged}} \\
\cmidrule(lr){2-3}
\cmidrule(lr){4-5}
\cmidrule(lr){6-7}
\cmidrule(lr){8-9}
\cmidrule(lr){10-11}
\cmidrule(lr){12-13}
\cmidrule(lr){14-15}
\cmidrule(lr){16-17}
& \textbf{TPR} & \textbf{Conf.}
& \textbf{TPR} & \textbf{Conf.}
& \textbf{TPR} & \textbf{Conf.}
& \textbf{TPR} & \textbf{Conf.}
& \textbf{TPR} & \textbf{Conf.}
& \textbf{TPR} & \textbf{Conf.}
& \textbf{Bal.Acc.} & \textbf{F1}
& \textbf{TNR} & \textbf{Conf.} \\
\midrule

\rowcolor{mypurple}
\multicolumn{17}{c}{\textbf{Multimodal Large Language Models}} \\
\midrule
DeepSeek-V4.1-Flash & 0.179 & 0.724 & 0.286 & 0.756 & 0.393 & 0.681 & 0.464 & 0.713 & 0.393 & 0.747 & 0.607 & 0.755 & 0.649 & 0.551 & 0.912 & 0.755\\
Gemini 3.1 Pro & 0.179 & 0.730 & 0.429 & 0.726 & 0.321 & 0.721 & 0.357 & 0.720 & 0.500 & 0.719 & 0.857 & 0.718 & 0.691 & 0.607 & 0.941 & 0.710\\
Gemini 3.5 Flash-Lite & 0.071 & 0.970 & 0.214 & 0.962 & 0.143 & 0.967 & 0.250 & 0.964 & 0.179 & 0.964 & 0.429 & 0.967 & 0.607 & 0.353 & 1.000 & 0.951\\
Gemini 3.8 Flash & 0.143 & 0.710 & 0.000 & — & 0.107 & 0.707 & 0.179 & 0.696 & 0.143 & 0.720 & 0.286 & 0.710 & 0.571 & 0.250 & 1.000 & 0.611\\
GPT-6 Astra & 0.107 & 0.787 & 0.036 & 0.740 & 0.179 & 0.822 & 0.321 & 0.867 & 0.357 & 0.835 & 0.500 & 0.846 & 0.625 & 0.400 & 1.000 & 0.939\\
\bottomrule
\end{tabular}
}
\end{table*}
\begin{table*}[!htp]
\centering
\caption{\textbf{Performance comparison of MLLMs on the Amazon Fashion subset of \method{}.} Results are reported across six AI image editing and generation subsets under the single-image setting without review context.}
\label{tab:T8}
\resizebox{\textwidth}{!}{
\begin{tabular}{c|cc cc cc cc cc cc|cc|cc}
\toprule
\multirow{2}[2]{*}{\textbf{Models}}
& \multicolumn{2}{c}{\textbf{GPT Image 2}}
& \multicolumn{2}{c}{\textbf{Grok Imagine}}
& \multicolumn{2}{c}{\textbf{Nano Banana 2}}
& \multicolumn{2}{c}{\textbf{Wan2.7-Image}}
& \multicolumn{2}{c}{\textbf{Qwen-Image-2.0}}
& \multicolumn{2}{c|}{\textbf{Qwen-Image-Edit}}
& \multicolumn{2}{c|}{\textbf{Overall}}
& \multicolumn{2}{c}{\textbf{Real-Damaged}} \\
\cmidrule(lr){2-3}
\cmidrule(lr){4-5}
\cmidrule(lr){6-7}
\cmidrule(lr){8-9}
\cmidrule(lr){10-11}
\cmidrule(lr){12-13}
\cmidrule(lr){14-15}
\cmidrule(lr){16-17}
& \textbf{TPR} & \textbf{Conf.}
& \textbf{TPR} & \textbf{Conf.}
& \textbf{TPR} & \textbf{Conf.}
& \textbf{TPR} & \textbf{Conf.}
& \textbf{TPR} & \textbf{Conf.}
& \textbf{TPR} & \textbf{Conf.}
& \textbf{Bal.Acc.} & \textbf{F1}
& \textbf{TNR} & \textbf{Conf.} \\
\midrule

\rowcolor{mypurple}
\multicolumn{17}{c}{\textbf{Multimodal Large Language Models}} \\
\midrule
DeepSeek-V4.1-Flash & 0.029 & 0.620 & 0.143 & 0.726 & 0.114 & 0.700 & 0.057 & 0.660 & 0.343 & 0.740 & 0.514 & 0.707 & 0.600 & 0.333 & 1.000 & 0.754\\
Gemini 3.1 Pro & 0.114 & 0.730 & 0.143 & 0.694 & 0.086 & 0.740 & 0.257 & 0.722 & 0.314 & 0.710 & 0.600 & 0.716 & 0.597 & 0.402 & 0.941 & 0.708\\
Gemini 3.5 Flash-Lite & 0.029 & 0.950 & 0.000 & — & 0.029 & 0.950 & 0.029 & 0.990 & 0.057 & 0.965 & 0.114 & 0.960 & 0.521 & 0.082 & 1.000 & 0.952\\
Gemini 3.8 Flash & 0.000 & — & 0.000 & — & 0.029 & 0.680 & 0.000 & — & 0.057 & 0.680 & 0.114 & 0.690 & 0.517 & 0.065 & 1.000 & 0.651\\
GPT-6 Astra & 0.057 & 0.760 & 0.029 & 0.760 & 0.143 & 0.796 & 0.314 & 0.800 & 0.371 & 0.815 & 0.514 & 0.802 & 0.619 & 0.385 & 1.000 & 0.938\\
\bottomrule
\end{tabular}
}
\end{table*}
\begin{table*}[!htp]
\centering
\caption{\textbf{Performance comparison of MLLMs on the Appliances subset of \method{}.} Results are reported across six AI image editing and generation subsets under the single-image setting without review context.}
\label{tab:T9}
\resizebox{\textwidth}{!}{
\begin{tabular}{c|cc cc cc cc cc cc|cc|cc}
\toprule
\multirow{2}[2]{*}{\textbf{Models}}
& \multicolumn{2}{c}{\textbf{GPT Image 2}}
& \multicolumn{2}{c}{\textbf{Grok Imagine}}
& \multicolumn{2}{c}{\textbf{Nano Banana 2}}
& \multicolumn{2}{c}{\textbf{Wan2.7-Image}}
& \multicolumn{2}{c}{\textbf{Qwen-Image-2.0}}
& \multicolumn{2}{c|}{\textbf{Qwen-Image-Edit}}
& \multicolumn{2}{c|}{\textbf{Overall}}
& \multicolumn{2}{c}{\textbf{Real-Damaged}} \\
\cmidrule(lr){2-3}
\cmidrule(lr){4-5}
\cmidrule(lr){6-7}
\cmidrule(lr){8-9}
\cmidrule(lr){10-11}
\cmidrule(lr){12-13}
\cmidrule(lr){14-15}
\cmidrule(lr){16-17}
& \textbf{TPR} & \textbf{Conf.}
& \textbf{TPR} & \textbf{Conf.}
& \textbf{TPR} & \textbf{Conf.}
& \textbf{TPR} & \textbf{Conf.}
& \textbf{TPR} & \textbf{Conf.}
& \textbf{TPR} & \textbf{Conf.}
& \textbf{Bal.Acc.} & \textbf{F1}
& \textbf{TNR} & \textbf{Conf.} \\
\midrule

\rowcolor{mypurple}
\multicolumn{17}{c}{\textbf{Multimodal Large Language Models}} \\
\midrule
DeepSeek-V4.1-Flash & 0.038 & 0.600 & 0.154 & 0.705 & 0.231 & 0.657 & 0.423 & 0.687 & 0.577 & 0.743 & 0.346 & 0.738 & 0.596 & 0.447 & 0.897 & 0.748\\
Gemini 3.1 Pro & 0.038 & 0.700 & 0.000 & — & 0.077 & 0.685 & 0.192 & 0.724 & 0.423 & 0.699 & 0.423 & 0.715 & 0.596 & 0.323 & 1.000 & 0.706\\
Gemini 3.5 Flash-Lite & 0.000 & — & 0.000 & — & 0.000 & — & 0.077 & 0.950 & 0.077 & 0.965 & 0.192 & 0.972 & 0.529 & 0.109 & 1.000 & 0.949\\
Gemini 3.8 Flash & 0.077 & 0.700 & 0.000 & — & 0.154 & 0.705 & 0.231 & 0.700 & 0.269 & 0.697 & 0.192 & 0.708 & 0.577 & 0.267 & 1.000 & 0.702\\
GPT-6 Astra & 0.077 & 0.765 & 0.077 & 0.830 & 0.346 & 0.822 & 0.615 & 0.848 & 0.731 & 0.847 & 0.692 & 0.845 & 0.712 & 0.595 & 1.000 & 0.929\\
\bottomrule
\end{tabular}
}
\end{table*}
\begin{table*}[!htp]
\centering
\caption{\textbf{Performance comparison of MLLMs on the Arts, Crafts \& Sewing subset of \method{}.} Results are reported across six AI image editing and generation subsets under the single-image setting without review context.}
\label{tab:T10}
\resizebox{\textwidth}{!}{
\begin{tabular}{c|cc cc cc cc cc cc|cc|cc}
\toprule
\multirow{2}[2]{*}{\textbf{Models}}
& \multicolumn{2}{c}{\textbf{GPT Image 2}}
& \multicolumn{2}{c}{\textbf{Grok Imagine}}
& \multicolumn{2}{c}{\textbf{Nano Banana 2}}
& \multicolumn{2}{c}{\textbf{Wan2.7-Image}}
& \multicolumn{2}{c}{\textbf{Qwen-Image-2.0}}
& \multicolumn{2}{c|}{\textbf{Qwen-Image-Edit}}
& \multicolumn{2}{c|}{\textbf{Overall}}
& \multicolumn{2}{c}{\textbf{Real-Damaged}} \\
\cmidrule(lr){2-3}
\cmidrule(lr){4-5}
\cmidrule(lr){6-7}
\cmidrule(lr){8-9}
\cmidrule(lr){10-11}
\cmidrule(lr){12-13}
\cmidrule(lr){14-15}
\cmidrule(lr){16-17}
& \textbf{TPR} & \textbf{Conf.}
& \textbf{TPR} & \textbf{Conf.}
& \textbf{TPR} & \textbf{Conf.}
& \textbf{TPR} & \textbf{Conf.}
& \textbf{TPR} & \textbf{Conf.}
& \textbf{TPR} & \textbf{Conf.}
& \textbf{Bal.Acc.} & \textbf{F1}
& \textbf{TNR} & \textbf{Conf.} \\
\midrule

\rowcolor{mypurple}
\multicolumn{17}{c}{\textbf{Multimodal Large Language Models}} \\
\midrule
DeepSeek-V4.1-Flash & 0.150 & 0.763 & 0.300 & 0.633 & 0.350 & 0.733 & 0.400 & 0.734 & 0.350 & 0.790 & 0.400 & 0.771 & 0.662 & 0.491 & 1.000 & 0.776\\
Gemini 3.1 Pro & 0.150 & 0.717 & 0.300 & 0.720 & 0.150 & 0.667 & 0.250 & 0.718 & 0.350 & 0.709 & 0.500 & 0.721 & 0.642 & 0.442 & 1.000 & 0.712\\
Gemini 3.5 Flash-Lite & 0.150 & 0.950 & 0.250 & 0.966 & 0.150 & 0.963 & 0.150 & 0.977 & 0.400 & 0.974 & 0.400 & 0.978 & 0.625 & 0.400 & 1.000 & 0.950\\
Gemini 3.8 Flash & 0.000 & — & 0.000 & — & 0.050 & 0.680 & 0.000 & — & 0.100 & 0.710 & 0.100 & 0.730 & 0.521 & 0.080 & 1.000 & 0.686\\
GPT-6 Astra & 0.000 & — & 0.050 & 0.780 & 0.200 & 0.835 & 0.300 & 0.813 & 0.250 & 0.810 & 0.300 & 0.848 & 0.592 & 0.310 & 1.000 & 0.918\\
\bottomrule
\end{tabular}
}
\end{table*}
\begin{table*}[!htp]
\centering
\caption{\textbf{Performance comparison of MLLMs on the Automotive subset of \method{}.} Results are reported across six AI image editing and generation subsets under the single-image setting without review context.}
\label{tab:T11}
\resizebox{\textwidth}{!}{
\begin{tabular}{c|cc cc cc cc cc cc|cc|cc}
\toprule
\multirow{2}[2]{*}{\textbf{Models}}
& \multicolumn{2}{c}{\textbf{GPT Image 2}}
& \multicolumn{2}{c}{\textbf{Grok Imagine}}
& \multicolumn{2}{c}{\textbf{Nano Banana 2}}
& \multicolumn{2}{c}{\textbf{Wan2.7-Image}}
& \multicolumn{2}{c}{\textbf{Qwen-Image-2.0}}
& \multicolumn{2}{c|}{\textbf{Qwen-Image-Edit}}
& \multicolumn{2}{c|}{\textbf{Overall}}
& \multicolumn{2}{c}{\textbf{Real-Damaged}} \\
\cmidrule(lr){2-3}
\cmidrule(lr){4-5}
\cmidrule(lr){6-7}
\cmidrule(lr){8-9}
\cmidrule(lr){10-11}
\cmidrule(lr){12-13}
\cmidrule(lr){14-15}
\cmidrule(lr){16-17}
& \textbf{TPR} & \textbf{Conf.}
& \textbf{TPR} & \textbf{Conf.}
& \textbf{TPR} & \textbf{Conf.}
& \textbf{TPR} & \textbf{Conf.}
& \textbf{TPR} & \textbf{Conf.}
& \textbf{TPR} & \textbf{Conf.}
& \textbf{Bal.Acc.} & \textbf{F1}
& \textbf{TNR} & \textbf{Conf.} \\
\midrule

\rowcolor{mypurple}
\multicolumn{17}{c}{\textbf{Multimodal Large Language Models}} \\
\midrule
DeepSeek-V4.1-Flash & 0.034 & 0.620 & 0.241 & 0.669 & 0.138 & 0.655 & 0.207 & 0.683 & 0.241 & 0.711 & 0.517 & 0.689 & 0.615 & 0.374 & 1.000 & 0.757\\
Gemini 3.1 Pro & 0.034 & 0.700 & 0.034 & 0.700 & 0.069 & 0.725 & 0.207 & 0.723 & 0.207 & 0.710 & 0.517 & 0.718 & 0.589 & 0.302 & 1.000 & 0.699\\
Gemini 3.5 Flash-Lite & 0.000 & — & 0.103 & 0.970 & 0.069 & 0.965 & 0.069 & 0.965 & 0.138 & 0.975 & 0.241 & 0.971 & 0.552 & 0.188 & 1.000 & 0.951\\
Gemini 3.8 Flash & 0.034 & 0.720 & 0.034 & 0.720 & 0.034 & 0.700 & 0.034 & 0.680 & 0.172 & 0.704 & 0.276 & 0.710 & 0.549 & 0.178 & 1.000 & 0.664\\
GPT-6 Astra & 0.138 & 0.770 & 0.207 & 0.805 & 0.207 & 0.835 & 0.655 & 0.811 & 0.655 & 0.843 & 0.759 & 0.845 & 0.718 & 0.608 & 1.000 & 0.933\\
\bottomrule
\end{tabular}
}
\end{table*}
\begin{table*}[!htp]
\centering
\caption{\textbf{Performance comparison of MLLMs on the Baby Products subset of \method{}.} Results are reported across six AI image editing and generation subsets under the single-image setting without review context.}
\label{tab:T12}
\resizebox{\textwidth}{!}{
\begin{tabular}{c|cc cc cc cc cc cc|cc|cc}
\toprule
\multirow{2}[2]{*}{\textbf{Models}}
& \multicolumn{2}{c}{\textbf{GPT Image 2}}
& \multicolumn{2}{c}{\textbf{Grok Imagine}}
& \multicolumn{2}{c}{\textbf{Nano Banana 2}}
& \multicolumn{2}{c}{\textbf{Wan2.7-Image}}
& \multicolumn{2}{c}{\textbf{Qwen-Image-2.0}}
& \multicolumn{2}{c|}{\textbf{Qwen-Image-Edit}}
& \multicolumn{2}{c|}{\textbf{Overall}}
& \multicolumn{2}{c}{\textbf{Real-Damaged}} \\
\cmidrule(lr){2-3}
\cmidrule(lr){4-5}
\cmidrule(lr){6-7}
\cmidrule(lr){8-9}
\cmidrule(lr){10-11}
\cmidrule(lr){12-13}
\cmidrule(lr){14-15}
\cmidrule(lr){16-17}
& \textbf{TPR} & \textbf{Conf.}
& \textbf{TPR} & \textbf{Conf.}
& \textbf{TPR} & \textbf{Conf.}
& \textbf{TPR} & \textbf{Conf.}
& \textbf{TPR} & \textbf{Conf.}
& \textbf{TPR} & \textbf{Conf.}
& \textbf{Bal.Acc.} & \textbf{F1}
& \textbf{TNR} & \textbf{Conf.} \\
\midrule

\rowcolor{mypurple}
\multicolumn{17}{c}{\textbf{Multimodal Large Language Models}} \\
\midrule
DeepSeek-V4.1-Flash & 0.032 & 0.700 & 0.000 & — & 0.097 & 0.630 & 0.065 & 0.610 & 0.226 & 0.689 & 0.226 & 0.703 & 0.545 & 0.193 & 0.982 & 0.783\\
Gemini 3.1 Pro & 0.000 & — & 0.032 & 0.720 & 0.000 & — & 0.129 & 0.710 & 0.000 & — & 0.226 & 0.721 & 0.523 & 0.121 & 0.982 & 0.699\\
Gemini 3.5 Flash-Lite & 0.000 & — & 0.000 & — & 0.032 & 0.950 & 0.000 & — & 0.032 & 0.950 & 0.032 & 0.980 & 0.508 & 0.032 & 1.000 & 0.951\\
Gemini 3.8 Flash & 0.000 & — & 0.000 & — & 0.000 & — & 0.000 & — & 0.000 & — & 0.065 & 0.720 & 0.505 & 0.021 & 1.000 & 0.665\\
GPT-6 Astra & 0.032 & 0.810 & 0.032 & 0.770 & 0.065 & 0.755 & 0.290 & 0.810 & 0.258 & 0.779 & 0.452 & 0.819 & 0.594 & 0.317 & 1.000 & 0.932\\
\bottomrule
\end{tabular}
}
\end{table*}
\begin{table*}[!htp]
\centering
\caption{\textbf{Performance comparison of MLLMs on the Beauty \& Personal Care subset of \method{}.} Results are reported across six AI image editing and generation subsets under the single-image setting without review context.}
\label{tab:T13}
\resizebox{\textwidth}{!}{
\begin{tabular}{c|cc cc cc cc cc cc|cc|cc}
\toprule
\multirow{2}[2]{*}{\textbf{Models}}
& \multicolumn{2}{c}{\textbf{GPT Image 2}}
& \multicolumn{2}{c}{\textbf{Grok Imagine}}
& \multicolumn{2}{c}{\textbf{Nano Banana 2}}
& \multicolumn{2}{c}{\textbf{Wan2.7-Image}}
& \multicolumn{2}{c}{\textbf{Qwen-Image-2.0}}
& \multicolumn{2}{c|}{\textbf{Qwen-Image-Edit}}
& \multicolumn{2}{c|}{\textbf{Overall}}
& \multicolumn{2}{c}{\textbf{Real-Damaged}} \\
\cmidrule(lr){2-3}
\cmidrule(lr){4-5}
\cmidrule(lr){6-7}
\cmidrule(lr){8-9}
\cmidrule(lr){10-11}
\cmidrule(lr){12-13}
\cmidrule(lr){14-15}
\cmidrule(lr){16-17}
& \textbf{TPR} & \textbf{Conf.}
& \textbf{TPR} & \textbf{Conf.}
& \textbf{TPR} & \textbf{Conf.}
& \textbf{TPR} & \textbf{Conf.}
& \textbf{TPR} & \textbf{Conf.}
& \textbf{TPR} & \textbf{Conf.}
& \textbf{Bal.Acc.} & \textbf{F1}
& \textbf{TNR} & \textbf{Conf.} \\
\midrule

\rowcolor{mypurple}
\multicolumn{17}{c}{\textbf{Multimodal Large Language Models}} \\
\midrule
DeepSeek-V4.1-Flash & 0.125 & 0.748 & 0.250 & 0.704 & 0.250 & 0.680 & 0.325 & 0.704 & 0.450 & 0.679 & 0.450 & 0.736 & 0.654 & 0.471 & 1.000 & 0.754\\
Gemini 3.1 Pro & 0.100 & 0.718 & 0.075 & 0.717 & 0.175 & 0.719 & 0.200 & 0.718 & 0.300 & 0.714 & 0.650 & 0.709 & 0.625 & 0.400 & 1.000 & 0.707\\
Gemini 3.5 Flash-Lite & 0.025 & 0.990 & 0.025 & 0.950 & 0.050 & 0.970 & 0.075 & 0.970 & 0.000 & — & 0.150 & 0.975 & 0.527 & 0.103 & 1.000 & 0.952\\
Gemini 3.8 Flash & 0.000 & — & 0.000 & — & 0.075 & 0.680 & 0.125 & 0.690 & 0.075 & 0.693 & 0.200 & 0.715 & 0.540 & 0.147 & 1.000 & 0.642\\
GPT-6 Astra & 0.075 & 0.783 & 0.075 & 0.793 & 0.100 & 0.787 & 0.475 & 0.836 & 0.450 & 0.809 & 0.575 & 0.827 & 0.646 & 0.452 & 1.000 & 0.939\\
\bottomrule
\end{tabular}
}
\end{table*}
\begin{table*}[!htp]
\centering
\caption{\textbf{Performance comparison of MLLMs on the Books subset of \method{}.} Results are reported across six AI image editing and generation subsets under the single-image setting without review context.}
\label{tab:T14}
\resizebox{\textwidth}{!}{
\begin{tabular}{c|cc cc cc cc cc cc|cc|cc}
\toprule
\multirow{2}[2]{*}{\textbf{Models}}
& \multicolumn{2}{c}{\textbf{GPT Image 2}}
& \multicolumn{2}{c}{\textbf{Grok Imagine}}
& \multicolumn{2}{c}{\textbf{Nano Banana 2}}
& \multicolumn{2}{c}{\textbf{Wan2.7-Image}}
& \multicolumn{2}{c}{\textbf{Qwen-Image-2.0}}
& \multicolumn{2}{c|}{\textbf{Qwen-Image-Edit}}
& \multicolumn{2}{c|}{\textbf{Overall}}
& \multicolumn{2}{c}{\textbf{Real-Damaged}} \\
\cmidrule(lr){2-3}
\cmidrule(lr){4-5}
\cmidrule(lr){6-7}
\cmidrule(lr){8-9}
\cmidrule(lr){10-11}
\cmidrule(lr){12-13}
\cmidrule(lr){14-15}
\cmidrule(lr){16-17}
& \textbf{TPR} & \textbf{Conf.}
& \textbf{TPR} & \textbf{Conf.}
& \textbf{TPR} & \textbf{Conf.}
& \textbf{TPR} & \textbf{Conf.}
& \textbf{TPR} & \textbf{Conf.}
& \textbf{TPR} & \textbf{Conf.}
& \textbf{Bal.Acc.} & \textbf{F1}
& \textbf{TNR} & \textbf{Conf.} \\
\midrule

\rowcolor{mypurple}
\multicolumn{17}{c}{\textbf{Multimodal Large Language Models}} \\
\midrule
DeepSeek-V4.1-Flash & 0.133 & 0.725 & 0.267 & 0.647 & 0.267 & 0.708 & 0.267 & 0.750 & 0.467 & 0.723 & 0.600 & 0.752 & 0.667 & 0.500 & 1.000 & 0.802\\
Gemini 3.1 Pro & 0.133 & 0.725 & 0.267 & 0.693 & 0.133 & 0.750 & 0.267 & 0.725 & 0.533 & 0.721 & 0.533 & 0.739 & 0.656 & 0.475 & 1.000 & 0.712\\
Gemini 3.5 Flash-Lite & 0.267 & 0.968 & 0.267 & 0.960 & 0.400 & 0.975 & 0.200 & 0.973 & 0.267 & 0.980 & 0.533 & 0.973 & 0.646 & 0.483 & 0.970 & 0.955\\
Gemini 3.8 Flash & 0.067 & 0.720 & 0.000 & — & 0.000 & — & 0.067 & 0.720 & 0.067 & 0.720 & 0.400 & 0.737 & 0.550 & 0.182 & 1.000 & 0.643\\
GPT-6 Astra & 0.067 & 0.910 & 0.133 & 0.835 & 0.000 & — & 0.200 & 0.900 & 0.533 & 0.821 & 0.533 & 0.905 & 0.622 & 0.393 & 1.000 & 0.942\\
\bottomrule
\end{tabular}
}
\end{table*}
\begin{table*}[!htp]
\centering
\caption{\textbf{Performance comparison of MLLMs on the CDs \& Vinyl subset of \method{}.} Results are reported across six AI image editing and generation subsets under the single-image setting without review context.}
\label{tab:T15}
\resizebox{\textwidth}{!}{
\begin{tabular}{c|cc cc cc cc cc cc|cc|cc}
\toprule
\multirow{2}[2]{*}{\textbf{Models}}
& \multicolumn{2}{c}{\textbf{GPT Image 2}}
& \multicolumn{2}{c}{\textbf{Grok Imagine}}
& \multicolumn{2}{c}{\textbf{Nano Banana 2}}
& \multicolumn{2}{c}{\textbf{Wan2.7-Image}}
& \multicolumn{2}{c}{\textbf{Qwen-Image-2.0}}
& \multicolumn{2}{c|}{\textbf{Qwen-Image-Edit}}
& \multicolumn{2}{c|}{\textbf{Overall}}
& \multicolumn{2}{c}{\textbf{Real-Damaged}} \\
\cmidrule(lr){2-3}
\cmidrule(lr){4-5}
\cmidrule(lr){6-7}
\cmidrule(lr){8-9}
\cmidrule(lr){10-11}
\cmidrule(lr){12-13}
\cmidrule(lr){14-15}
\cmidrule(lr){16-17}
& \textbf{TPR} & \textbf{Conf.}
& \textbf{TPR} & \textbf{Conf.}
& \textbf{TPR} & \textbf{Conf.}
& \textbf{TPR} & \textbf{Conf.}
& \textbf{TPR} & \textbf{Conf.}
& \textbf{TPR} & \textbf{Conf.}
& \textbf{Bal.Acc.} & \textbf{F1}
& \textbf{TNR} & \textbf{Conf.} \\
\midrule

\rowcolor{mypurple}
\multicolumn{17}{c}{\textbf{Multimodal Large Language Models}} \\
\midrule
DeepSeek-V4.1-Flash & 0.053 & 0.620 & 0.211 & 0.720 & 0.421 & 0.776 & 0.316 & 0.787 & 0.632 & 0.664 & 0.579 & 0.692 & 0.639 & 0.525 & 0.909 & 0.757\\
Gemini 3.1 Pro & 0.105 & 0.735 & 0.053 & 0.720 & 0.211 & 0.718 & 0.368 & 0.726 & 0.263 & 0.730 & 0.263 & 0.720 & 0.605 & 0.348 & 1.000 & 0.702\\
Gemini 3.5 Flash-Lite & 0.000 & — & 0.000 & — & 0.105 & 0.970 & 0.053 & 0.990 & 0.158 & 0.960 & 0.105 & 0.970 & 0.535 & 0.131 & 1.000 & 0.953\\
Gemini 3.8 Flash & 0.000 & — & 0.000 & — & 0.053 & 0.760 & 0.211 & 0.725 & 0.053 & 0.720 & 0.158 & 0.720 & 0.539 & 0.146 & 1.000 & 0.642\\
GPT-6 Astra & 0.000 & — & 0.053 & 0.870 & 0.158 & 0.823 & 0.158 & 0.843 & 0.316 & 0.825 & 0.211 & 0.815 & 0.575 & 0.260 & 1.000 & 0.937\\
\bottomrule
\end{tabular}
}
\end{table*}
\begin{table*}[!htp]
\centering
\caption{\textbf{Performance comparison of MLLMs on the Cell Phones \& Accessories subset of \method{}.} Results are reported across six AI image editing and generation subsets under the single-image setting without review context.}
\label{tab:T16}
\resizebox{\textwidth}{!}{
\begin{tabular}{c|cc cc cc cc cc cc|cc|cc}
\toprule
\multirow{2}[2]{*}{\textbf{Models}}
& \multicolumn{2}{c}{\textbf{GPT Image 2}}
& \multicolumn{2}{c}{\textbf{Grok Imagine}}
& \multicolumn{2}{c}{\textbf{Nano Banana 2}}
& \multicolumn{2}{c}{\textbf{Wan2.7-Image}}
& \multicolumn{2}{c}{\textbf{Qwen-Image-2.0}}
& \multicolumn{2}{c|}{\textbf{Qwen-Image-Edit}}
& \multicolumn{2}{c|}{\textbf{Overall}}
& \multicolumn{2}{c}{\textbf{Real-Damaged}} \\
\cmidrule(lr){2-3}
\cmidrule(lr){4-5}
\cmidrule(lr){6-7}
\cmidrule(lr){8-9}
\cmidrule(lr){10-11}
\cmidrule(lr){12-13}
\cmidrule(lr){14-15}
\cmidrule(lr){16-17}
& \textbf{TPR} & \textbf{Conf.}
& \textbf{TPR} & \textbf{Conf.}
& \textbf{TPR} & \textbf{Conf.}
& \textbf{TPR} & \textbf{Conf.}
& \textbf{TPR} & \textbf{Conf.}
& \textbf{TPR} & \textbf{Conf.}
& \textbf{Bal.Acc.} & \textbf{F1}
& \textbf{TNR} & \textbf{Conf.} \\
\midrule

\rowcolor{mypurple}
\multicolumn{17}{c}{\textbf{Multimodal Large Language Models}} \\
\midrule
DeepSeek-V4.1-Flash & 0.087 & 0.620 & 0.152 & 0.683 & 0.217 & 0.676 & 0.326 & 0.707 & 0.370 & 0.711 & 0.457 & 0.707 & 0.598 & 0.419 & 0.929 & 0.766\\
Gemini 3.1 Pro & 0.109 & 0.712 & 0.087 & 0.718 & 0.217 & 0.718 & 0.304 & 0.713 & 0.391 & 0.717 & 0.543 & 0.715 & 0.638 & 0.432 & 1.000 & 0.707\\
Gemini 3.5 Flash-Lite & 0.022 & 0.990 & 0.022 & 0.950 & 0.087 & 0.958 & 0.065 & 0.977 & 0.152 & 0.964 & 0.239 & 0.974 & 0.549 & 0.178 & 1.000 & 0.946\\
Gemini 3.8 Flash & 0.043 & 0.700 & 0.043 & 0.680 & 0.065 & 0.713 & 0.152 & 0.703 & 0.261 & 0.706 & 0.391 & 0.711 & 0.580 & 0.275 & 1.000 & 0.642\\
GPT-6 Astra & 0.174 & 0.741 & 0.239 & 0.767 & 0.326 & 0.807 & 0.478 & 0.842 & 0.587 & 0.862 & 0.630 & 0.847 & 0.703 & 0.577 & 1.000 & 0.925\\
\bottomrule
\end{tabular}
}
\end{table*}
\begin{table*}[!htp]
\centering
\caption{\textbf{Performance comparison of MLLMs on the Clothing, Shoes \& Jewelry subset of \method{}.} Results are reported across six AI image editing and generation subsets under the single-image setting without review context.}
\label{tab:T17}
\resizebox{\textwidth}{!}{
\begin{tabular}{c|cc cc cc cc cc cc|cc|cc}
\toprule
\multirow{2}[2]{*}{\textbf{Models}}
& \multicolumn{2}{c}{\textbf{GPT Image 2}}
& \multicolumn{2}{c}{\textbf{Grok Imagine}}
& \multicolumn{2}{c}{\textbf{Nano Banana 2}}
& \multicolumn{2}{c}{\textbf{Wan2.7-Image}}
& \multicolumn{2}{c}{\textbf{Qwen-Image-2.0}}
& \multicolumn{2}{c|}{\textbf{Qwen-Image-Edit}}
& \multicolumn{2}{c|}{\textbf{Overall}}
& \multicolumn{2}{c}{\textbf{Real-Damaged}} \\
\cmidrule(lr){2-3}
\cmidrule(lr){4-5}
\cmidrule(lr){6-7}
\cmidrule(lr){8-9}
\cmidrule(lr){10-11}
\cmidrule(lr){12-13}
\cmidrule(lr){14-15}
\cmidrule(lr){16-17}
& \textbf{TPR} & \textbf{Conf.}
& \textbf{TPR} & \textbf{Conf.}
& \textbf{TPR} & \textbf{Conf.}
& \textbf{TPR} & \textbf{Conf.}
& \textbf{TPR} & \textbf{Conf.}
& \textbf{TPR} & \textbf{Conf.}
& \textbf{Bal.Acc.} & \textbf{F1}
& \textbf{TNR} & \textbf{Conf.} \\
\midrule

\rowcolor{mypurple}
\multicolumn{17}{c}{\textbf{Multimodal Large Language Models}} \\
\midrule
DeepSeek-V4.1-Flash & 0.128 & 0.753 & 0.191 & 0.694 & 0.213 & 0.728 & 0.149 & 0.659 & 0.362 & 0.736 & 0.511 & 0.681 & 0.608 & 0.410 & 0.957 & 0.779\\
Gemini 3.1 Pro & 0.085 & 0.718 & 0.085 & 0.715 & 0.128 & 0.708 & 0.128 & 0.720 & 0.340 & 0.702 & 0.362 & 0.714 & 0.594 & 0.316 & 1.000 & 0.682\\
Gemini 3.5 Flash-Lite & 0.064 & 0.950 & 0.021 & 0.950 & 0.043 & 0.950 & 0.021 & 0.950 & 0.149 & 0.950 & 0.106 & 0.964 & 0.534 & 0.126 & 1.000 & 0.957\\
Gemini 3.8 Flash & 0.021 & 0.680 & 0.000 & — & 0.064 & 0.707 & 0.021 & 0.720 & 0.000 & — & 0.128 & 0.693 & 0.520 & 0.075 & 1.000 & 0.692\\
GPT-6 Astra & 0.085 & 0.755 & 0.064 & 0.837 & 0.064 & 0.843 & 0.213 & 0.829 & 0.170 & 0.797 & 0.468 & 0.812 & 0.589 & 0.301 & 1.000 & 0.939\\
\bottomrule
\end{tabular}
}
\end{table*}
\begin{table*}[!htp]
\centering
\caption{\textbf{Performance comparison of MLLMs on the Delivery, Pickup \& Dine-Out subset of \method{}.} Results are reported across six AI image editing and generation subsets under the single-image setting without review context.}
\label{tab:T18}
\resizebox{\textwidth}{!}{
\begin{tabular}{c|cc cc cc cc cc cc|cc|cc}
\toprule
\multirow{2}[2]{*}{\textbf{Models}}
& \multicolumn{2}{c}{\textbf{GPT Image 2}}
& \multicolumn{2}{c}{\textbf{Grok Imagine}}
& \multicolumn{2}{c}{\textbf{Nano Banana 2}}
& \multicolumn{2}{c}{\textbf{Wan2.7-Image}}
& \multicolumn{2}{c}{\textbf{Qwen-Image-2.0}}
& \multicolumn{2}{c|}{\textbf{Qwen-Image-Edit}}
& \multicolumn{2}{c|}{\textbf{Overall}}
& \multicolumn{2}{c}{\textbf{Real-Damaged}} \\
\cmidrule(lr){2-3}
\cmidrule(lr){4-5}
\cmidrule(lr){6-7}
\cmidrule(lr){8-9}
\cmidrule(lr){10-11}
\cmidrule(lr){12-13}
\cmidrule(lr){14-15}
\cmidrule(lr){16-17}
& \textbf{TPR} & \textbf{Conf.}
& \textbf{TPR} & \textbf{Conf.}
& \textbf{TPR} & \textbf{Conf.}
& \textbf{TPR} & \textbf{Conf.}
& \textbf{TPR} & \textbf{Conf.}
& \textbf{TPR} & \textbf{Conf.}
& \textbf{Bal.Acc.} & \textbf{F1}
& \textbf{TNR} & \textbf{Conf.} \\
\midrule

\rowcolor{mypurple}
\multicolumn{17}{c}{\textbf{Multimodal Large Language Models}} \\
\midrule
DeepSeek-V4.1-Flash & 0.040 & 0.758 & 0.070 & 0.689 & 0.070 & 0.651 & 0.080 & 0.657 & 0.150 & 0.689 & 0.210 & 0.743 & 0.537 & 0.186 & 0.970 & 0.772\\
Gemini 3.1 Pro & 0.080 & 0.716 & 0.120 & 0.721 & 0.090 & 0.712 & 0.280 & 0.719 & 0.290 & 0.717 & 0.620 & 0.714 & 0.593 & 0.393 & 0.940 & 0.700\\
Gemini 3.5 Flash-Lite & 0.010 & 0.950 & 0.060 & 0.950 & 0.030 & 0.960 & 0.120 & 0.968 & 0.060 & 0.955 & 0.290 & 0.969 & 0.537 & 0.173 & 0.980 & 0.949\\
Gemini 3.8 Flash & 0.010 & 0.720 & 0.000 & — & 0.010 & 0.720 & 0.020 & 0.720 & 0.010 & 0.730 & 0.120 & 0.704 & 0.514 & 0.055 & 1.000 & 0.666\\
GPT-6 Astra & 0.040 & 0.780 & 0.110 & 0.787 & 0.070 & 0.787 & 0.280 & 0.804 & 0.220 & 0.804 & 0.540 & 0.828 & 0.585 & 0.345 & 0.960 & 0.926\\
\bottomrule
\end{tabular}
}
\end{table*}
\begin{table*}[!htp]
\centering
\caption{\textbf{Performance comparison of MLLMs on the Electronics subset of \method{}.} Results are reported across six AI image editing and generation subsets under the single-image setting without review context.}
\label{tab:T19}
\resizebox{\textwidth}{!}{
\begin{tabular}{c|cc cc cc cc cc cc|cc|cc}
\toprule
\multirow{2}[2]{*}{\textbf{Models}}
& \multicolumn{2}{c}{\textbf{GPT Image 2}}
& \multicolumn{2}{c}{\textbf{Grok Imagine}}
& \multicolumn{2}{c}{\textbf{Nano Banana 2}}
& \multicolumn{2}{c}{\textbf{Wan2.7-Image}}
& \multicolumn{2}{c}{\textbf{Qwen-Image-2.0}}
& \multicolumn{2}{c|}{\textbf{Qwen-Image-Edit}}
& \multicolumn{2}{c|}{\textbf{Overall}}
& \multicolumn{2}{c}{\textbf{Real-Damaged}} \\
\cmidrule(lr){2-3}
\cmidrule(lr){4-5}
\cmidrule(lr){6-7}
\cmidrule(lr){8-9}
\cmidrule(lr){10-11}
\cmidrule(lr){12-13}
\cmidrule(lr){14-15}
\cmidrule(lr){16-17}
& \textbf{TPR} & \textbf{Conf.}
& \textbf{TPR} & \textbf{Conf.}
& \textbf{TPR} & \textbf{Conf.}
& \textbf{TPR} & \textbf{Conf.}
& \textbf{TPR} & \textbf{Conf.}
& \textbf{TPR} & \textbf{Conf.}
& \textbf{Bal.Acc.} & \textbf{F1}
& \textbf{TNR} & \textbf{Conf.} \\
\midrule

\rowcolor{mypurple}
\multicolumn{17}{c}{\textbf{Multimodal Large Language Models}} \\
\midrule
DeepSeek-V4.1-Flash & 0.135 & 0.677 & 0.308 & 0.684 & 0.212 & 0.696 & 0.212 & 0.708 & 0.423 & 0.717 & 0.385 & 0.709 & 0.587 & 0.432 & 0.895 & 0.768\\
Gemini 3.1 Pro & 0.096 & 0.704 & 0.096 & 0.700 & 0.135 & 0.710 & 0.173 & 0.717 & 0.481 & 0.711 & 0.635 & 0.712 & 0.621 & 0.423 & 0.974 & 0.695\\
Gemini 3.5 Flash-Lite & 0.077 & 0.975 & 0.058 & 0.973 & 0.096 & 0.958 & 0.154 & 0.980 & 0.231 & 0.966 & 0.250 & 0.968 & 0.572 & 0.252 & 1.000 & 0.950\\
Gemini 3.8 Flash & 0.019 & 0.700 & 0.019 & 0.720 & 0.096 & 0.688 & 0.192 & 0.710 & 0.269 & 0.709 & 0.250 & 0.706 & 0.571 & 0.247 & 1.000 & 0.670\\
GPT-6 Astra & 0.231 & 0.770 & 0.231 & 0.787 & 0.365 & 0.790 & 0.615 & 0.852 & 0.577 & 0.851 & 0.750 & 0.846 & 0.731 & 0.632 & 1.000 & 0.933\\
\bottomrule
\end{tabular}
}
\end{table*}
\begin{table*}[!htp]
\centering
\caption{\textbf{Performance comparison of MLLMs on the Grocery \& Gourmet Food subset of \method{}.} Results are reported across six AI image editing and generation subsets under the single-image setting without review context.}
\label{tab:T20}
\resizebox{\textwidth}{!}{
\begin{tabular}{c|cc cc cc cc cc cc|cc|cc}
\toprule
\multirow{2}[2]{*}{\textbf{Models}}
& \multicolumn{2}{c}{\textbf{GPT Image 2}}
& \multicolumn{2}{c}{\textbf{Grok Imagine}}
& \multicolumn{2}{c}{\textbf{Nano Banana 2}}
& \multicolumn{2}{c}{\textbf{Wan2.7-Image}}
& \multicolumn{2}{c}{\textbf{Qwen-Image-2.0}}
& \multicolumn{2}{c|}{\textbf{Qwen-Image-Edit}}
& \multicolumn{2}{c|}{\textbf{Overall}}
& \multicolumn{2}{c}{\textbf{Real-Damaged}} \\
\cmidrule(lr){2-3}
\cmidrule(lr){4-5}
\cmidrule(lr){6-7}
\cmidrule(lr){8-9}
\cmidrule(lr){10-11}
\cmidrule(lr){12-13}
\cmidrule(lr){14-15}
\cmidrule(lr){16-17}
& \textbf{TPR} & \textbf{Conf.}
& \textbf{TPR} & \textbf{Conf.}
& \textbf{TPR} & \textbf{Conf.}
& \textbf{TPR} & \textbf{Conf.}
& \textbf{TPR} & \textbf{Conf.}
& \textbf{TPR} & \textbf{Conf.}
& \textbf{Bal.Acc.} & \textbf{F1}
& \textbf{TNR} & \textbf{Conf.} \\
\midrule

\rowcolor{mypurple}
\multicolumn{17}{c}{\textbf{Multimodal Large Language Models}} \\
\midrule
DeepSeek-V4.1-Flash & 0.059 & 0.690 & 0.176 & 0.738 & 0.176 & 0.737 & 0.294 & 0.709 & 0.265 & 0.754 & 0.559 & 0.757 & 0.589 & 0.398 & 0.923 & 0.738\\
Gemini 3.1 Pro & 0.059 & 0.700 & 0.029 & 0.720 & 0.176 & 0.727 & 0.265 & 0.711 & 0.265 & 0.713 & 0.706 & 0.708 & 0.625 & 0.400 & 1.000 & 0.709\\
Gemini 3.5 Flash-Lite & 0.059 & 0.965 & 0.059 & 0.965 & 0.147 & 0.964 & 0.147 & 0.974 & 0.206 & 0.971 & 0.441 & 0.969 & 0.573 & 0.298 & 0.969 & 0.951\\
Gemini 3.8 Flash & 0.000 & — & 0.000 & — & 0.029 & 0.720 & 0.059 & 0.730 & 0.029 & 0.720 & 0.206 & 0.720 & 0.527 & 0.102 & 1.000 & 0.666\\
GPT-6 Astra & 0.029 & 0.830 & 0.088 & 0.867 & 0.059 & 0.885 & 0.353 & 0.860 & 0.235 & 0.850 & 0.471 & 0.860 & 0.603 & 0.341 & 1.000 & 0.936\\
\bottomrule
\end{tabular}
}
\end{table*}
\begin{table*}[!htp]
\centering
\caption{\textbf{Performance comparison of MLLMs on the Handmade Products subset of \method{}.} Results are reported across six AI image editing and generation subsets under the single-image setting without review context.}
\label{tab:T21}
\resizebox{\textwidth}{!}{
\begin{tabular}{c|cc cc cc cc cc cc|cc|cc}
\toprule
\multirow{2}[2]{*}{\textbf{Models}}
& \multicolumn{2}{c}{\textbf{GPT Image 2}}
& \multicolumn{2}{c}{\textbf{Grok Imagine}}
& \multicolumn{2}{c}{\textbf{Nano Banana 2}}
& \multicolumn{2}{c}{\textbf{Wan2.7-Image}}
& \multicolumn{2}{c}{\textbf{Qwen-Image-2.0}}
& \multicolumn{2}{c|}{\textbf{Qwen-Image-Edit}}
& \multicolumn{2}{c|}{\textbf{Overall}}
& \multicolumn{2}{c}{\textbf{Real-Damaged}} \\
\cmidrule(lr){2-3}
\cmidrule(lr){4-5}
\cmidrule(lr){6-7}
\cmidrule(lr){8-9}
\cmidrule(lr){10-11}
\cmidrule(lr){12-13}
\cmidrule(lr){14-15}
\cmidrule(lr){16-17}
& \textbf{TPR} & \textbf{Conf.}
& \textbf{TPR} & \textbf{Conf.}
& \textbf{TPR} & \textbf{Conf.}
& \textbf{TPR} & \textbf{Conf.}
& \textbf{TPR} & \textbf{Conf.}
& \textbf{TPR} & \textbf{Conf.}
& \textbf{Bal.Acc.} & \textbf{F1}
& \textbf{TNR} & \textbf{Conf.} \\
\midrule

\rowcolor{mypurple}
\multicolumn{17}{c}{\textbf{Multimodal Large Language Models}} \\
\midrule
DeepSeek-V4.1-Flash & 0.040 & 0.620 & 0.060 & 0.617 & 0.160 & 0.639 & 0.120 & 0.722 & 0.240 & 0.683 & 0.320 & 0.722 & 0.578 & 0.271 & 1.000 & 0.741\\
Gemini 3.1 Pro & 0.060 & 0.717 & 0.140 & 0.700 & 0.180 & 0.713 & 0.180 & 0.709 & 0.300 & 0.708 & 0.580 & 0.715 & 0.620 & 0.387 & 1.000 & 0.708\\
Gemini 3.5 Flash-Lite & 0.000 & — & 0.000 & — & 0.040 & 0.950 & 0.060 & 0.950 & 0.160 & 0.950 & 0.160 & 0.964 & 0.535 & 0.131 & 1.000 & 0.950\\
Gemini 3.8 Flash & 0.000 & — & 0.000 & — & 0.000 & — & 0.020 & 0.720 & 0.040 & 0.700 & 0.100 & 0.704 & 0.513 & 0.052 & 1.000 & 0.688\\
GPT-6 Astra & 0.020 & 0.860 & 0.020 & 0.720 & 0.120 & 0.802 & 0.360 & 0.806 & 0.440 & 0.815 & 0.460 & 0.825 & 0.618 & 0.383 & 1.000 & 0.918\\
\bottomrule
\end{tabular}
}
\end{table*}
\input{Table/T22}
\input{Table/T23}
\input{Table/T24}
\input{Table/T25}
\input{Table/T26}
\input{Table/T27}
\input{Table/T28}
\input{Table/T29}
\input{Table/T30}
\input{Table/T31}
\input{Table/T32}
\input{Table/T33}
\input{Table/T34}
\input{Table/T35}

\begin{table*}[!htp]
\centering
\caption{\textbf{Human evaluator performance on \method{}.} Results are reported under the single-image setting without review context.}
\label{tab:human_eval}
\resizebox{\textwidth}{!}{%
\begin{tabular}{l|cccccc|ccc|c}
\toprule
\multirow{2}[2]{*}{\textbf{Evaluator}}
  & \multirow{2}[2]{*}{\textbf{GPT Image 2}}
  & \multirow{2}[2]{*}{\textbf{Grok Imagine}}
  & \multirow{2}[2]{*}{\textbf{Nano Banana 2}}
  & \multirow{2}[2]{*}{\textbf{Wan2.7-Image}}
  & \multirow{2}[2]{*}{\textbf{Qwen-Image-2.0}}
  & \multirow{2}[2]{*}{\textbf{Qwen-Image-Edit}}
  & \multicolumn{3}{c|}{\textbf{Overall}}
  & \multirow{2}[2]{*}{\textbf{Real-Damaged}} \\
\cmidrule(lr){8-10}
 & & & & & & & \textbf{Bal.Acc} & \textbf{F1} & \textbf{Acc} & \\
\midrule
\rowcolor{mypurple}
\multicolumn{11}{c}{\textbf{Human Evaluators}} \\
\midrule
Human 1 & 0.483 & 0.577 & 0.596 & 0.613 & 0.696 & 0.695 & 0.610 & 0.728 & 0.610 & 0.611\\
Human 2 & 0.195 & 0.255 & 0.305 & 0.413 & 0.534 & 0.560 & 0.676 & 0.546 & 0.464 & 0.974\\
Human 3 & 0.582 & 0.695 & 0.717 & 0.767 & 0.859 & 0.871 & 0.772 & 0.839 & 0.755 & 0.795\\
Average & 0.420 & 0.509 & 0.539 & 0.598 & 0.697 & 0.709 & 0.686 & 0.704 & 0.610 & 0.793\\
\bottomrule
\end{tabular}}
\end{table*}
\input{Table/Abl_Damage_Type}
\input{Table/Abl_T1_Baseline}
\input{Table/Abl_T2_Merged}
\input{Table/Abl_T3_No_Artifacts}
\input{Table/Abl_T4_Generic}
\input{Table/Abl_T5_Minimal}
\input{Table/Abl_Mismatched_Prompt}
\input{Table/Abl_Mismatch_Compare}

\newpage
\subsection{Supplementary Evaluation Explorations}
\label{app:supp_eval_explorations}

This section presents eight supplementary explorations that extend the primary results along axes not fully captured by the main benchmark settings. Appendix~\ref{app:category_difficulty} examines how detection difficulty varies systematically across the 29 product and service categories in \method{}. Appendix~\ref{app:damage_type_sensitivity} analyzes how the physical form of the claimed defect shapes detectability. Appendix~\ref{app:image_text_mismatch} probes whether review-conditioned detection reflects genuine cross-modal claim verification or weaker semantic shortcuts. Appendix~\ref{app:prompt_sensitivity} quantifies the sensitivity of MLLM detection performance to prompt formulation. Appendix~\ref{app:threshold_free} evaluates the specialized detectors without a fixed decision threshold, and Appendix~\ref{app:ifdl} evaluates image forgery detection and localization methods. Appendix~\ref{app:localization} compares how MLLMs and IFDL methods localize the manipulated region, and Appendix~\ref{app:complementarity} examines whether MLLMs and specialized detectors detect the same fake-damaged images. Together, these explorations provide a more granular characterization of current detector limitations and motivate targeted directions for future work.

\subsubsection{Category-Level Detection Difficulty}
\label{app:category_difficulty}

The per-category breakdown across Tab.~\ref{tab:T7}--\ref{tab:T35} reveals that detection difficulty varies substantially and systematically across product and service categories. Averaged across the 16 MLLMs in Tab.~\ref{tab:main}, category-level balanced accuracy spans from 0.534 in Baby Products to 0.646 in Hotels \& Accommodations, with the middle half of categories between 0.562 and 0.604. The per-category tables also list the 14 other evaluated MLLMs in a separate block, which is not included in these averages. This spread is sufficiently large to affect practical conclusions about model competence and indicates that \method{} probes genuine category-dependent variation in forensic difficulty.

Categories with more structurally distinctive or semantically grounded evidence submissions tend to be easier for current MLLMs. Hotels \& Accommodations, All Beauty, and Musical Instruments represent the higher end of the performance distribution (0.646, 0.630, and 0.628). In hotel evidence, damage claims typically involve recognizable indoor fixtures such as bedding, bathroom fittings, and walls, whose physical states carry clear expectations and whose AI-generated manipulations produce spatially localized artifacts against a structured background. Beauty products and musical instruments similarly feature highly constrained visual elements, including printed labels, typography, and regular object geometry, which make AI-generated edits to surfaces, seams, or labels more identifiable through local consistency checks. For these categories, the presence of text, branded visual elements, and regular geometry provides MLLMs with forensic anchors that are absent in more visually heterogeneous categories.

By contrast, categories such as Baby Products (0.534), Tools \& Home Improvement (0.536), Delivery, Pickup \& Dine-Out (0.539), and Amazon Fashion (0.553) cluster at the lower end of performance. Baby product evidence predominantly consists of plush toys, soft packaging, and infant accessories, whose surfaces lack the sharp geometric regularity or semantic distinctiveness needed to make AI-generated damage boundaries stand out. Apparel evidence in Amazon Fashion is similarly affected: fabric textures, natural wrinkles, and variable lighting conditions create substantial within-category appearance diversity, making it difficult to distinguish AI-generated damage artifacts from authentic surface variation. The Delivery, Pickup \& Dine-Out category poses a distinct challenge: food presentation varies greatly across vendors and occasions, and damage patterns such as crushed packaging, contamination, or spillage can be visually subtle or ambiguous against backgrounds that are themselves cluttered and uncontrolled.

Model-level patterns further illuminate these category effects. Among the 16 MLLMs in Tab.~\ref{tab:main}, no single model dominates: Qwen3.8-Max achieves the highest balanced accuracy, alone or tied, in 11 of the 29 categories, GPT-6 Astra in 10, and Qwen3.8-Flash in six. Even the strongest model varies widely across categories: Qwen3.8-Max reaches 0.777 in All Beauty and 0.744 in Hotels \& Accommodations but only 0.542 in Baby Products and 0.589 in Delivery, Pickup \& Dine-Out, a gap of more than 20 percentage points. These model-by-category interactions suggest that category-specific forensic competence is not captured by a single aggregate metric and that different model architectures may respond differently to the visual forensic cues dominant in each category.

These results carry a practical implication for evaluating refund-evidence detectors. Because product category is not independent of damage type, generator behavior, or evidence style, aggregate benchmarking can mask systematic category-level blind spots. A detector that performs acceptably on average may nevertheless be unreliable precisely for categories such as fashion, food delivery, and everyday household goods, which represent the highest volume and most heterogeneous distribution of real platform disputes. Future work should therefore stratify evaluations by category as a first-class axis, and prioritize improvement on categories with heterogeneous object appearance, unconstrained imaging conditions, and damage patterns that overlap with natural product variation.

\subsubsection{Damage-Type Sensitivity}
\label{app:damage_type_sensitivity}

The damage-type breakdown reveals that detection difficulty is not uniformly distributed across fake-damage categories: the physical form of the claimed defect systematically shapes how detectable it is for current MLLMs. As shown in Tab.~\ref{tab:damage_type}, structurally salient or materially discontinuous defects are generally more accessible to MLLM detection. Averaged across the 16 MLLMs in Tab.~\ref{tab:main}, cracked, leaking, shattered/broken, and melted evidence yield average TPRs of 0.331, 0.288, 0.282, and 0.266. By contrast, packaging-damaged, torn/ripped, mouldy/spoiled, and dented evidence remain substantially harder, with average TPRs of 0.144, 0.152, 0.163, and 0.192. This performance gap suggests that current MLLMs are more responsive to obvious object-level breakage, material discontinuities, or fluid-related damage than to subtle deformation, surface-level complaints, or packaging-level defects.

The difficulty further varies across models. GPT-6 Astra achieves 0.534 TPR on cracked evidence and Qwen3.8-Max 0.524 on melted evidence, while multiple models approach zero TPR on packaging-damaged samples; Qwen3-VL-Plus, for instance, obtains only 0.001 TPR on this damage type. GPT-6 Luna reaches high TPR on every damage type (0.803 to 0.990) only because it flags most images as AI-generated, including most genuine damaged evidence (TNR 0.206). These results indicate that detector performance depends not only on the image generator or product category, but also on the physical manifestation of the claimed defect, independently of other evaluation axes.

This finding carries direct implications for refund-evidence verification in practice. Many genuine disputes involve subtle or localized damage such as packaging deformation, shallow dents, stains, or minor tears, rather than visually dramatic breakage. A detector that performs reliably only on conspicuous defects therefore risks overestimating its practical utility in real platform settings. Future work should treat damage type as a first-class evaluation dimension, examined in conjunction with generator identity, product category, and input modality. Promising directions include material-aware visual reasoning that accounts for object composition and physical properties, damage-plausibility modeling grounded in mechanical and structural constraints, and local comparison methods that evaluate whether a claimed defect is geometrically and physically consistent with the object and scene context.

\subsubsection{Image-Text Mismatch}
\label{app:image_text_mismatch}
The image–text mismatch experiment is designed to test whether review-conditioned detection reflects genuine cross-modal claim verification or merely exploits weaker semantic shortcuts present in the review text. In the standard review-conditioned setting, each image is paired with the customer review from the same product or service instance, enabling models to integrate both visual forensic cues and semantic consistency between the image and the stated claim. In the mismatched-review condition, the matched review is replaced with a genuine review sampled uniformly at random from a different Amazon product category, while the image input, prompt structure, and output schema are held fixed. Because every image then receives a genuine review, the condition also removes the stylistic difference between generated and genuine reviews. This controlled design provides a diagnostic for whether model predictions are sensitive to semantic coherence between review text and visual content, rather than driven by visual evidence alone.

As shown in Tab.~\ref{tab:mismatch_compare} and detailed in Tab.~\ref{tab:mismatch_results}, the benefit of review text disappears once the review no longer matches the evidence. Averaged over the six MLLMs, balanced accuracy on the same 579 images is 0.595 without review text, 0.622 with the matched review, and 0.587 with a mismatched review. Gemini 3.8 Flash illustrates the pattern most clearly, rising from 0.541 to 0.633 with the matched review but only to 0.556 with a mismatched one. Mismatched reviews can also harm genuine evidence: the TNR of DeepSeek-V4.1-Flash falls from 0.966 without review text to 0.828 under mismatch and that of Kimi K3 from 0.974 to 0.935, indicating that these models treat an incoherent review as a sign of manipulation. GPT-5.6 Terra is the exception, performing worse with the matched review (0.599) than without review text (0.631) or with a mismatched review (0.630).

The nine additional MLLMs in the second block of Tab.~\ref{tab:mismatch_compare} and~\ref{tab:mismatch_results} depend on the review even less. The matched review raises their average balanced accuracy only from 0.573 to 0.586, and a mismatched review yields a similar value (0.587). The mismatched review nevertheless lowers their average real-image TNR from 0.931 without review text to 0.907, as five of the nine models flag more genuine evidence.

Comparing the mismatched condition against the standard review-conditioned setting further clarifies the role of textual context. In the primary results, adding review text produces only a marginal average improvement in single-image balanced accuracy (0.591 to 0.619 for the six MLLMs in Tab.~\ref{tab:main_results_review}). Since the matched gain vanishes when both classes receive genuine reviews, it may reflect either image--text coherence or the stylistic difference between generated and genuine reviews, suggesting that current MLLMs do not consistently anchor their decisions in semantic alignment between the review and the visual evidence. The mismatched condition probes the complementary failure mode: whether a model is inappropriately influenced by plausible but semantically irrelevant review text. A reliable claim-verification model should neither treat coherent review text as confirmatory evidence nor discard textual context entirely; instead, it should assess whether the submitted visual evidence specifically supports the stated claim. These results motivate future evaluation protocols that include explicit image–text consistency stress tests: hard negative reviews from visually similar product categories, partially correct reviews that name the correct object but the wrong defect type, and adversarial reviews that plausibly describe a defect absent from the image. Such designs would better distinguish generic AI-image detection from genuine refund-evidence verification, where the core requirement is to determine whether specific visual evidence substantiates a specific textual claim.

\subsubsection{Prompt Sensitivity}
\label{app:prompt_sensitivity}

The prompt-sensitivity study demonstrates that MLLM-based forensic detection is sensitive to prompt formulation, yet prompt engineering alone is insufficient to resolve the difficulty of fake-damaged evidence verification. As summarized in Tab.~\ref{tab:prompt_ablation} and detailed in Tab.~\ref{tab:prompt_v1}--\ref{tab:prompt_v5}, no single prompt is best for all six models, and even the best configuration reaches an F1 of only 0.532.

Replacing the forensic role with a generic assistant identity is the most consistently harmful change: it lowers F1 for five of the six models, for example from 0.400 to 0.207 for GPT-5.6 Terra and from 0.222 to 0.095 for Grok 4.6. The Baseline prompt yields the highest F1 for Gemini 3.8 Flash (0.150) and Grok 4.6 (0.222), and the second-highest for GPT-5.6 Terra (0.400, behind 0.416 under No Checklist).

The trend is not monotonic across models. Merging the role instructions into the user turn raises F1 for Kimi K3 from 0.171 to 0.263. DeepSeek-V4.1-Flash attains its highest F1 under the Minimal (0.532) and Merged (0.526) prompts, but these gains come from flagging more genuine evidence: its real-image TNR falls from 0.954 under Baseline to 0.750 and 0.698, and its balanced accuracy is highest under Baseline (0.610). Consequently, the ablation does not identify a universally optimal template; rather, it reveals that prompt structure itself constitutes a measurable source of evaluation variance, with effects that differ across model families. The ten additional MLLMs in the second block of Tab.~\ref{tab:prompt_v1}--\ref{tab:prompt_v5}, evaluated with the same five prompts on the same 579 images, show the same pattern. The generic assistant identity lowers F1 for all ten, the Baseline prompt yields the highest F1 for seven of them and the Merged prompt for the other three, and the highest F1 among them is 0.541, reached by Gemini 3 Flash with the Baseline prompt.

\subsubsection{Threshold-Free Evaluation of Specialized Detectors}
\label{app:threshold_free}

The threshold-free evaluation reveals that the ranking of specialized detectors depends on how their operating point is chosen. Tab.~\ref{tab:main} reports every specialized detector at its default decision threshold of 0.5, whereas refund verification assigns different costs to false positives and false negatives. We therefore compute AUROC, the standardized partial AUROC for FPR $\leq 0.1$, and the TPR at fixed false-positive rates of 1\%, 5\%, and 10\% from the detector scores on the same 6,940 images, as reported in Tab.~\ref{tab:threshold_free}.

The ranking by AUROC agrees closely with the balanced-accuracy ranking in Tab.~\ref{tab:main} (Spearman $\rho = 0.929$) and with the ranking by TPR at 5\% FPR ($\rho = 0.976$), but it is nearly unrelated to the ranking by TPR at the default threshold ($\rho = 0.071$). IAPL [SD-v1.4] illustrates this gap most clearly: it achieves the highest TPR at the default threshold (0.956) but flags 55.2\% of the real-damaged evidence, and its TPR falls to 0.275 once the FPR is held at 5\%. Effort [Chameleon] reaches an AUROC of only 0.399, meaning that its scores rank real-damaged evidence above fake-damaged evidence more often than not. ForgeLens [GenImage] is the strongest detector under every threshold-free metric, with an AUROC of 0.913 and a TPR of 0.840 at 5\% FPR.

These results indicate that the default-threshold TPR alone overstates the practical utility of over-sensitive detectors, and that detector comparisons for refund verification should be reported at operating points with a bounded false-positive rate. MLLMs return a binary verdict with a coarse self-reported confidence rather than a calibrated score, so we report them only at their fixed operating points in Tab.~\ref{tab:main}. Since real and synthetic evidence also differ in file-level properties such as image size and format, the detector scores may partly reflect such properties.

\input{Table/Abl_Threshold_Free}

\subsubsection{Image Forgery Detection and Localization Baselines}
\label{app:ifdl}

The IFDL evaluation reveals that existing forgery detection and localization methods do not solve \method{} either. As shown in Tab.~\ref{tab:ifdl}, all six IFDL checkpoints remain at or below chance, with balanced accuracy between 0.468 and 0.498, whereas the AI-generated image detectors in Tab.~\ref{tab:main} range from 0.452 to 0.904.

Their failure modes mirror the checkpoint dependence observed for specialized detectors. MVSS-Net [CASIA] flags almost every image as manipulated, with TPR between 0.890 and 0.989 but a TNR of only 0.034, whereas MVSS-Net [DEFACTO] flags few images, with TPR between 0.112 and 0.181 and a TNR of 0.836. TruFor and FakeShield retain most real-damaged evidence, with TNR of 0.825 and 0.823, but detect at most 0.199 and 0.181 of the fake-damaged images of any generator. These methods are trained on spliced, copy-moved, or inpainted manipulations, whose traces differ from those of generative editing, in which the generator re-renders the whole frame. This suggests that broader manipulation coverage alone is insufficient, and that forgery detection for refund evidence requires exposure to generative editing.

\input{Table/Abl_IFDL}

\subsubsection{Localization of the Manipulated Region}
\label{app:localization}

The localization evaluation reveals that neither MLLMs nor IFDL methods reliably locate the edited region, and that the two families point at different parts of the image. As shown in Tab.~\ref{tab:localization}, Gemini 3 Flash reaches a mean IoU of 0.208 with the manipulated area on the 786 images it flags with a box, compared with 0.087 for FakeShield on the same images and at most 0.101 for the IFDL checkpoints scored on every fake-damaged image. Its advantage partly reflects coarse boxes, which cover 38.8\% of the frame on average, whereas the masks of FakeShield cover only 5.5\%. Neither method tracks the extent of the edit, as the correlation between the predicted and the true area is $-0.055$ for Gemini 3 Flash and $-0.040$ for FakeShield. Both families also localize manipulations that do not exist: Gemini 3 Flash returns a box for 289 of the 1,012 real-damaged images, and FakeShield returns a mask for 155. The IoU of MVSS-Net [CASIA] changes gradually with the size of the ground truth, from 0.066 to 0.101 and 0.128 as it takes 30\%, 50\%, and 70\% of the difference.

The two families also point at different parts of the image. On the 786 images localized by both methods, the IoU between the Gemini 3 Flash box and the FakeShield mask is only 0.123 on average (median 0.032), the two predictions do not overlap at all for 19.0\% of the images, and the per-image IoUs of the two methods are only weakly correlated (Pearson $r = 0.346$). Together with the complementarity analysis in Appx.~\ref{app:complementarity}, this suggests that MLLMs and forensic methods rely on different evidence, rather than being weaker and stronger versions of the same detector.

These results indicate that neither family yet provides the region-level evidence that a refund decision would require. Region-level annotations such as those in Appx.~\ref{appx:region_annotations} allow future work to evaluate localization and claim grounding directly, rather than only the binary verdict.

\begin{table}[!htp]
\centering
\caption{\textbf{Localization of the manipulated region.} All methods are scored against the same ground truth, the smallest set of pixels carrying half of the smoothed difference between each fake-damaged image and its real-undamaged source, which covers 0.178 of the frame on average. $n$ is the number of scored images. FakeShield is scored on the 957 fake-damaged images for which it returns a mask, Gemini 3 Flash on the 786 of them that it also flags with a box, and MVSS-Net and ResFCN on every fake-damaged image. Area is the mean fraction of the frame covered by the prediction.}
\label{tab:localization}
\resizebox{\textwidth}{!}{
\begin{tabular}{l|c|c|ccc|c}
\toprule
\textbf{Methods} & \textbf{Output} & $n$ & \textbf{Mean IoU} & \textbf{Median IoU} & \textbf{Pixel F1} & \textbf{Area} \\
\midrule
Gemini 3 Flash & Box & 786 & 0.208 & 0.191 & 0.328 & 0.388 \\
FakeShield & Mask & 957 & 0.086 & 0.026 & 0.137 & 0.059 \\
FakeShield (same images as Gemini 3 Flash) & Mask & 786 & 0.087 & 0.026 & 0.137 & 0.055 \\
\midrule
MVSS-Net [CASIA] & Mask & 5,928 & 0.101 & 0.071 & 0.168 & -- \\
MVSS-Net [DEFACTO] & Mask & 5,928 & 0.006 & 0.000 & 0.010 & -- \\
ResFCN [CASIA] & Mask & 5,928 & 0.024 & 0.001 & 0.041 & -- \\
ResFCN [DEFACTO] & Mask & 5,928 & 0.012 & 0.000 & 0.020 & -- \\
\bottomrule
\end{tabular}}
\end{table}

\subsubsection{Complementarity of MLLMs and Specialized Detectors}
\label{app:complementarity}

The complementarity analysis reveals that MLLMs and specialized detectors succeed on largely different fake-damaged images. For each pair of an MLLM in Tab.~\ref{tab:main} and one of the 14 detector checkpoints in Tab.~\ref{tab:main} and~\ref{tab:ifdl}, we compare over the 5,928 fake-damaged images whether each method detects each image. Across the resulting 224 pairs, the two outcomes are nearly independent, with a median $\phi$ coefficient of 0.031 and $|\phi| < 0.1$ for 193 pairs. For example, Qwen3.8-Max detects 174 of the 910 fake-damaged images missed by ForgeLens [GenImage].

Complementary errors do not, however, translate into large gains under simple fusion. Flagging an image whenever either Qwen3.8-Max or ForgeLens [GenImage] flags it raises the pooled balanced accuracy of ForgeLens [GenImage] by only 0.010, because the additional detections are accompanied by additional false positives on real-damaged evidence. This motivates fusion strategies that weigh the two families according to the evidence each relies on, rather than combining their verdicts directly.
\newpage

\section{Scope of \method{} and the Role of MLLMs}
\label{app:positioning}

\paragraph{What \method{} Measures.} \method{} asks whether a refund claim is supported by genuine visual evidence, which requires three capabilities. The first is \textit{\textbf{visual verification}}, i.e., deciding whether a photograph of damage is genuine when the negative class consists of real damaged evidence rather than clean photographs. Every detector family can be evaluated on this capability, and it is the one measured in Tab.~\ref{tab:main}. The second is \textit{\textbf{claim--image consistency}}, i.e., checking that the evidence shows the damage the customer describes, which requires reading the review text. The third is \textit{\textbf{multi-view aggregation}} across the photographs submitted with one claim. Specialized AI-generated image detectors and IFDL methods score one image at a time and take no text, so only MLLMs can be evaluated on the second and third capabilities, and only MLLMs return a rationale that a platform could present to a customer whose claim is rejected. Following prior forensic benchmarks~\citep{wang2025dfbench,du2025forensichub}, we therefore evaluate MLLMs alongside specialized detectors, rather than because we expect them to be stronger forensic classifiers.

\paragraph{Where Current Models Fall Short.} Our results place the current bottleneck at visual verification. In the single-image setting, the 16 MLLMs in Tab.~\ref{tab:main} retain genuine damaged evidence with an average TNR of 0.938 but detect only 0.230 of the fake-damaged images. Review text is not yet used as a claim to be verified. Across the 18 MLLMs evaluated with and without review text (Tab.~\ref{tab:T2} and~\ref{tab:T2b}), the average balanced accuracy rises only from 0.588 to 0.606, with 12 models improving and six declining, and on the mismatch subset the gain from the matched review (0.595 to 0.622) disappears once the review describes a different product (0.587, Appx.~\ref{app:image_text_mismatch}). Multiple views are used more effectively. Without review text, the average balanced accuracy of the six MLLMs in Tab.~\ref{tab:main_results_review} rises from 0.591 for single images to 0.621 for jointly presented images and 0.700 in the multi-step setting (Tab.~\ref{tab:T2}, \ref{tab:T3}, and~\ref{tab:T5}).

\paragraph{Interpretable Rationales.} A single example illustrates both the value and the risk of MLLM rationales. For a crack that Grok Imagine added to a translucent plastic case (Electronics, Review\_010), Gemini 3 Flash flags the image with a confidence of 0.95, stating that the crack ``exhibits physically impossible geometry, appearing as a solid black void rather than a fracture in translucent material'', whereas ForgeLens [GenImage] assigns it a score of 0.058. Qwen3.8-Max and GPT-6 Astra accept the same image with the opposite argument, e.g., that ``the dark sofa shows through the puncture hole exactly where expected''. Rationales thus make an MLLM decision checkable, but a fluent rationale can still be wrong, which motivates evaluating rationale faithfulness in future work.

\section{Limitations}
\label{app:limitation}

\method{} provides a controlled benchmark for AI-generated refund-evidence verification, but its scope is necessarily bounded. 

The benchmark covers selected e-commerce, food-delivery, travel-service, and self-collected scenarios, and therefore reflects the platforms, categories, languages, and review-media availability included in our collection. It should not be interpreted as covering the full distribution of refund disputes, complaint behaviors, platform policies, or user-submitted evidence formats.

Another limitation concerns the synthetic threat model. 
The fake-damaged evidence is generated using 12 image editing and generation models under a shared prompt-construction pipeline. 
This design enables controlled comparison across generators and damage patterns, but it cannot exhaustively represent future generative models, post-processing strategies, prompt optimization methods, or human-assisted manipulation workflows. As generative models and editing interfaces evolve, detector performance and relative model rankings may change.

The evaluation setting is also narrower than real refund adjudication. 
Our experiments focus on visual and review-conditioned evidence verification, whereas real platform decisions may additionally involve order histories, logistics records, seller responses, transaction metadata, platform policies, and human communication. Accordingly, results on \method{} should be interpreted as evidence of model behavior under the benchmarked verification settings, not as proof of deployment readiness for autonomous refund decisions.

Finally, several evaluated MLLMs are proprietary API-based systems. 
Although we report model identifiers, prompt templates, access details, and evaluation settings, exact reproducibility may be affected by provider-side model updates, backend changes, request-size limits, safety filters, or changes in default inference behavior. We therefore treat the reported API-based results as a snapshot of model behavior under the specified evaluation period and settings.

\section{Compute Reporting}
\label{sec:compute_reporting}

We report the computational resources used for constructing and evaluating \method{}. The study does not train models from scratch, fine-tune MLLMs, or conduct hyperparameter search. Local computation is used for data preprocessing, image downloading and filtering, synthetic-image generation orchestration, specialized-detector inference, and evaluation scripts. MLLM evaluation is performed through official proprietary API services; therefore, provider-side inference compute is not directly observable from our local hardware logs.

\paragraph{Local Compute Environment.}
As mentioned in Appx.~\ref{appx:testbed}, all local experiments are conducted on a single Alibaba Cloud Elastic Compute Service (ECS) instance, \texttt{ecs.gn7e-c16g1.4xlarge}. The instance is equipped with an Intel Xeon Platinum 8369B CPU exposing 16 available vCPUs, 125~GiB memory, 2,048~GiB storage, and one NVIDIA A100 SXM GPU with 80~GB memory. We do not use distributed training, multi-node computation, or multi-GPU computation.

\paragraph{Local Runtime.}
The approximate local runtime is 6 hours for data preprocessing, 6 hours for image downloading and filtering, 24 hours for synthetic-image generation, and 1 GPU-hour for specialized-detector evaluation. These estimates describe the local execution time of our benchmark construction and evaluation pipeline and exclude provider-side compute for proprietary API inference.

\paragraph{API-Based MLLM Evaluation.}
We evaluate 30 MLLMs through official API services. The reported results rest on 510,277 successful API calls: 208,200 single-image review-free calls (30 MLLMs $\times$ 6,940 images), 124,920 single-image review-conditioned calls (18 MLLMs), 27,024 multi-image review-free and 27,024 multi-image review-conditioned calls (eight MLLMs on all 29 categories and ten further MLLMs on the 27 Amazon categories, one call per review), 34,056 multi-step review-free and 34,056 multi-step review-conditioned calls (six MLLMs, one call per image turn), 46,317 prompt-ablation calls (16 MLLMs), and 8,680 mismatched-review calls (15 MLLMs). The multi-image runs of these 18 MLLMs additionally issued 34,564 calls on single-image reviews of the reported categories, which are excluded from the multi-image metrics. By provider, the successful calls comprise 274,937 for Alibaba Cloud DashScope-hosted models, 97,108 for OpenAI-hosted models, 66,010 for Google-hosted models, and 72,222 for xAI-hosted models.

\paragraph{Retries and Failures.}
Our scripts retry transient API errors up to three times with exponential back-off and, for request-size and image-fetch failures, re-encode the images at up to ten progressively smaller resolution tiers, as detailed in Appx.~\ref{appx:other_detail}. The stored records show at least 5,596 additional attempts beyond the successful calls, about 1.1\%; attempts made before a failed record was re-run are not logged, so this is a lower bound. Calls that still failed after all retries were re-issued in a later pass until every reported record had a valid response, except for eight responses in the earlier prompt-ablation and mismatch runs of the additional MLLMs, which could not be parsed and are excluded from the corresponding tables.

\paragraph{Training, Tuning, and Ablations.}
No model is trained from scratch or fine-tuned in this study. No hyperparameter search is performed. The additional model-call experiments consist of fixed prompt-ablation variants and image-text mismatch evaluation. Both are conducted through API-based inference and do not involve parameter updates or model selection.

\section{Impact Statement}
\label{sec:impact}

\paragraph{Intended Use and Deployment Scope.}
This benchmark is intended to support research on detecting AI-generated or AI-modified visual evidence in refund- and complaint-related review scenarios. 
Its primary use is controlled evaluation of detector robustness, failure modes, and multimodal evidence verification. 
It is not designed to serve as an automated decision system for accepting or rejecting refund claims, moderating user reviews, or making consequential decisions about individual users. 
Model outputs should therefore be treated as signals for further review, not as final determinations.

\paragraph{Dual-Use and Misuse Risks.}
The benchmark has dual-use implications. 
While it can support the development and evaluation of more reliable detectors, it may also reveal detector weaknesses that malicious actors could use to produce more convincing, deceptive evidence. 
We mitigate this risk by framing the dataset as a research and evaluation resource, documenting intended uses and limitations in the dataset card, and avoiding operational guidance for generating successful fraudulent claims. 
The synthetic examples are released to enable controlled detector evaluation and failure analysis, not to facilitate deception, fraud, or harassment.

\paragraph{Over-Reliance and User Harm.}
A key downstream risk is over-reliance on imperfect detection models. 
False positives may cause genuinely damaged evidence to be incorrectly flagged as AI-generated, potentially disadvantaging legitimate customers in refund or dispute-resolution workflows. 
False negatives may allow synthetic or manipulated evidence to pass undetected. 
For this reason, the benchmark should not be interpreted as evidence that any evaluated model is ready for autonomous deployment. 
Any real-world use should include human review, uncertainty handling, appeal mechanisms, and platform-specific validation before model outputs affect user outcomes.

\paragraph{Bias, Fairness, and Representativeness.}
The benchmark does not collect protected attributes and is not intended to infer or predict sensitive characteristics such as gender, race, ethnicity, religion, sexuality, health status, or political affiliation. 
Nevertheless, performance may vary across platforms, languages, regions, product categories, service domains, imaging conditions, and user devices. 
Models trained or evaluated on this benchmark should therefore not be assumed to generalize uniformly to all review ecosystems or user populations. 
We avoid claims of universal representativeness and document the dataset scope, source platforms, category coverage, and language coverage to reduce the risk of inappropriate generalization.

\paragraph{Privacy, Surveillance, and Human Rights.}
The benchmark is based on review evidence rather than surveillance data, and it is not intended for identity recognition, user profiling, behavioral monitoring, or inference of protected attributes. 
However, related detection systems could be misused if deployed as broad monitoring tools over user-generated content or linked to punitive actions without transparency or recourse. 
We therefore recommend that downstream applications restrict use to content-integrity assessment, avoid identity-linked profiling, and provide meaningful human oversight for any decision that may affect users' access to services, refunds, or dispute outcomes.

\paragraph{Security and Environmental Considerations.}
The main security-related concern is adversarial adaptation, where attackers use benchmark feedback to improve synthetic evidence against current detectors. This risk is inherent to evaluation work on deceptive media, and we address it by emphasizing aggregate evaluation, failure-mode analysis, and responsible use rather than operational fraud guidance. Our experiments are limited to benchmark construction, specialized-detector evaluation on a single GPU instance, and API-based inference for proprietary MLLMs; the work does not involve training large foundation models from scratch. While API-based evaluation incurs additional upstream compute across proprietary model providers that is not directly observable from our local hardware logs, the associated energy footprint remains substantially different from that of large-scale foundation-model training, and no model parameters are updated in our study. Future work involving repeated large-scale MLLM evaluation or detector training should report compute resources, API usage, energy estimates, and evaluation scale transparently, particularly as provider-side energy reporting becomes more widely available.

\section{Responsible Research Practices}
\label{sec:responsible_practice}

\paragraph{Privacy and Personally Identifiable Information.}
The real-review portion of the benchmark is derived from publicly accessible sources, including the Amazon Reviews 2023 dataset and publicly accessible Trip.com and GrabFood review records obtained through API-based requests and direct collection. 
Because user-generated review media can contain personally identifiable information, we apply automated MLLM-based filtering followed by human inspection to reduce privacy risk before release. 
We remove images identified as containing sensitive personal information, such as visible faces, license plates, addresses, screenshots, or other directly identifying content.

For released metadata, we retain only fields needed for benchmark use, traceability, and reproducibility. Display names, avatar links, and review URLs of Trip.com and GrabFood reviews are retained as publicly displayed by the source platforms, and we add no identifying information beyond them.

\paragraph{Consent and Public Data Use.}
The dataset is constructed from publicly accessible review data and public data sources. We did not obtain explicit consent from each review author because the records were publicly available at the time of collection, and contacting individual review authors at scale was not feasible. To mitigate this limitation, we restrict the released benchmark to research use, add no identifying metadata beyond what the source platforms display, filter images for personally sensitive content, and document the data sources, collection procedure, and intended use in the dataset card. We also document platform-specific access constraints in the dataset card and instruct users to comply with the applicable terms of the original platforms and data sources. We have additionally confirmed that all primary data sources used in this study, including Amazon Reviews 2023~\citep{hou2024bridging}, remain publicly available and have not been deprecated or withdrawn by their original authors at the time of submission.

\paragraph{Copyright and Usage Restrictions.}
The benchmark contains original review images and metadata together with synthetic images generated by our pipeline. 
We do not claim ownership over third-party review text, images, or platform metadata. 
Because the copyright and licensing status of review media is not uniform across source platforms, the dataset card documents data provenance, intended research use, and usage restrictions. 
Users of the benchmark are responsible for complying with the applicable terms of the original platforms and data sources. 
Synthetic images generated by our pipeline are released to support reproducibility and controlled evaluation.

\paragraph{Representative Evaluation Practice.}
The benchmark is designed to evaluate AI-generated or AI-modified evidence in refund- and complaint-related review scenarios, rather than to represent all online reviews or all forms of e-commerce fraud. The Amazon subset covers 27 product categories selected according to platform taxonomy and common product-review categories, while Trip.com and GrabFood provide two additional service-review domains covering hotel and food-delivery scenarios. 

The review text is not restricted to English; when available, we retain the original language, including Simplified Chinese, Traditional Chinese, Japanese, Korean, Spanish, Russian, and other languages. At the same time, the benchmark should not be interpreted as universally representative of all platforms, regions, languages, cultures, product types, service domains, or fraud behaviors. 
Amazon and Trip.com serve broad international user populations, whereas GrabFood is primarily associated with Southeast Asian food-delivery markets. We therefore avoid claims of universal coverage and state in the limitations that the benchmark reflects the selected platforms, available review media, and the product and service categories included in our collection.

\end{document}